\documentclass[12pt]{article}
\usepackage[hidelinks]{hyperref}
\usepackage{arxiv}
\usepackage[english]{babel}
\usepackage[utf8]{inputenc}
\usepackage{array}
\usepackage{booktabs}
\usepackage{caption}
\usepackage[T1]{fontenc}    
\usepackage{microtype}      
\usepackage{longtable}
\usepackage{url}
\usepackage{makecell}

\usepackage{nicefrac}       
\usepackage{doi}

\usepackage{varwidth}
\usepackage[symbol]{footmisc}
\usepackage{amsmath}
\usepackage{graphicx}
\usepackage[inkscapearea=page, inkscapelatex=true]{svg}
\usepackage[super,numbers,sort&compress]{natbib}
\usepackage{natmove}

\def\c{\textcolor{green}{\tikz\fill[scale=0.4](0,.35) -- (.25,0) -- (1,.7) -- (.25,.15) -- cycle;}}

\def\view{\includegraphics[width=10px, height=10px]{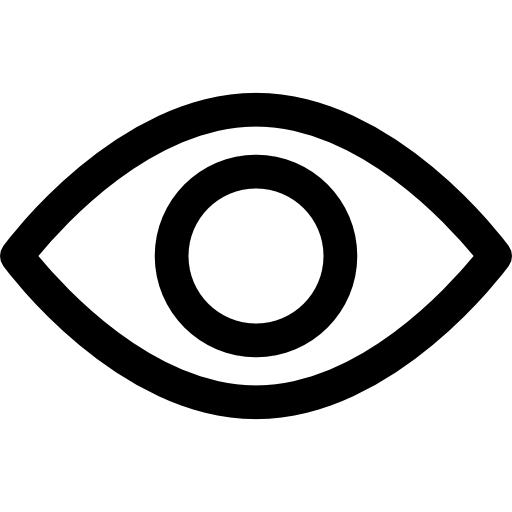}}
\def\audio{\includegraphics[width=10px, height=10px]{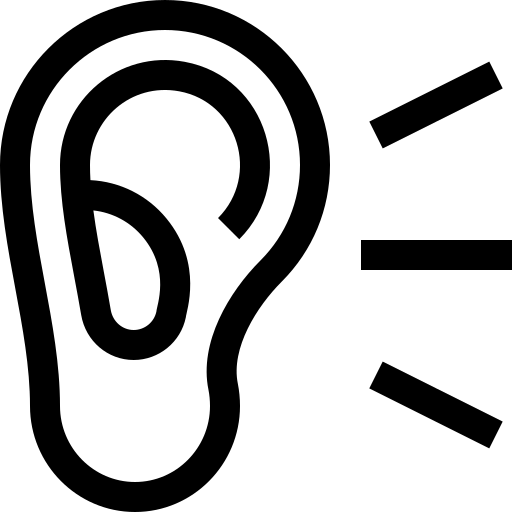}}
\def\W{\includegraphics[width=10px, height=10px]{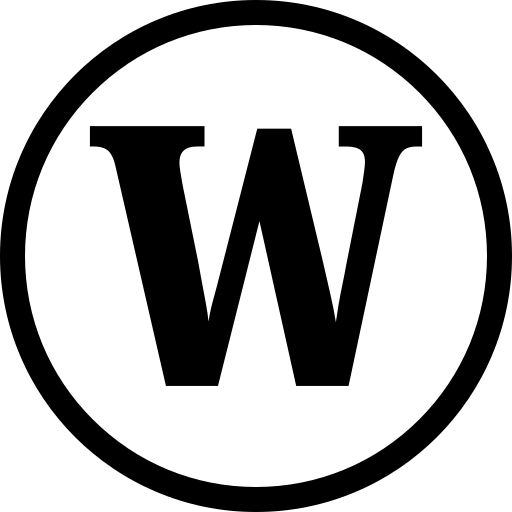}}
\def\S{\includegraphics[width=10px, height=10px]{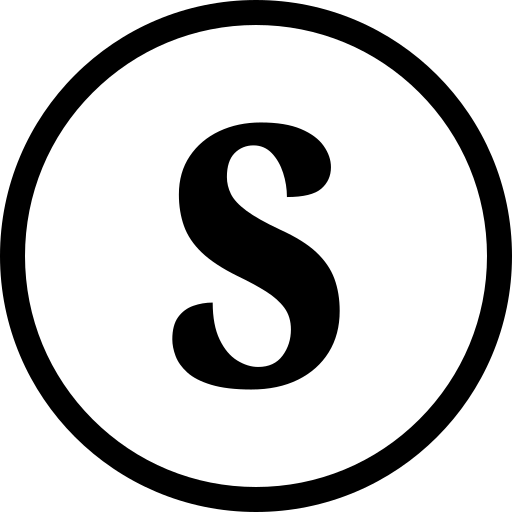}}
\def\L{\includegraphics[width=10px, height=10px]{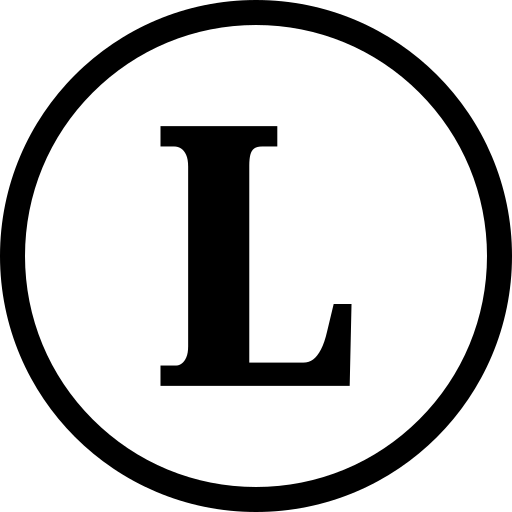}}
\def\agent{\includegraphics[width=10px, height=10px]{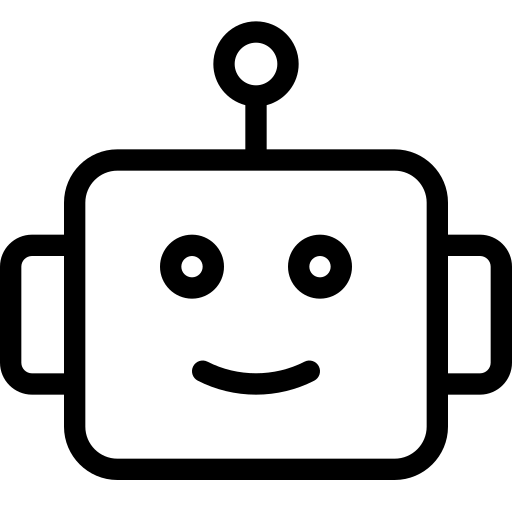}}

\usepackage[export]{adjustbox}
\usepackage{enumitem}

\usepackage{setspace} 
\usepackage{colortbl}
\definecolor{lightgray}{gray}{0.9}
\definecolor{C0}{RGB}{ 27, 158, 119}
\definecolor{C1}{RGB}{217,  95,   2}
\definecolor{C2}{RGB}{117, 112, 179}
\definecolor{C3}{RGB}{231,  41, 138}

\usepackage[edges]{forest}
\usepackage{tikz}
\usetikzlibrary{shapes, arrows, positioning}
\tikzset{%
    root/.style = {align=center,text width=1.0cm,rounded corners=3pt, line width=0.3mm, fill=C0!20, draw=gray},
    task/.style = {align=center,text width=1.5cm,rounded corners=3pt, fill=C1!20, draw=gray},
    application/.style = {align=center,text width=2.0cm,rounded corners=3pt, fill=C2!20, draw=gray},
    study/.style = {align=center,text width=2.5cm,rounded corners=3pt, fill=C3!20, draw=gray},
}

\title{A Review of Large Language Models and Autonomous Agents in Chemistry}
\date{\today}
\author{
    \href{https://orcid.org/0000-0001-5336-2847}{\includegraphics[scale=0.06]{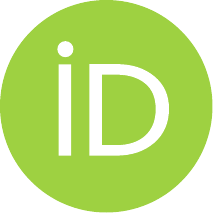}\hspace{1mm}Mayk Caldas Ramos}\\
	FutureHouse Inc., San Francisco, CA\\
    Department of Chemical Engineering \\
	University of Rochester, Rochester, NY\\
	\texttt{mcaldasr@ur.rochester.edu} \\
\And
	\href{https://orcid.org/}{\includegraphics[scale=0.06]{orcid.pdf}\hspace{1mm}Christopher J. Collison} \\
	School of Chemistry and Materials Science \\
	Rochester Institute of Technology, Rochester, NY\\
	\texttt{cjcscha@rit.edu} \\
\And
	\href{https://orcid.org/0000-0002-6647-3965}{\includegraphics[scale=0.06]{orcid.pdf}\hspace{1mm}Andrew D. White}\thanks{Corresponding author} \\
    FutureHouse Inc., San Francisco, CA\\
	Department of Chemical Engineering \\
	University of Rochester, Rochester, NY\\
	\texttt{andrew@futurehouse.org} \\
}

\renewcommand{\shorttitle}{LLMs and Autonomous LLM-based Agents in Chemistry}

\hypersetup{
pdftitle={A Review of Large Language Models and Autonomous Agents in Chemistry},
pdfsubject={},
pdfauthor={Caldas Ramos, Mayk and Collison, Christopher J and White, Andrew D},
pdfkeywords={Large Language Model, LLM, LLM agent, agent, science, chemistry},
}

\begin{document}

\maketitle

\begin{abstract}
Large language models (LLMs) have emerged as powerful tools in chemistry, significantly impacting molecule design, property prediction, and synthesis optimization. 
This review highlights LLM capabilities in these domains and their potential to accelerate scientific discovery through automation. 
We also review LLM-based autonomous agents: LLMs with a broader set of tools to interact with their surrounding environment. These agents perform diverse tasks such as paper scraping, interfacing with automated laboratories, and synthesis planning. As agents are an emerging topic, we extend the scope of our review of agents beyond chemistry and discuss across any scientific domains.
This review covers the recent history, current capabilities, and design of LLMs and autonomous agents, addressing specific challenges, opportunities, and future directions in chemistry.
Key challenges include data quality and integration, model interpretability, and the need for standard benchmarks, while future directions point towards more sophisticated multi-modal agents and enhanced collaboration between agents and experimental methods.
Due to the quick pace of this field, a repository has been built to keep track of the latest studies: \url{https://github.com/ur-whitelab/LLMs-in-science}.

\end{abstract}

\tableofcontents

\section{Introduction}

The integration of Machine Learning (ML) and Artificial Intelligence (AI) into chemistry has spanned several decades\cite{willett2011chemoinformatics, griffen_chemists_2020, baum_artificial_2021, ayres_taking_2021, yang_concepts_2019, mater2019deep, shi2023machine, keith2021combining, kuntz2022machine, meuwly2021machine}. Although applications of computational methods in quantum chemistry and molecular modeling from the 1950s-1970s were not considered AI, they laid the groundwork. Subsequently in the 1980s expert systems like DENDRAL\cite{lederberg1969applications, lindsay1993dendral} were expanded to infer molecular structures from mass spectrometry data.\cite{buchanan_applications_1976} At the same time, Quantitative Structure-Activity Relationship (QSAR) Models were developed\cite{yang_concepts_2019} that would use statistical methods to predict the effects of chemical structure on activity.\cite{hansch1962correlation, hansch1964p, hansch1995exploring, topliss1972chance} In the 1990s, neural networks, and associated Kohonen Self-Organizing Maps were introduced to domains such as drug design,\cite{weinstein_neural_1992, van1994use} as summarized well by \citet{yang_concepts_2019} and \citet{goldman2006machine}, although they were limited by the computational resources of the time. With an explosion of data from High-Throughput Screening (HTS)\cite{pereira2007origin, medina-franco_shifting_2013}, models then started to benefit from vast datasets of molecular structures and their biological activities. Furthermore, ML algorithms such as Support Vector Machines and Random Forests became popular for classification and regression tasks in cheminformatics,\cite{willett2011chemoinformatics} offering improved performance over traditional statistical methods.\cite{butler_machine_2018}

Deep learning transformed the landscape of ML in chemistry and materials science in the 2010s.\cite{rupp_fast_2012} Recurrent Neural Networks (RNNs),\cite{Olivecrona2017-is, Segler2018-ef, segler_generating_2018, Gupta2018-pj, Karpov2020-yf} Convolutional Neural Networks (CNNs)\cite{schutt_schnet_2018, Hirohara2018-wx, Coley2019-it} and later, Graph Neural Networks (GNNs),\cite{dwivedi2023benchmarking, sanchez2021gentle, bronstein2017geometric, wu2020comprehensive, gilmer2017neural, gomez-bombarelli_automatic_2018} made great gains in their application to molecular property prediction, drug discovery,\cite{gaudelet2021utilizing} and synthesis prediction.\cite{choudhary_recent_2022} Such methods were able to capture complex patterns in data, and therefore enabled the identification of novel materials for high-impact needs such as energy storage and conversion.\cite{Fung2021-ve, Reiser2022-ph}

In this review, we explore the next phase of AI in chemistry, namely the use of Large Language Models (LLMs) and autonomous agents. Inspired by successes in natural language processing (NLP), LLMs 
were adapted for chemical language (e.g., Simplified Molecular Input Line Entry System (SMILES)\cite{weininger_smiles_1988}) to tackle tasks from synthesis prediction to molecule generation\cite{Chithrananda2020-cd, li_mol-bert_2021, Wang2023-jd}. We will then explore the integration of LLMs into autonomous agents as illustrated by \citet{Bran2023-jk} and \citet{Boiko2023-ot}, which may be used for data interpretation or, for example, to experiment with robotic systems. 
We are at a crossroads where AI enables chemists to solve major global problems faster and streamline routine lab tasks. This enables, for instance, the development of larger, consistent experimental datasets and shorter lead times for drug and material commercialization.
As such, language has been the preferred mechanism for describing and disseminating research results and protocols in chemistry for hundreds of years.\cite{White2023-gs}

\subsection{Challenges in Chemistry}

\begin{figure}[h!]
    \centering
    \includegraphics[width=0.5\linewidth]{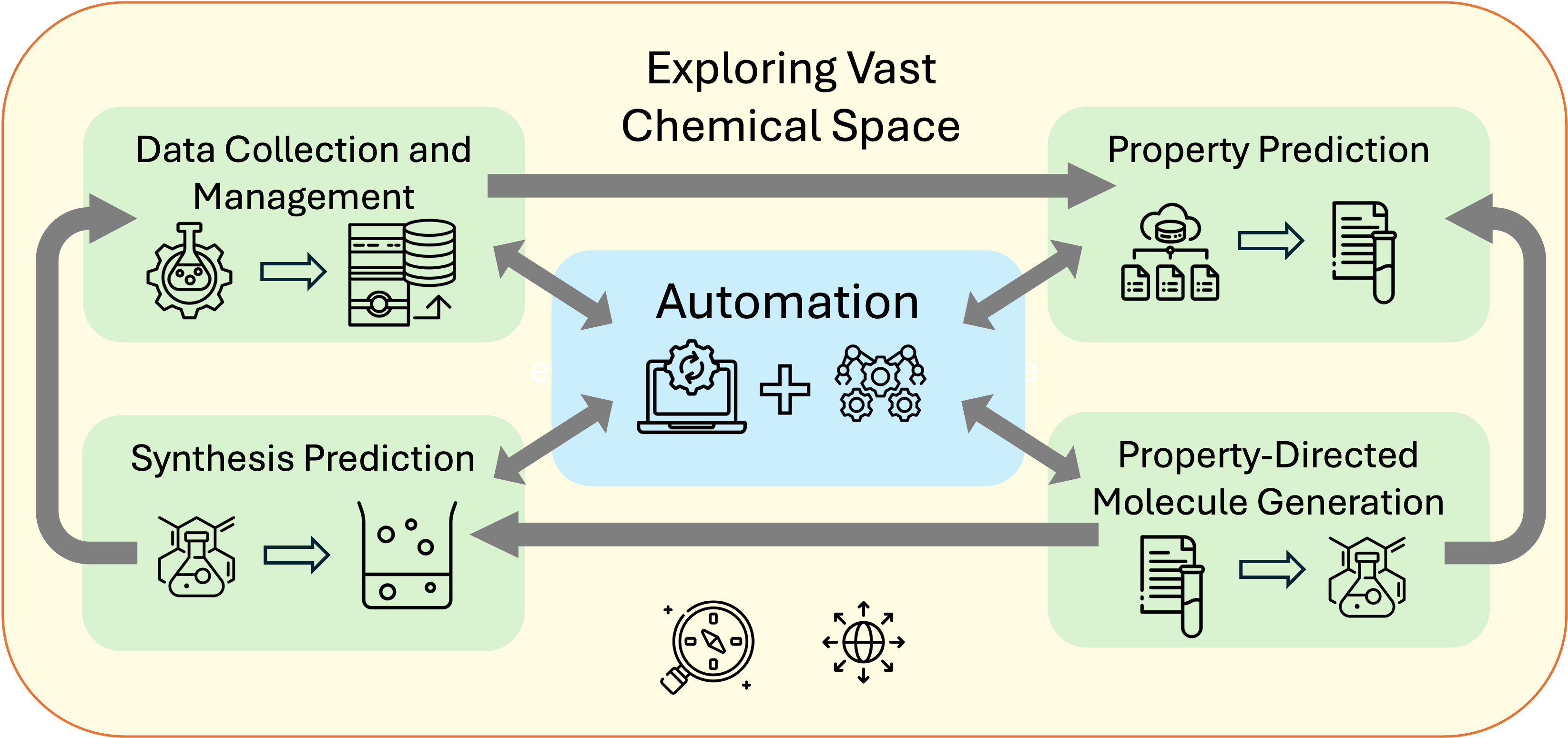}
    \caption{AI-powered LLMs accelerate chemical discovery with models that address key challenges in Property Prediction, Property Directed Molecule Generation, and Synthesis Prediction. Autonomous agents connect these models and additional tools thereby enabling rapid exploration of vast chemical spaces.}
    \label{fig:challenge_scheme}
\end{figure}

We categorize some key challenges that can be addressed by AI in chemistry as: Property Prediction, Property-Directed Molecule Generation, and Synthesis Prediction. These categories, as illustrated in Figure \ref{fig:challenge_scheme} can be connected to a fourth challenge in automation.
The first task is to predict a property for a given compound to decide if it should be synthesized for a specific application, such as an indicator,\cite{collison_complexation_2008} light harvester,\cite{wiegand_directional_2024} or catalyst.\cite{ahmadov_dual_2024} To achieve better models for property prediction, high-quality data is crucial. We discuss the caveats and issues with the current datasets in Section \ref{sec:benchmark} and illustrate state-of-the-art findings in Section \ref{sec:sciEncoder}.

The second task is to generate novel chemical structures that meet desired chemical profiles or exhibit properties.\cite{fischer_approaching_2019} Success in this area would accelerate progress in various chemical applications, but reliable reverse engineering (inverse design)\cite{wang_data-driven_2022} is not yet feasible over the vast chemical space.\cite{sridharan_modern_2022} For instance, inverse design, when coupled with automatic selection of novel structures (\textit{de novo} molecular design) could lead to the development of drugs targeting specific proteins while retaining properties like solubility, toxicity, and blood-brain barrier permeability.\cite{wu_moleculenet_2018} The complexity of connecting \textit{de novo} design with property prediction is high and we show how state-of-the-art models currently perform in Section \ref{sec:SciDecoder}. 

Once a potential target molecule has been identified, the next challenge is predicting its optimal synthesis using inexpensive, readily available, and non-toxic starting materials. In a vast chemical space, there will always be an alternative molecule ''B''  that has similar properties to molecule ''A'' but is easier to synthesize. Exploring this space to find a new molecule with the right properties and a high-yield synthesis route brings together these challenges. The number of possible stable chemicals is estimated to be up to $10^{180}$.\cite{restrepo_chemical_2022, kirkpatrick2004chemical, mullard2017drug, llanos2019exploration} Exploring this vast space requires significant acceleration beyond current methods.\cite{schrier_pursuit_2023} As \citet{restrepo_chemical_2022} emphasizes, cataloguing failed syntheses is essential to building a comprehensive dataset of chemical features. Autonomous chemical resources can accelerate database growth and tackle this challenge. Thus, automation is considered a fourth major task in chemistry.\cite{gromski2019explore, steiner2019organic, burger2020mobile, macleod2020self} The following discussion explores how LLMs and autonomous agents can provide the most value. Relevant papers are discussed in Section \ref{sec:sciEncDec}

This review is organized within the context of these categories. The structure of the review is as follows. Section 2 provides an introduction to transformers, including a brief description of encoder-only, decoder-only and encoder-decoder architectures. Section 3 provides a detailed survey of work with LLMs, where we connect each transformer architecture to the areas of chemistry that it is best suited to support. We then progress into a description of autonomous agents in section 4, and a survey of how such LLM-based agents are finding application in chemistry-centered scientific research, section 5. After providing some perspective on future challenges and opportunities in section 6, and we conclude in section 7. We distinguish between ``text-based'' and ``mol-based'' inputs and outputs, with ``text'' referring to natural language and ``mol'' referring to the chemical syntax for material structures, as introduced by \citet{Zhang2024-na}.

\section{Large Language Models}

The prior state-of-the-art for sequence-to-sequence (seq2seq) tasks had been the Recurrent Neural Network (RNN),\cite{rumelhart1986learning} typically as implemented by \citet{hochreiter1997long}. In a seq2seq task, an input sequence, such as a paragraph in English, is processed to generate a corresponding output sequence, such as a translation into French. The RNN retains ``memory'' of previous steps in a sequence to predict later parts. However, as sequence length increases, gradients can become vanishingly small or explosively large\cite{ribeiro_beyond_2020, or_exploding_2023}, preventing effective use of earlier information in long sequences. Due to these limitations, RNNs have thus fallen behind Large Language Models (LLMs), which primarily implement transformer architectures, introduced by \citet{Vaswani2017-kq}.
LLMs are deep neural networks (NN) characterized by their vast number of parameters and, though transformers dominate, other architectures for handling longer input sequences are being actively explored.\cite{Gu2023-gp,Jelassi2024-dj,peng2023rwkv, beck_xlstm_2024} A detailed discussion of more generally applied LLMs can be found elsewhere.\cite{Minaee2024-ii} Since transformers are well-developed in chemistry and are the dominant paradigm behind nearly all state-of-the-art sequence modeling results, they are a focus in this review.

\subsection{The Transformer}

\begin{figure}[h!]
    \centering
    \includegraphics[width=1.0\textwidth]{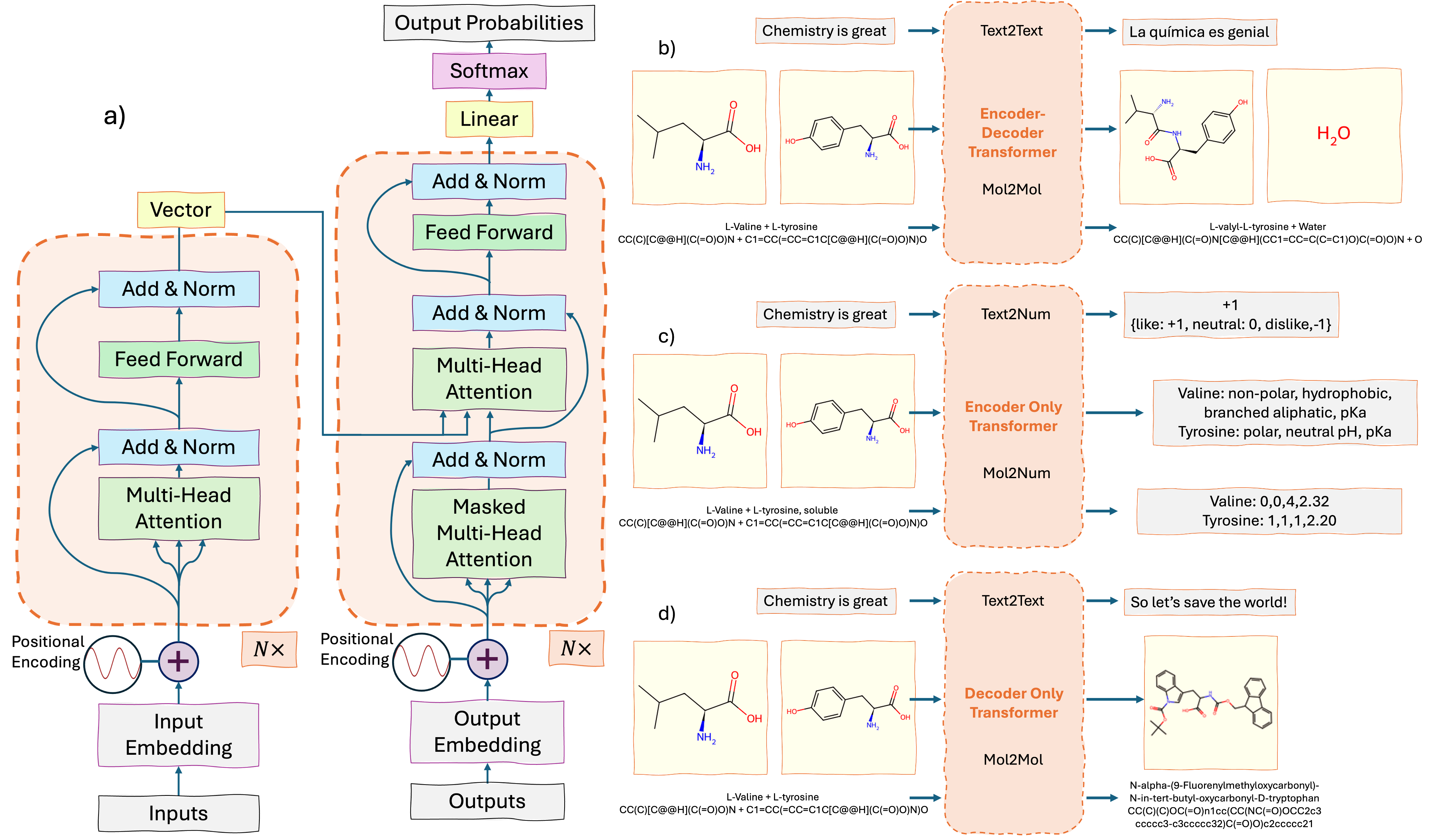}
    \caption{a) The generalized encoder-decoder transformer: The encoder on the left converts an input into a vector, while the decoder on the right predicts the next token in a sequence. b) Encoder-decoder transformers are traditionally used for translation tasks and, in chemistry, for reaction prediction, translating reactants into products. c) Encoder-only transformers provide a vector output and are typically used for sentiment analysis. In chemistry, they are used for property prediction or classification tasks. d) Decoder-only transformers generate likely next tokens in a sequence. In chemistry, they are used to generate new molecules given an instruction and description of molecules.
    }
    \label{fig:transformer}
\end{figure}

The transformer was introduced in, ``Attention is all you need'' by \citet{Vaswani2017-kq} in 2017. A careful line-by-line review of the model can be found in ``The Annotated Transformer'' \cite{annotated_transformer}. The transformer was the first seq2seq model based entirely on attention mechanisms, although attention had been a feature for RNNs some years prior.\cite{bahdanau_neural_2016} The concept of ``attention'' is a focus applied to certain words of the input, which would convey the most importance, or the context of the passage, and thereby would allow for better decision-making and greater accuracy. However, in a practical sense, ``attention'' is implemented simply as the dot-product between token embeddings and a learned non-linear function, which will be described further below.

\paragraph{Context Window} Large language models are limited by the size of their context window, which represents the maximum number of input tokens they can process at once. 
This constraint arises from the quadratic computational cost of the transformer's attention mechanism, which restricts effective input to a few thousand tokens.\cite{Li2024-zp}
Hence, LLM-based agents struggle to maintain coherence and capture long-range dependencies in extensive texts or complex dialogues, impacting their performance in applications requiring deep contextual understanding.\cite{Zhang2024-xw}
These limitations and strategies to overcome them are better discussed in Section~\ref{sec:agents}.

\paragraph{Tokenization} In NLP tasks, the natural language text sequence, provided in the context window, is first converted to a list of tokens, which are integers that each represent a fragment of the sequence. Hence the input is numericized according to the model's vocabulary following a specific tokenization scheme.\cite{Kudo2018-bf, Kudo2018-aq, Song2020-lo, Rust2020-vs, Berglund2023-nz} 

\paragraph{Input Embeddings} Each token is then converted into a vector in a process called input embedding. This vector is a learned representation that positions tokens in a continuous space based on their semantic relationships. This process allows the model to capture similarities between tokens, which is further refined through mechanisms like attention (discussed below) that weigh and enhance these semantic connections.

\paragraph{Positional Encoding} A positional encoding is then added, which plays a major role in transformer success. It is added to the input embeddings to provide information about the order of elements in a sequence, as transformers lack a built-in notion of sequence position. \citet{Vaswani2017-kq} reported similar performance with both fixed positional encoding based on sine and cosine functions, and learned encodings. However, many options for positional embeddings exist.\cite{Gehring2017-zj}. In fixed positional encoding, the position of each element in a sequence is encoded using sine and cosine functions with different frequencies, depending on the element's position. This encoding is then added to the word's vector representation (generated during the tokenization and embedding process). The result is a modified vector that encodes both the meaning of the word and its position within the sequence. These sine and cosine functions generate values within a manageable range of -1 to 1, ensuring that each positional encoding is unique and that the encoding is unaffected by sequence length.

\paragraph{Attention} 
The concept of ``attention'' is central to the transformer's success, especially during training. Attention enables the model to focus on the most relevant parts of the input data. It operates by comparing each element in a sequence, such as a word, to every other element. Each element serves as a \emph{query}, compared against other elements called \emph{keys}, each associated with a corresponding value. The alignment between a \emph{query} and a \emph{keys}, determines the strength of their connection, represented by an \emph{attention weight}.\cite{Devlin2018-nl}
These weights highlight the importance of certain elements by scaling their associated values accordingly.
During training, the model learns to adjust these weights, capturing relationships and contextual information within the sequence. Once trained, the model uses these learned weights to integrate information from different parts of the sequence, ensuring that its output remains coherent and contextually aligned with the input.

The transformer architecture is built around two key modules: the encoder and the decoder. Figure \ref{fig:transformer}a provides a simplified diagram of the general encoder-decoder transformer architecture.
The input is The input is tokenized, from the model's vocabulary,\cite{Kudo2018-bf, Kudo2018-aq, Song2020-lo, Rust2020-vs, Berglund2023-nz} embedded and positionally encoded, as described above. The encoder consists of multiple stacked layers (six layers in the original model),\cite{Vaswani2017-kq} with each layer building on the outputs of the previous one. Each token is represented as a vector, that gets passed through these layers. At each encoder layer, a self-attention mechanism is applied, which calculates the attention between tokens, as discussed earlier. Afterward, the model uses normalization and adds the output back to the input through what's called a residual connection. Residual connection is represented in Figure \ref{fig:transformer}a by the ``by-passing'' arrow. This bypass helps prevent issues with vanishing gradients,\cite{ribeiro_beyond_2020, or_exploding_2023} ensuring that information flows smoothly through the model. The final step in each encoder layer is a feed-forward neural network with an activation function (such as ReLU,\cite{nair2010rectified} SwiGLU,\cite{shazeer_glu_2020} GELU,\cite{hendrycks_gaussian_2023} etc) that further refines the representation of the input.

The decoder works similarly to the encoder but with key differences. 
It starts with an initial input token -- usually a special start token—embedded into a numerical vector. 
This token initiates the output sequence generation. 
Positional encodings are applied to preserve the token order. 
The decoder is composed of stacked layers, each containing a masked self-attention mechanism that ensures the model only attends to the current and previous tokens, preventing access to future tokens. Additionally, an encoder-decoder attention mechanism aligns the decoder's output with relevant encoder inputs, as depicted by the connecting arrows in Figure \ref{fig:transformer}a. 
This alignment helps the model focus on the most critical information from the input sequence. Each layer also employs normalization, residual connections, and a feed-forward network. The final layer applies a softmax function, converting the scores into a probability density over the vocabulary of tokens. The decoder generates the sequence autoregressively, predicting each token based on prior outputs until an end token signals termination.

\subsection{Model training}

The common lifetime of an LLM consists of being first pretrained using self-supervised techniques, generating what is called a base model. Effective prompt engineering may lead to successful task completion but this base model is often fine-tuned for specific applications using supervised techniques and this creates the ``instruct model.'' It is called the ``instruct model'' because the fine-tuning is usually done for it to follow arbitrary instructions, removing the need to specialize fine-tuning for each downstream task\cite{Ouyang2023-qd}. Finally, the instruct model can be further tuned with reward models to improve human preference or some other non-differentiable and sparse desired character\cite{stiennon2020learning}. These concepts are expanded on below.

\paragraph{Self-supervised Pretraining}

A significant benefit implied in all the transformer models described in this review is that self-supervised learning takes place with a vast corpus of text. Thus, the algorithm learns patterns from unlabeled data, which opens up the model to larger datasets that may not have been explicitly annotated by humans. The advantage is to discover underlying structures or distributions without being provided with explicit instructions on what to predict, nor with labels that might indicate the correct answer.

\paragraph{Prompt Engineering}

The model's behavior can be guided by carefully crafting input prompts that leverage the pretrained capabilities of LLMs. Since the original LLM remains unchanged, it retains its generality and can be applied across various tasks.\cite{Brown2020-ej} However, this approach relies heavily on the assumption that the model has adequately learned the necessary domain knowledge during pretraining to achieve an appropriate level of accuracy in a specific domain. Prompt engineering can be sensitive to subtle choices of language; small changes in wording can lead to significantly different outputs, making it challenging to achieve consistent results and to quantify the accuracy of the outputs.\cite{Errica2024-nm} 

\paragraph{Supervised Fine-tuning} After this pretraining, many models described herein are fine-tuned on specific downstream tasks (e.g., text classification, question answering) using supervised learning. In supervised learning, models learn from labeled data, and map inputs to known outputs. Such fine-tuning allows the model to be adjusted with a smaller, task-specific dataset to perform well on that downstream task.

\paragraph{LLM Alignment}
A key step after model training is aligning the output with human preferences. This process is critical to ensure that the large language model (LLM) produces outputs that are not only accurate but also reflect appropriate style, tone, and ethical considerations. Pretraining and fine-tuning often do not incorporate human values, so alignment methods are essential to adjust the model's behavior, including reducing harmful outputs.\cite{Shen2023-em}

One important technique for LLM alignment is instruction tuning. This method refines the model by training it on datasets that contain specific instructions and examples of preferred responses. By doing so, the model learns to generalize from these examples and follow user instructions more effectively, leading to outputs that are more relevant and safer for real-world applications.\cite{Lee2024-yx, Hewitt2024-vr}
Instruction tuning establishes a baseline alignment, which can then be further improved in the next phase using reinforcement learning (RL).\cite{Zhang2023-di}

In RL-based alignment, the model generates tokens as actions and receives rewards based on the quality of the output, guiding the model to optimize its behavior over time. Unlike post-hoc human evaluations, RL actively integrates preference feedback during training, refining the model to maximize cumulative rewards. This approach eliminates the need for token-by-token supervised fine-tuning by focusing on complete outputs, which better capture human preferences.\cite{Duan2016-qs, Ziegler2019-da, Mazuz2023-gb}

The text generation process in RL is typically modeled as a Markov Decision Process (MDP), where actions are tokens, and rewards reflect how well the final output aligns with human intent.\cite{Laskin2022-cn} A popular method, Reinforcement Learning with Human Feedback (RLHF),\cite{OpenAI2022-RLHF} leverages human input to shape the reward system, ensuring alignment with user preferences. Variants such as reinforcement learning with synthetic feedback (RLSF)\cite{Kim2023-hf}, Proximal Policy Optimization (PPO)\cite{Schulman2017-lw}, and REINFORCE\cite{Zhang2020-ji} offer alternative strategies for assigning rewards and refining model policies.\cite{Duan2016-qs, Laskin2022-cn, Shinn2023-kt, Akyurek2023-ms} A broader exploration of RL's potential in fine-tuning LLMs is available in works by \citet{Cao2024-yf} and \citet{Shen2023-em}

There are ways to reformulate the RLHF process into a direct optimization problem with a different loss. This is known as reward-free metods. Among the main examples of reward-free methods, we have the direct preference optimization (DPO)\cite{Rafailov2023-hc}, Rank Responses to align Human Feedback (RRHF)\cite{Yuan2023-yq}, and Preference Ranking Optimization (PRO)\cite{Song2023-hi}.
These models are popular competitors to PPO and other reward-based methods due to its simplicity. It overcomes the lack of token-by-token loss signal by comparing two completions at a time. 
The discussions about which technique is superior remain very active in the literature.\cite{Xu2024-vg}

Finally, the alignment may not be to human preferences but to downstream tasks that do not provide token-by-token rewards. For example, \citet{bou2024acegen} and \citet{hayes2024simulating} both use RL on a language model for improving its outputs on a downstream scientific task.

\subsection{Model types}

While the Vaswani Transformer\cite{Vaswani2017-kq} employed an encoder-decoder structure for sequence-to-sequence tasks, the encoder and decoder were ultimately seen as independent models, leading to ``encoder-only'', and ``decoder-only'' models described below.

Examples of how such models can be used are provided in Figures~\ref{fig:transformer}b, c, and d.
Figure \ref{fig:transformer}b illustrates the encoder-decoder model's capability to transform sequences, such as translating from English to Spanish or predicting reaction products by mapping atoms from reactants (amino acids) to product positions (a dipeptide and water). This architecture has large potential on sequence-to-sequence transformations.\cite{Pei2023-bp, Pei2024-ph}
Figure \ref{fig:transformer}c highlights the strengths of an encoder-only model in extracting properties or insights directly from input sequences. For example, in text analysis, it can assign sentiment scores or labels, such as tagging the phrase ``Chemistry is great'' with a positive sentiment. In chemistry, it can predict molecular properties, like hydrophobicity or pKa, from amino acid representations, demonstrating its applications in material science and cheminformatics.\cite{Li2021-xw, Qian2023-ya, nguyen-vo_predicting_2022}
Finally, Figure \ref{fig:transformer}d depicts a decoder-only architecture, ideal for tasks requiring sequence generation or completion. This model excels at inferring new outputs from input prompts. For instance, given that ``chemistry is great,'' it can propose broader implications or solutions. It can also generate new peptide sequences from smaller amino acid fragments, showcasing its ability to create novel compounds. This generative capacity is particularly valuable in drug design, where the goal is to discover new molecules or expand chemical libraries.\cite{Jin2019-uw, Chithrananda2020-cd, ahmad_chemberta-2_2022, Taylor2022-pu}

\subsubsection{Encoder-only Models}

Beyond Vaswani's transformer,\cite{Vaswani2017-kq} used for sequence-to-sequence tasks, another significant evolutionary step forward came in the guise of the Bidirectional Encoder Representations from Transformers, or ``BERT'', described in October 2018 by \citet{Devlin2018-nl}
BERT utilized only the encoder component, achieving state-of-the-art performance on sentence-level and token-level tasks, outperforming prior task-specific architectures.\cite{Devlin2018-nl} The key difference was BERT's bidirectional transformer pretraining on unlabeled text, meaning the model processes the context both to the left and right of the word in question, facilitated by a Masked Language Model (MLM). This encoder-only design allowed BERT to develop more comprehensive representations of input sequences, rather than mapping input sequences to output sequences. In pretraining, BERT also uses Next Sentence Prediction (NSP). ``Sentence'' here means an arbitrary span of contiguous text. The MLM task randomly masks tokens and predicts them by considering both preceding and following contexts simultaneously, inspired by Taylor\cite{taylor_cloze_1953}. NSP predicts whether one sentence logically follows another, training the model to understand sentence relationships. This bidirectional approach allows BERT to recognize greater nuance and richness in the input data.

Subsequent evolutions of BERT include, for example, RoBERTa, (Robustly optimized BERT approach), described in 2019 by \citet{liu_roberta_2019}. RoBERTa was trained on a larger corpus, for more iterations, with larger mini-batches, and longer sequences, improving model understanding and generalization. By removing the NSP task and focusing on the MLM task, performance improved. RoBERTa dynamically changed masked positions during training and used different hyperparameters. Evolutions of BERT also include domain-specific pretraining and creating specialist LLMs for fields like chemistry, as described below (see Section \ref{sec:LLM_in_science}).

\subsubsection{Decoder-only Models}

In June 2018, \citet{radford2018improving} proposed the Generative Pretrained Transformer (GPT) in their paper, ``Improving Language Understanding by Generative Pretraining''. GPT used a decoder-only, left-to-right unidirectional language model to predict the next word in a sequence based on previous words, without an encoder. Unlike earlier models, GPT could predict the next sequence, applying a general language understanding to specific tasks with smaller annotated datasets.

GPT employed positional encodings to maintain word order in its predictions. Its self-attention mechanism prevented tokens from attending to future tokens, ensuring each word prediction depended only on preceding words.
Hence a decoder-only architecture represents a so-called causal language model, one that generates each item in a sequence based on the previous items. This approach is also referred to as ``autoregressive'', meaning that each new word is predicted based on the previously generated words, with no influence from future words. The generation of each subsequent output is causally linked to the history of generated outputs and nothing ahead of the current word affects its generation.

\subsubsection{Encoder-decoder Models}

Evolving further, BART (Bidirectional and Auto-Regressive Transformers) was introduced by \citeauthor{lewis_bart_2019} in 2019.\cite{lewis_bart_2019} BART combined the context learning strengths of the bidirectional BERT, and the autoregressive capabilities of models like GPT, which excel at generating coherent text. BART was thus a hybrid seq2seq model, consisting of a BERT-like bidirectional encoder and a GPT-like autoregressive decoder. This is nearly the same architecture as \citet{Vaswani2017-kq}; the differences are in the pretraining. BART was pretrained using a task that corrupted text by, for example, deleting tokens, and shuffling sentences. It then learned to reconstruct the original text with left-to-right autoregressive decoding as in GPT models.

\subsubsection{Multi-task and Multi-modal Models}

In previous sections, we discussed LLMs that take natural language text as input and then output either a learned representation or another text sequence. These models traditionally perform tasks like translation, summarization, and classification. However, multi-task models are capable of performing several different tasks using the same model, even if those tasks are unrelated. This allows a single model to be trained on multiple objectives, enhancing its versatility and efficiency, as it can generalize across various tasks during inference.

Multi-task models, such as the Text-to-Text Transfer Transformer (T5) developed by \citet{Raffel2019-uy} demonstrate that various tasks can be reframed into a text-to-text format, allowing the same model architecture and training procedure to be applied universally. By doing so, the model can be used for diverse tasks, but all with the same set of weights. This reduces the need for task-specific models and increases the model's adaptability to new problems. The relevance of this approach is particularly significant as it enables researchers to tackle multiple tasks without needing to retrain separate models, saving both computational resources and time.
For instance, Flan-T5\cite{Chung2022-yp} used instruction fine-tuning with chain-of-thought prompts, enabling it to generalize to unseen tasks, such as generating rationales before answering. This fine-tuning expands the model's ability to tackle more complex problems. More advanced approaches have since been proposed to build robust multi-task models that can flexibly switch between tasks at inference time.\cite{Tan2023-gt, Shen2024-dh, Son2024-nm, Feng2024-iv}

Additionally, LLMs have been extended to process different input modalities, such as image and sound, even though they initially only processed text. For example, Fuyu\cite{FUYU8B-2023} uses linear projection to adapt image representations into the token space of an LLM, allowing a decoder-only model to generate captions for figures. Expanding on this, next-GPT\cite{Wu2023-pw} was developed as an ``any-to-any'' model, capable of processing multiple modalities, such as text, audio, image, and video, through modality-specific encoders. The encoded representation is projected into a decoder-only token space, and the LLM's output is processed by a domain-specific diffusion model to generate each modality's output. Multitask or multimodel methods are further described below as these methods start to connect LLMs with autonomous agents.

\begin{figure}
    \centering
    \includegraphics[width=\textwidth]{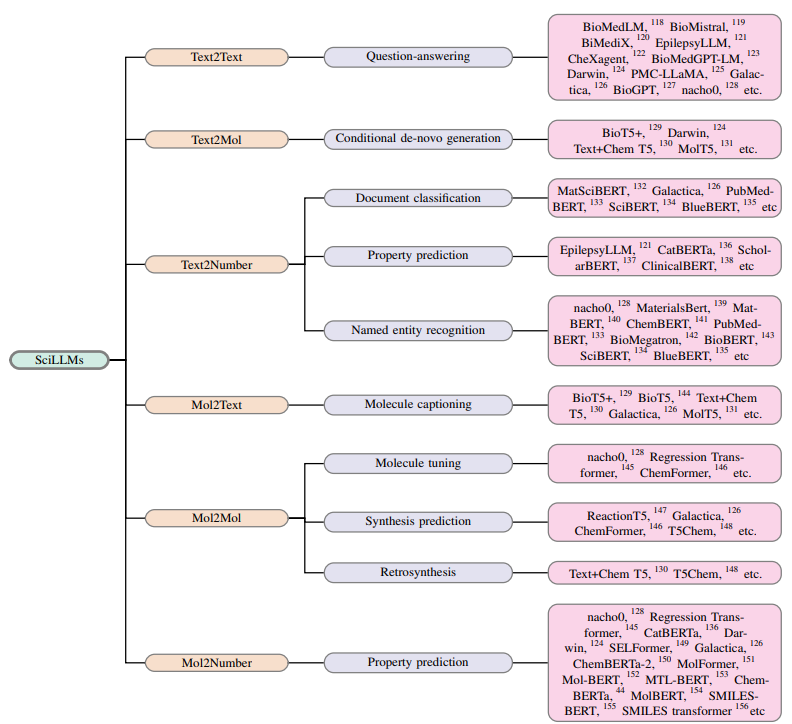}
    \caption{Classification of LLMs in chemistry and biochemistry according to their application. }
    \label{fig:applications}
\end{figure}

\section{LLMs for Chemistry and Biochemistry}\label{sec:LLM_in_science}

The integration of large language models (LLMs) into chemistry and biochemistry is opening new frontiers in molecular design, property prediction, and synthesis. As these models evolve, they increasingly align with specific chemical tasks, capitalizing on the strengths of their architectures. Specifically, encoder-only models excel at property prediction\cite{Li2021-xw}, decoder-only models are suited for inverse design\cite{Bhattacharya2024-ik}, and encoder-decoder models are applied to synthesis prediction\cite{Vaskevicius2023-mk}. 
However, with the development improvement of decoder-only models\cite{Radford2022-wq} and the suggestion that regression tasks can be reformulated as a text completion task\cite{Born2023-nc}, decoder-only models started being also applied for property prediction.\cite{Mao2023-bo, shoghi_molecules_2024, Jablonka2023-bm, Jacobs2024-dv}
This section surveys key LLMs that interpret chemical languages like SMILES and InChI, as well as those that process natural language descriptions relevant to chemistry.

\begin{figure}
    \centering
    \includegraphics[height=0.95\textheight, angle=0]{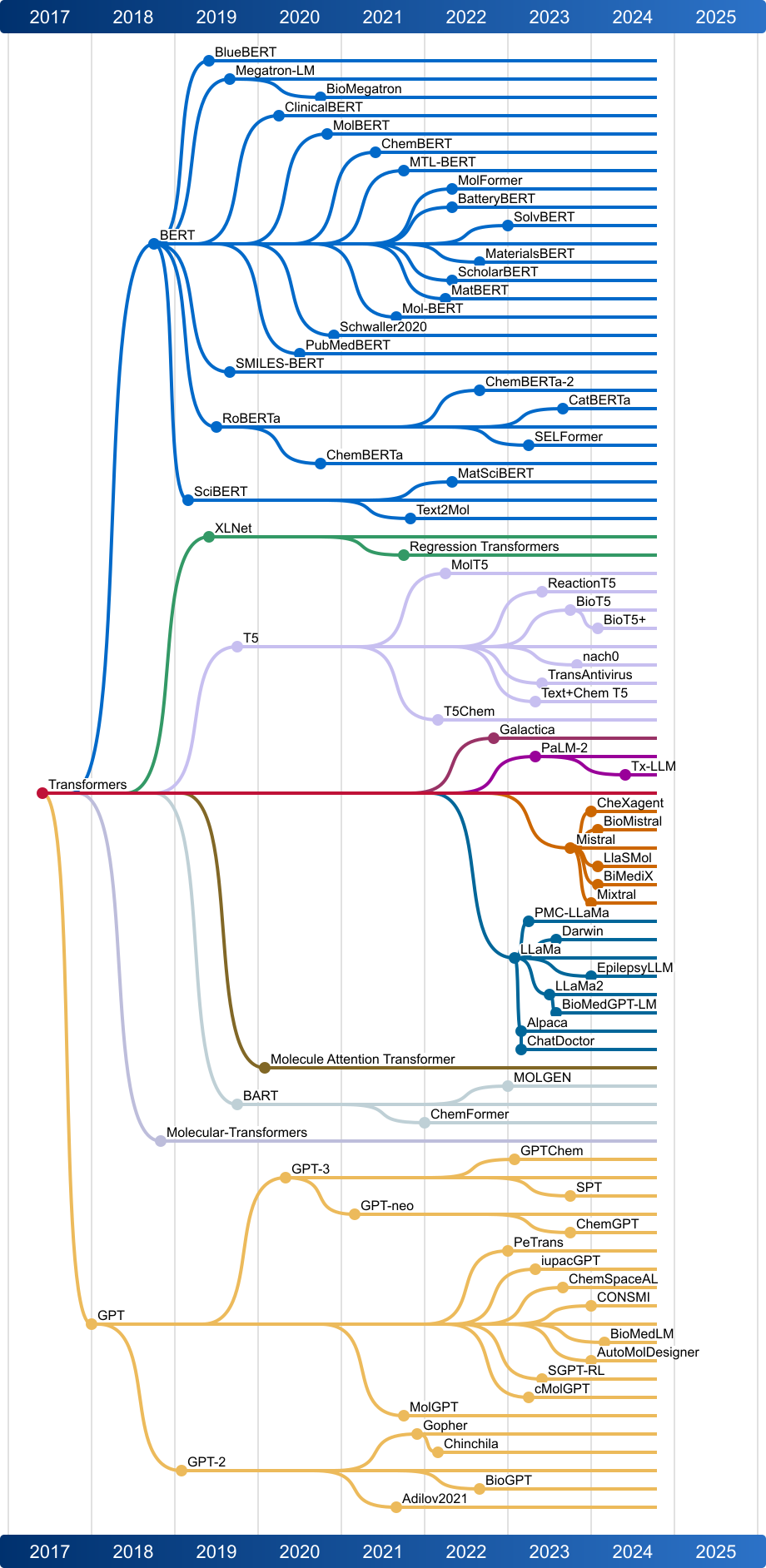}
    \caption{Illustration of how Large Language Models (LLMs) evolved chronologically. The dates display the first publication of each model.}
    \label{fig:sci_llms}
\end{figure}

We provide a chronological perspective on the evolution of LLMs in this field (Figure \ref{fig:sci_llms}), presenting broadly on the design, functionality, and value of each model. Our approach primarily centers on models that use chemical representations like SMILES strings as inputs, but we also examine how natural language models extract valuable data from scientific literature to enhance chemical research.

Ultimately, this discussion underscores the potential for mol-based and text-based LLMs to work together, addressing the growing opportunity for automation in chemistry. This sets the stage for a broader application of autonomous agents in scientific discovery. Figure \ref{fig:applications} illustrates the capabilities of different LLMs available currently, while Figure \ref{fig:sci_llms} presents a chronological map of LLM development in chemistry and biology.

Of critical importance, this section starts by emphasizing the role of trustworthy datasets and robust benchmarks. Without well-curated, diverse datasets, models may fail to generalize across real-world applications. Benchmarks that are too narrowly focused can limit the model's applicability, preventing a true measure of its potential. While natural language models take up a smaller fraction of this section, these models will be increasingly used to curate these datasets, ensuring data quality becomes a key part of advancing LLM capabilities in chemistry.

\subsection{Molecular Representations, Datasets, and Benchmarks}\label{sec:benchmark}

Molecules can be described in a variety of ways, ranging from two-dimensional structural formulas to more complex three-dimensional models that capture electrostatic potentials. Additionally, molecules can be characterized through properties such as solubility, reactivity, or spectral data from techniques like NMR or mass spectrometry. However, to leverage these descriptions in machine learning, they must be converted into a numerical form that a computer can process. Given the diversity of data in chemistry-based machine learning, multiple methods exist for representing molecules,\cite{lo2018machine, david_molecular_2020, Atz2021-nh, Walters2021-ck, Karthikeyan2022-mr, Li2022-io} highlighting this heterogeneity. Common representations include molecular graphs\cite{chen2019graph, hu2020open, kearnes2016molecular}, 3D point clouds\cite{wang2022point, thomas2018tensor, wang2021seppcnet, ahmadi2024machine}, and quantitative feature descriptors.\cite{singh_molecular_2023, Bran2023-oi, shilpa_recent_2023, Wigh2022-kf, david_molecular_2020} 
In this review, we focus specifically on string-based representations of molecules, given the interest in language models. 
Among the known string representations, we can cite IUPAC names, SMILES,\cite{weininger_smiles_1988} DeepSMILES,\cite{oboyle_deepsmiles_2018} SELFIES,\cite{krenn_self-referencing_2020} and InChI,\cite{heller_inchi_2015} as recently reviewed by \citet{Das2024-ad}

\begin{figure}[h!]
    \centering
    \includegraphics[width=0.6\textwidth]{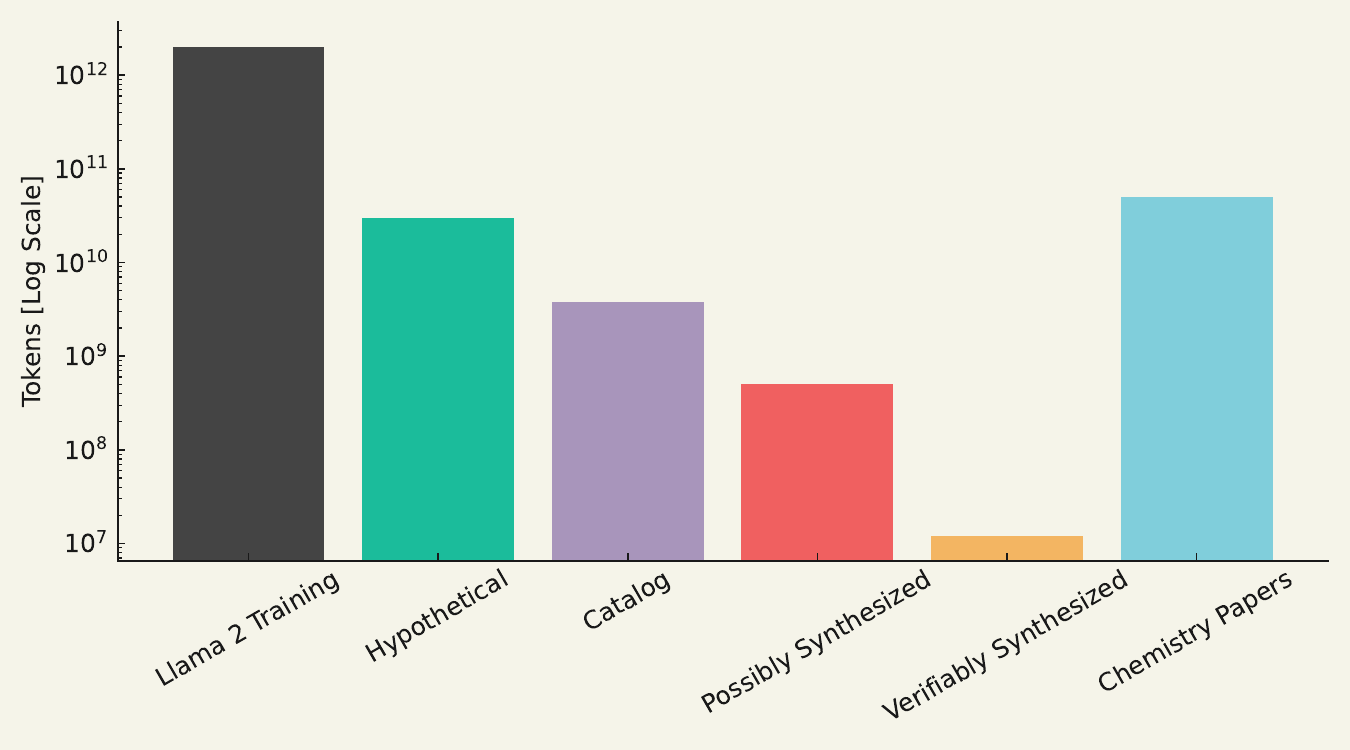}
    \caption{Number of training tokens (on log scale) available from various chemical sources compared with typical LLM training runs. The numbers are drawn from ZINC\cite{irwin2012zinc}, PubChem\cite{kim2016pubchem}, \citet{touvron2023llama}, ChEMBL\cite{Gaulton2012-og}, and \citet{Kinney2023-fj}}.
    \label{fig:tokens}
\end{figure}

Regarding datasets, there are two types of data used for training LLMs, namely training data and evaluation data. Training data should be grounded in real molecular structures to ensure the model develops an accurate representation of what constitutes a valid molecule. This is similar to how natural language training data, such as that used in models like GPT-4, must be based on real sentences or code to avoid generating nonsensical outputs. 
Figure \ref{fig:tokens} shows a comparison of the number of tokens in common chemistry datasets with those used to train LLaMA2, based on literature data.\cite{irwin2012zinc, kim2016pubchem, touvron2023llama, Gaulton2012-og, Kinney2023-fj} With this in mind, we note the largest chemical training corpus, which largely comprises hypothetical chemical structures, amounts to billions of tokens, almost two orders of magnitude fewer than the trillions of tokens used to train LLaMA2. When excluding hypothetical structures from datasets like ZINC,\cite{irwin2012zinc} (Figure \ref{fig:tokens}), the number of tokens associated with verifiably synthesized compounds is over five orders of magnitude lower than that of LLaMA2's training data. To address this gap, efforts such as the Mol-instructions dataset, curated by \citet{Fang2023-oj}, prioritize quality over quantity, providing $\sim$2M biomolecular and protein-related instructions. Mol-instructions\cite{Fang2023-oj} was selectively built from multiple data sources,\cite{Wei2010-rn, Krallinger2015-op, Li2016-wn, wu_moleculenet_2018, Islamaj-Dogan2019-gz, Ashburner2000-xq, Hendrycks2020-ak, Kim2021-tk, Pal2022-xh, Lu2022-ms, UniProt-Consortium2023-vf} with rigorous quality control.
Given the success of literature-based LLMs, one may naturally assume that large datasets are of paramount importance for chemistry. However, it is crucial not to overlook the importance of data quality. \citet{segler2018planning} demonstrated that even using the Reaxys dataset, a very small, human-curated collection of chemical reactions, was sufficient to achieve state-of-the-art results in retrosynthesis. Therefore, the issue is not merely a lack of data, but rather a lack of high-quality data that may be the pivotal factor holding back the development of better scientific LLMs. Ultimately, the focus must shift from sheer quantity to the curation of higher-quality datasets to advance these models.

To evaluate the accuracy of these models, we compare their performance against well-established benchmarks. However, if the benchmarks are not truly representative of the broader chemistry field, it becomes difficult to gauge the expected impact of these models.  Numerous datasets, curated by the scientific community, are available for this benchmarking.\cite{_jablonka_awesome-chemistry-datasets, Mirza2024-qw} Among them, MoleculeNet\cite{wu_moleculenet_2018}, first published in 2017, is the most commonly used labeled dataset for chemistry. However, MoleculeNet has several limitations: it is small, contains errors and inconsistencies, and lacks relevance to a larger number of real-world chemistry problems.\cite{Gloriam2019-yf, irwin_chemformer_2022, liu_multi-modal_2023, Livne2023-bu} Pat Walters, a leader in ML for drug discovery, has emphasized, ``I think the best way to make progress on applications of machine learning to drug discovery is to fund a large public effort that will generate high-quality data and make this data available to the community''\cite{pat_walters_blog3}.

Walters provides several constructive critiques noting, for example, that the QM7, QM8, and QM9 datasets, intended for predicting quantum properties from 3D structures, are often misused with predictions based incorrectly on their 1D SMILES strings, which inadequately represent 3D molecular conformations.
He also suggests more relevant benchmarks and also datasets with more valid entries. For example, he points to the Absorption, Distribution, Metabolism, and Excretion (ADME) data curated by \citet{Fang2023-aj}, as well as the Therapeutic Data Commons (TDC)\cite{TDC-ss, Huang2021-ug} and TDC-2\cite{Velez-Arce2024-xk}. These datasets contain measurements of real compounds, making them grounded in reality. Moreover, ADME is crucial for determining a drug candidate's success, while therapeutic results in diverse modalities align with metrics used in drug development.

Here, we hypothesize that the lack of easily accessible, high-quality data in the correct format for training foundational chemical language models is a major bottleneck to the development of the highly desired ``super-human'' AI-powered digital chemist. A more optimistic view is presented by \citet{inductive_bio_blog}
They argue that we do not need to wait for the creation of new benchmarks. Instead, they suggest that even the currently available, messy public data can be carefully curated to create benchmarks that approximate real-world applications. In addition, we argue that extracting data from scientific chemistry papers might be an interesting commitment to generating data of high quality, grounded to the truth, and on a large scale.\cite{Hira2024-hj}
Some work has been done in using LLMs for data extraction.\cite{Dagdelen2024-le, Circi2024-du}
Recently, a few benchmarks following these ideas were created for evaluating LLMs' performance in biology (LAB-Bench\cite{Laurent2024-dj}) and material science (MatText\cite{Alampara2024-zx}, MatSci-NLP\cite{Song2023-oh} and MaScQA\cite{Zaki2024-mt}).

\subsection{Property Prediction and Encoder-only Mol-LLMs}\label{sec:sciEncoder}

Encoder-only transformer architectures are primarily composed of an encoder, making them well-suited for chemistry tasks that require extracting meaningful information from input sequences, such as classification and property prediction. Since encoder-only architectures are mostly applied to capturing the underlying structure-property relationships, we describe here the relative importance of the property prediction task. \citet{Sultan2024-xt} also discussed the high importance of this task, the knowledge obtained in the last years, and the remaining challenges regarding molecular property prediction using LLMs.

\begin{longtable}{@{\extracolsep{\fill}}m{2.5cm}|m{1.5cm}m{3.8cm}m{1.8cm}m{3cm}m{2cm}}
    
    \caption{Encoder-only scientific LLMs. The release date column displays the date of the first publication for each paper. When available, the publication date of the last updated version is displayed between parentheses.
    $a$:``Model Size'' is reported as the number of parameters.
    $b$: The authors report they not used as many encoder layers as it was used in the original BERT paper. But the total number of parameters was not reported.} \label{tab:enc_llms} \\

    LLM   & Model Size$^a$ & Training Data & Architecture & Application & Release date \\\hline
    \endfirsthead

    \multicolumn{6}{c}{{\bfseries \tablename\ \thetable{} -- continued from previous page}} \\
    LLM   & Model Size$^a$ & Training Data & Architecture & Application & Release date \\\hline
    \endhead

    \hline \multicolumn{6}{r}{{Continued on next page}} \\ \hline
    \endfoot
    
    \hline
    \endlastfoot
    
    CatBERTa\cite{ock_catalyst_2023}
    & 355M & OpenCatalyst2020 (OC20) & RoBERTa & Property prediction & 2023.09 (2023.11) \\
    SELFormer\cite{Yuksel2023-iy}
    & $\sim$86M & $\sim$2M compounds from ChEMBL & RoBERTa & Property prediction & 2023.04 (2023.06) \\
    ChemBERTa-2\cite{ahmad_chemberta-2_2022}
    & 5M - 46M & 77M SMILES from PubChem & RoBERTa & Property prediction & 2022.09\\
    MaterialsBERT\cite{Yoshitake2022-xh} 
    & 110M & 2.4M material science abstracts + 750 annotated abstract for NER & BERT & NER and property extraction & 2022.09 (2023.04) \\
    SolvBERT\cite{Yu2022-fr}
    & \textit{b} & 1M SMILES of solute-solvent pairs from CombiSolv-QM and LogS from \citet{boobier2020machine} & BERT & Property prediction & 2022.07 (2023.01) \\
    ScholarBERT\cite{Hong2022-mv} 
    &340M, 770M& Public.Resource.Org, Inc &BERT& Property prediction & 2022.05 (2023.05) \\
    BatteryBERT\cite{huang_batterybert_2022} 
    & $\sim$ 110M & $\sim$ 400k papers from RSC, Elsevier and Springer &BERT& Document classification & 2022.05 \\
    MatBERT\cite{Trewartha2022-ez} 
    & 110M & Abstracts from solid state articles and abstracts and methods from gold nanoparticle articles & BERT & NER & 2022.04 \\
    MatSciBERT\cite{Gupta2022-hy} 
    & 110M & $\sim$150K material science paper downloaded from Elsevier & BERT & NER and text classification & 2021.09 (2022.05) \\
    Mol-BERT\cite{Li2021-xw}
    & 110M & $\sim$4B SMILES from ZINC15 and ChEMBL27 & BERT & Property prediction & 2021.09 \\
    MolFormer\cite{Ross2022-yp}
    & $b$  & PubChem and ZINC & BERT & Property prediction & 2021.06 (2022.12) \\
    ChemBERT\cite{Guo2022-kr}
    & 110M & $\sim$200k extracted using ChemDataExtractor & BERT & NER & 2021.06 \\
    MolBERT\cite{Fabian2020-zd} 
    & $\sim$85M & ChemBench & BERT & Property prediction & 2020.11\\
    ChemBERTa\cite{Chithrananda2020-cd}
    &  & 10M SMILES from PubChem & RoBERTa & Property prediction & 2020.10\\
    BioMegatron\cite{Shin2020-cp}
    & 345M, 800M, 1.2B &Wikipedia, CC-Stories,  Real-News,  and  OpenWebtext& Megatron-LM & NER and QA & 2020-10 \\
    PubMedBERT\cite{Gu2021-yn} 
    & 110M & 14M abstracts from PubMed & BERT & NER, QA, and document classification & 2020.07 (2021.10) \\
    Molecule Attention Transformer\cite{Maziarka2020-jy}
    & $b$ & ZINC15 & Encoder with GCN features & Property prediction & 2020.02 \\
    SMILES-BERT\cite{Wang2019-nq}
    & $b$ & $\sim$18M SMILES from ZINC & BERT & Property prediction & 2019.09\\
    BlueBERT\cite{Peng2019-yb} 
    & 110M & PubMed and MIMIC-III & BERT & NER, and document classification & 2019.06 \\
    ClinicalBERT\cite{Huang2019-gw} 
    & 110M & MIMIC-III & BERT & Patient readmission probability & 2019.04 \\
    SciBERT\cite{Beltagy2019-ri} 
    & 110M &  1.14M papers from Semantic Scholar & BERT & NER and sentence classification & 2019.03 (2019.11) \\
    BioBERT\cite{Lee2020-vl} 
    & 110M & PubMed and PMC & BERT & NER and QA & 2019.01 (2019.09) \\
\end{longtable}


\subsubsection{Property Prediction}

The universal value of chemistry lies in identifying and understanding the properties of compounds to optimize their practical applications. In the pharmaceutical industry, therapeutic molecules interact with the body in profound ways.\cite{wu_bloodbrain_2023, bissantz2010medicinal, roughley2011medicinal} Understanding these interactions and modifying molecular structures to enhance those therapeutic benefits can lead to significant medical advancements.\cite{doytchinova_drug_2022} Similarly, in polymer science, material properties depend on chemical structure, polymer chain length, and packing,\cite{intro_to_polymer_science} and a protein's function similarly depends on its structure and folding.
Historically, chemists have identified new molecules from natural products\cite{newman_natural_2016} and screened them against potential targets\cite{ferreira_drug_2023} to test their properties for diseases. Once a natural product shows potential, chemists synthesize scaled-up quantities for further testing or derivatization,\cite{kolb_growing_2003, castellino_late-stage_2023, sharma_peptide-based_2023} a costly and labor-intensive process.\cite{reizman_feedback_2016, dimasi_innovation_2016}
Traditionally, chemists have used their expertise to hypothesize the properties of new molecules derived from those natural products, hence aiming for the best investment of synthesis time and labor.
Computational chemistry has evolved to support the chemical industry in more accurate property prediction.\cite{lewars_computational} 
Techniques such as quantum theoretical calculations and force-field-based molecular dynamics offer great support for property prediction and the investigation of molecular systems, though both require substantial computational resources.\cite{bidault_how_2023, pyzer-knapp_accelerating_2021,fredericks_pyxtal_2021, case_convergence_2016, kazantsev_efficient_2011}
Property prediction can now be enhanced through machine learning tools,\cite{huang_application_2023, shilpa_recent_2023, martinez-mayorga_pursuit_2024, Wellawatte2023-uv} and more recent advancements in LLMs lead to effective property prediction without the extensive computational demands of quantum mechanics and MD calculations. Combined with human insight, AI can revolutionize material development, enabling the synthesis of new materials with a high likelihood of possessing desired properties for specific applications.

\subsubsection{Encoder-only Mol-LLMs}

Encoder-only models are exemplified by the BERT architecture, which is commonly applied in natural language sentiment analysis to extract deeper patterns from prose.\cite{deng_llms_2023} The human chemist has been taught to look at a 2D image of a molecular structure and to recognize its chemical properties or classify the compound. Therefore, encoder-only models would ideally convert SMILES strings, empty of inherent chemical essence, into a vector representation, or latent space, which would reflect those chemical properties. This vector representation can then be used directly for various downstream tasks.

While encoder-only LLMs are predominantly used for property prediction, they are also applicable for synthesis classification.\citet{Schwaller2020-xh} used a BERT model to more accurately classify complex synthesis reactions by generating reaction fingerprints from raw SMILES strings, without the need to separate reactants from reagents in the input data, thereby simplifying data preparation. The BERT model achieved higher accuracy (98.2\%) compared to the encoder-decoder model (95.2\%) for classifying reactions. Accurate classification aids in understanding reaction mechanisms, vital for reaction design, optimization, and retrosynthesis. \citet{Toniato2023-in} also used a BERT architecture to classify reaction types for downstream retrosynthesis tasks that would enable the manufacture of any molecular target. Further examples of BERT use include self-supervised reaction atom-to-atom mapping \cite{schwaller_extraction_2021, Schwaller2021-vo}. These chemical classifications would accelerate research and development in organic synthesis, described further below.

Beyond synthesis classification, encoder-only models like BERT have shown great promise for molecular property prediction, especially when labeled data is limited. Recognizing this, \citeauthor{wang_smiles-bert_2019} introduced a semi-supervised SMILES-BERT model, which was pretrained on a large unlabeled dataset with a Masked SMILES Recovery task.\cite{wang_smiles-bert_2019} The model was then fine-tuned for various molecular property prediction tasks, outperforming state-of-the-art methods in 2019 on three chosen datasets varying in size and property. This marked a shift from using BERT for reaction classification towards property prediction and drug discovery. \citet{Maziarka2020-jy} also claimed state-of-the-art performance in property prediction after self-supervised pretraining in their Molecule Attention Transformer (MAT), which adapted BERT to chemical molecules by augmenting the self-attention with inter-atomic distances and molecular graph structure.

\citet{Zhang2022-ar} also tackled the issue of limited property-labeled data and the lack of correlation between any two datasets labeled for different properties, hindering generalizability. They introduced multitask learning BERT (MTL-BERT), which used large-scale pretraining and multitask learning with unlabeled SMILES strings from ChEMBL,\cite{Gaulton2012-og} which is a widely-used database containing bioactive molecules with drug-like properties, designed to aid drug discovery. The MTL-BERT approach mined contextual information and extracted key patterns from complex SMILES strings, improving model interpretability. The model was fine-tuned for relevant downstream tasks, achieving better performance than state-of-the-art methods in 2022 on 60 molecular datasets from ADMETlab\cite{Xiong2021-lf} and MoleculeNet.\cite{wu_moleculenet_2018}

In 2021, \citeauthor{Li2021-xw}\cite{Li2021-xw} introduced Mol-BERT, pretrained on four million unlabeled drug SMILES from the ZINC15\cite{sterling_zinc_2015} and ChEMBL27\cite{Gaulton2012-og} databases to capture molecular substructure information for property prediction. Their work leveraged the underutilized potential of large unlabeled datasets like ZINC, which contains over 230 million commercially available compounds, and is designed for virtual screening and drug discovery. Mol-BERT consisted of three components: a PretrainingExtractor, Pretraining Mol-BERT, and Fine-Tuning Mol-BERT. It treated Morgan fingerprint fragments as ``words'' and complete molecular compounds as ``sentences,'' using RDKit and the Morgan algorithm for canonicalization and substructure identification. This approach generated comprehensive molecular fingerprints from SMILES strings, used in a Masked Language Model (MLM) task for pretraining. Mol-BERT was fine-tuned on labeled samples, providing outputs as binary values or continuous scores for classification or regression, and it outperformed existing sequence and graph-based methods by at least 2\% in ROC-AUC scores on Tox21, SIDER, and ClinTox benchmark datasets.\cite{wu_moleculenet_2018}

\citet{Ross2022-xf} introduced MoLFormer, a large-scale self-supervised BERT model, with the intention to provide molecular property predictions with competitive accuracy and speed when compared to Density Functional Theory calculations or wet-lab experiments. They trained MoLFormer with rotary positional embeddings on SMILES sequences of 1.1 billion unlabeled molecules from ZINC,\cite{sterling_zinc_2015} and PubChem,\cite{kim2016pubchem} another database of chemical properties and activities of millions of small molecules, widely used in drug discovery and chemical research. The rotary positional encoding captures token positions more effectively than traditional methods, \cite{Vaswani2017-kq} improving modeling of sequence relationships. MoLFormer outperformed state-of-the-art GNNs on several classification and regression tasks from ten MoleculeNet\cite{wu_moleculenet_2018} datasets, while performing competitively on two others. It effectively learned spatial relationships between atoms, predicting various molecular properties, including quantum-chemical properties. Additionally, the authors stated how MoLFormer represents an efficient and environment-friendly use of computational resources, claiming a reduced GPU usage in training by a factor of 60 (16 GPUs instead of 1000).

With ChemBERTa, \citet{Chithrananda2020-cd} explored the impact of pretraining dataset size, tokenization strategy, and the use of SMILES or SELFIES, distinguishing their work from other BERT studies. 
They used HuggingFace's RoBERTa transformer,\cite{wolf2020huggingfaces}  and referenced a DeepChem\cite{wu_moleculenet_2018} tutorial for accessibility.
Their results showed improved performance on downstream tasks (BBBP, ClinTox, HIV, Tox21 from MoleculeNet\cite{wu_moleculenet_2018}) as the pretraining dataset size increased from 100K to 10M.
Although ChemBERTa did not surpass state-of-the-art GNN-based baselines like Chemprop (which used 2048-bit Morgan Fingerprints from RDKit),\cite{yang_analyzing_2019} the authors suggested that with expansion to larger datasets they would eventually beat those baselines.
The authors compared Byte-Pair Encoder (BPE) with a custom SmilesTokenizer and its regular expression developed by \cite{Schwaller2019-lc} while exploring tokenization strategies. They found the SmilesTokenizer slightly outperformed BPE, suggesting more relevant sub-word tokenization is beneficial. 
No difference was found between SMILES and SELFIES, but the paper highlighted how attention heads in transformers could be visualized with BertViz,\cite{vig2019bertviz} showing certain neurons selective for functional groups. This study underscored the importance of appropriate benchmarking and addresses the carbon footprint of AI in molecular property prediction.

In ChemBERTa-2, \citet{ahmad_chemberta-2_2022} aimed to create a foundational model applicable across various tasks. They addressed a criticism that LLMs were not so generalizable because the training data was biased or non-representative. They addressed this criticism by training on 77M samples and adding a Multi-Task Regression component to the pretraining. ChemBERTa-2 matched state-of-the-art architectures on MoleculeNet.\cite{wu_moleculenet_2018} As with ChemBERTa, the work was valuable because of additional exploration, in this case into how pretraining improvements affected certain downstream tasks more than others, depending on the type of fine-tuning task, the structural features of the molecules in the fine-tuning task data set, or the size of that fine-tuning dataset. The result was that pretraining the encoder-only model is important, but gains could be made by considering the chemical application itself, and the associated fine-tuning dataset.

In June 2023, \citet{Yuksel2023-iy} introduced SELFormer, building on ideas from ChemBERTa2\cite{ahmad_chemberta-2_2022} and using SELFIES for large data input. 
\citet{Yuksel2023-iy} argue that SMILES strings have validity and robustness issues, hindering effective chemical interpretation of the data, although this perspective is not universally held.
\cite{skinnider_invalid_2024} SELFormer uses SELFIES and is pretrained on two million drug-like compounds, fine-tuned for diverse molecular property prediction tasks (BBBP, SIDER, Tox21, HIV, BACE, FreeSolv, ESOL, PDBbind from MoleculeNet).
\cite{wu_moleculenet_2018} SELFormer outperformed all competing methods for some tasks and produced comparable results for the rest. It could also discriminate molecules with different structural properties. 
The paper suggests future directions in multimodal models combining structural data with other types of molecular information, including text-based annotations. We will discuss such multimodal models below.

In 2022, \citet{Yu2022-fr} published SolvBERT, a multi-task BERT-based regression model that could predict both solvation free energy and solubility from the SMILES notations of solute-solvent complexes. 
It was trained on the CombiSolv-QM dataset,\cite{vermeire2021transfer} a curation of experimental solvent free energy data called CombiSolv-Exp-8780,\cite{Mobley2014-mb, marenich2020minnesota, moine2017estimation, grubbs2010mathematical} and the solubility dataset from \citet{boobier2020machine}. 
SolvBERT's performance was benchmarked against advanced graph-based models\cite{yang2019learned, Rong2020-as}
This work is powerful because there is an expectation that solvation free energy depends on 3-dimensional conformational properties of the molecules, or at least 2D properties that would be well characterized by graph-based molecular representations. It shows an overachieving utility of using SMILES strings in property prediction, and aligns with other work by \citet{winter_smile_2022}, regarding activity coefficients. SolvBERT showed comparable performance to a Directed Message Passing Neural Network (DMPNN) in predicting solvation free energy, largely due to its effective clustering feature in the pretraining phase as shown by TMAP (Tree Map of All Possible) visualizations. Furthermore, SolvBERT outperformed Graph Representation Of Molecular Data with Self-supervision (GROVER)\cite{Rong2020-as} in predicting experimentally evaluated solubility data for new solute-solvent combinations. This underscores the significance of SolvBERT's ability to capture the dynamic and spatial complexities of solvation interactions in a text-based model.

While models like SolvBERT have achieved impressive results in solvation free energy prediction, challenges such as limited labeled data continue to restrict the broader application of transformer models in chemistry. Recognizing this issue, \citeauthor{jiang_intransformer_2024} introduced INTransformer in 2024,\cite{jiang_intransformer_2024} a method designed to enhance property prediction by capturing global molecular information more effectively, even when data is scarce. By incorporating perturbing noise and using contrastive learning to artificially augment smaller datasets, INTransformer delivered improved performance on several tasks. Ongoing work continues to explore various transformer strategies for smaller datasets. Again using contrastive learning, which maximizes the difference between representations of similar and dissimilar data points, but in a different context, MoleculeSTM\cite{Liu2022-sk} uses LLM encoders to create representations for SMILES and for descriptions of molecules extracted from PubChem\cite{Kim2023-nv}.
Similar work was performed by \citet{Xu2023-ws}
The authors curated a dataset with descriptions of proteins.
Subsequently, to train ProtST, a protein language model (PLM) was used to encode amino acid sequences and LLMs to encode the descriptions.

In this section, we outlined the advancements of encoder-only models like BERT and their evolution for property prediction and synthesis classification. Chemists traditionally hypothesize molecular properties, but these models, ranging from Mol-BERT to SolvBERT, showcase the growing efficiency of machine learning in property prediction. Approaches such as multitask learning and contrastive learning, as seen in INTransformer, offer solutions to challenges posed by limited labeled data.

\subsection{Property Directed Inverse Design and Decoder-only mol-LLMs}\label{sec:SciDecoder}

Decoder-only GPT-like architectures offer significant value for property-directed molecule generation and \textit{de novo} chemistry applications because they excel at generating novel molecular structures by learning from vast datasets of chemical compounds. These models can capture intricate patterns and relationships within molecular sequences, proposing viable new compounds that adhere to desired chemical properties and constraints. This enables rapid exploration and innovation within an almost infinite chemical space. Moreover, such large general-purpose models can be fine-tuned with small amounts of domain-specific scientific data,\cite{Jablonka2023-bm, Hocky2024-tc} allowing them to support specific applications efficiently. In this section, we first describe property-directed inverse design from a chemistry perspective and then examine how decoder-only LLMs have propelled inverse design forward. 

\begin{longtable}{@{\extracolsep{\fill}}m{2.5cm}|m{1.5cm}m{3.8cm}m{1.5cm}m{3.8cm}m{2cm}}

    \caption{\ Decoder-only scientific LLMs. The release date column displays the date of the first publication for each paper. When available, the publication date of the last updated version is displayed between parentheses.
    $a$:``Model Size'' is reported as the number of parameters. ``PubMed'' refer to the PubMed abstracts dataset, while PMC (PubMed Corpus) refers to the full-text corpus dataset. $b$: The total number of parameters was not reported.}\label{tab:dec_llms}\\

    LLM   & Model Size$^a$ & Training Data & Architecture & Application & Release date \\\hline
    \endfirsthead

    \multicolumn{6}{c}{{\bfseries \tablename\ \thetable{} -- continued from previous page}} \\
    LLM   & Model Size$^a$ & Training Data & Architecture & Application & Release date \\\hline
    \endhead

    \hline \multicolumn{6}{r}{{Continued on next page}} \\ \hline
    \endfoot
    
    \hline
    \endlastfoot

    Tx-LLM\cite{Chaves2024-xi}
    & $b$ & TDC datasets  & PaLM-2 & Property prediction and retrosynthesis & 2024.06 \\
    BioMedLM\cite{Bolton2024-al}
    & 2.7B & PubMed abstracts and full articles & GPT & QA & 2024.03\\
    LlasMol\cite{yu_llasmol_2024}
    & $\sim$ 7B & SMolInstruct & Galactica, LLaMa, Mistral & Property prediction, molecule captioning, molecule generation, retrosynthesis, name conversion & 2024.02 (2024.08) \\
    BioMistral\cite{Labrak2024-xs} 
    & 7B & PubMed Central (PMC) & Mistral & QA & 2024.02 (2024.08) \\
    BiMediX\cite{Pieri2024-jy}
    & 8x7B & 1.3M Arabic-English instructions (BiMed) & Mixtral & QA & 2024.02\\
    EpilepsyLLM\cite{Zhao2024-hi} 
    & 7B & Data from the Japan Epilepsy Association, Epilepsy Information Center, and Tenkan Net & LLaMa & QA & 2024.01 \\
    CheXagent\cite{Chen2024-lg}
    & 7B & 28 publicly available datasets, including PMC, MIMIC, wikipedia, PadChest, and BIMCV-COVID-19 & Mistral & QA, Image understanding & 2024.01 \\
    ChemSpaceAL\cite{kyro_chemspaceal_2024}
    & $b$ & ChEMBL 33, GuacaMol v1, MOSES, and BindingDB 08-2023 & GPT & Molecule Generation & 2023.09 (2024.02) \\
    BioMedGPT-LM\cite{Luo2023-gw} 
    & 7B and 10B & 5.5M biomedical papers from S2ORC & LLaMA2 & QA & 2023.08 \\
    Darwin\cite{Xie2023-nh} 
    & 7B & SciQ and Web of Science & LLaMA & QA, Property prediction, NER, and Molecule Generation & 2023.08 \\ 
    cMolGPT\cite{Wang2023-jd}
    & $b$ & MOSES & GPT & Molecule Generation & 2023.05 \\
    PMC-LLaMA\cite{Wu2023-tb} 
    & 7B and 13B & MedC-k and MedC-I & LLaMA & QA & 2023.04 (2024.04) \\
    GPTChem\cite{Jablonka2023-bm}
    & 175B & Curation of multiple classification and regression benchmarks & GPT-3 & Property prediction and inverse design & 2023.02 (2024.02) \\
    Galactica\cite{Taylor2022-pu} 
    &125M, 1.3B, 6.7B, 30B, 120B&The galactica corpus, a curation with ~62B scientific documents& Decoder-only & QA, NER, Document Summarization, Property Prediction & 2022.11 \\
    BioGPT\cite{Luo2022-sv}
    & 355M & 15M of Title and abstract from PubMed & GPT-2 & QA, NER, and Document Classification & 2022-09 (2023.04) \\
    SMILES-to-properties-transformer\cite{winter_smile_2022}
    & 6.5M & Synthetic data generated with the thermodynamic  model  COSMO-RS & GPT-3 & Property prediction & 2022.06 (2022.09)\\
    ChemGPT\cite{frey_neural_2023} 
    & $\sim$ 1B & 10M molecules from PubChem & GPT-neo & Molecule generation & 2022.05 (2023.11) \\ 
    Regression Transformer\cite{Born2023-nc}
    & $\sim$27M & ChEMBL, MoleculeNet, USPTO, etc & XLNet & Property prediction, Molecule tuning, Molecule generation & 2022.02 (2023.04) \\
    MolGPT\cite{Bagal2022-tg}
    & 6M & MOSES and GuacaMol & GPT & Molecule Generation & 2021.10 \\
    Adilov2021\cite{Adilov2021-ih}
    & 13.4M & 5M SMILES from ChemBERTa's PubChem-10M. & GPT-2 & Property prediction and molecule generation & 2021.09 \\
\end{longtable}

\subsubsection{Property Directed Inverse Design}

Nature has long been a rich source of molecules that inhibit disease proliferation, because organisms have evolved chemicals for self-defense. Historically, most pharmaceuticals are derived from these natural products,\cite{li_drug_2009, newman_natural_2012} which offer benefits such as cell permeability, target specificity, and a vast chemical diversity.\cite{afarha_strategies_2016}
However, the high costs and complexities associated with high-throughput screening and synthesizing natural products limit the exploration of this space.\cite{afarha_strategies_2016, li_drug_2009}

While natural products have been a valuable starting point, we are not confined to their derivatives. AI, particularly generative LLMs, allows us to go beyond nature and explore a much larger chemical space. \textit{In-silico} molecular design enables rapid modification, akin to random mutation,\cite{nigam_beyond_2021} where only valid, synthesizable molecules that meet predefined property criteria remain in the generated set.\cite{Wellawatte2023-uv, Gandhi2022-mx} This approach allows us to test modifications \textit{in-silico}, expanding exploration beyond the boundaries of natural products.

The true innovation of AI-driven molecular design, however, lies in its ability to directly generate candidate molecules based on desired properties, without the need for iterative stepwise modifications.\cite{Du2024-jf} This ``inverse design'' capability allows us to start with a target property and directly generate candidate molecules that meet the predefined property requirements. Generative LLMs applied to sequences of atoms and functional groups offer a powerful opportunity for out-of-the-box exploration, tapping into the vast chemical space that extends far beyond the confines of nature. This accelerates the path from concept to viable therapeutic agents, aligning seamlessly with decoder-only LLM architectures.

\subsubsection{Decoder-only Mol-LLMs}

One of the first applications of decoder-only models in chemistry was Adilov's (2021) ``Generative pretraining from Molecules''\cite{Adilov2021-ih}. This work pretrained a GPT-2-like causal transformer for self-supervised learning using SMILES strings. By introducing ``adapters'' between attention blocks for task-specific fine-tuning,\cite{houlsby2019parameter} this method provided a versatile approach for both molecule generation and property prediction, requiring minimal architectural changes. It aimed to surpass encoder-only models, such as ChemBERTa,\cite{Chithrananda2020-cd} with a more scalable and resource-efficient approach, demonstrating the power of decoder-only models in chemical generation.

A key advancement then came with MolGPT\cite{Bagal2022-tg}, a 6-million-parameter decoder-only model designed for molecular generation. MolGPT introduced masked self-attention, enabling the learning of long-range dependencies in SMILES strings. The model ensured chemically valid SMILES representations, respecting structural rules like valency and ring closures. It also utilized salience measures for interpretability, aiding in predicting SMILES tokens and understanding which parts of the molecule were most influential in the model's predictions. MolGPT outperformed many existing Variational Auto-Encoder (VAE)-based approaches,\cite{fuhr_deep_22, han_revolutionizing_23, koutroumpa_systematic_23, kell_deep_20, bilodeau_generative_22, gangwal_generative_24, vogt_using_22, talluri_novel_24} in predicting novel molecules with specified properties, being trained on datasets like MOSES\cite{Polykovskiy2020-fv} and GuacaMol.\cite{Brown2019-ul}

While MolGPT's computational demands may be higher than traditional VAEs, its ability to generate high-quality, novel molecules justifies this trade-off. MolGPT demonstrated strong performance on key metrics such as validity, which measures the percentage of generated molecules that are chemically valid according to bonding rules; uniqueness, the proportion of generated molecules that are distinct from one another; Frechet ChemNet Distance (FCD),\cite{preuer2018frechet} which compares the distribution of generated molecules to that of real molecules in the training set, indicating how closely the generated molecules resemble real-world compounds; and KL divergence,\cite{Brown2019-ul} a measure of how the probability distribution of generated molecules deviates from the true distribution of the training data. These metrics illustrate MolGPT's ability to generate high-quality, novel molecules while maintaining a balance between diversity and similarity to known chemical spaces. A brief summary of advancements in transformer-based models for \textit{de-novo} molecule generation from 2023 and 2024 follows, which continue to refine and expand upon the foundational work laid by models like MolGPT.

\citet{haroon_generative_2023} further developed a GPT-based model with relative attention for \textit{de novo} drug design, showing improved validity, uniqueness, and novelty. This work was followed by \citet{frey_neural_2023}, who introduced ChemGPT to explore hyperparameter tuning and dataset scaling in new domains. ChemGPT's contribution lies in refining generative models to better fit specific chemical domains, advancing the understanding of how data scale impacts generative performance. Both \citet{wang_explore_2023} and \citet{mao_transformer-based_2024} presented work that surpassed MolGPT. 
Furthermore, \citet{Mao2023-bo} showed that decoder-only models could generate novel compounds using IUPAC names directly.

This marked a departure from typical SMILES-based molecular representations, as IUPAC names offer a standardized, human-readable format that aligns with how chemists conceptualize molecular structures. By integrating these chemical semantics into the model, iupacGPT\cite{Mao2023-bo} bridges the gap between computational predictions and real-world chemical applications. The IUPAC name outputs are easier to understand, validate, and apply, facilitating smoother integration into workflows like regulatory filings, chemical databases, and drug design. Focusing on pretraining with a vast dataset of IUPAC names and fine-tuning with lightweight networks, iupacGPT excels in molecule generation, classification, and regression tasks, providing an intuitive interface for chemists in both drug discovery and material science.

In a similar vein, \citet{zhang_simple_2023} proposed including target 3D structural information in molecular generative models, even though their approach is not LLM-based. However, it serves as a noteworthy contribution to the field of structure-based drug design. Integrating biological data, such as 3D protein structures, can significantly improve the relevance and specificity of generated molecules, making this method valuable for future LLM-based drug design.
Similarly, \citet{wang_petrans_2023} discussed PETrans, a deep learning method that generates target-specific ligands using protein-specific encoding and transfer learning. This study further emphasizes the importance of using transformer models for generating molecules with high binding affinity to specific protein targets. The significance of these works lies in their demonstration that integrating both human-readable formats (like IUPAC names) and biological context (such as protein structures) into generative models can lead to more relevant, interpretable, and target-specific drug candidates. This reflects a broader trend in AI-driven chemistry to combine multiple data sources for more precise molecular generation, accelerating the drug discovery process.

In 2024, \citet{yoshikai_novel_2024} discussed the limitations of transformer architectures in recognizing chirality from SMILES representations, which impacts the prediction accuracy of molecular properties. To address this, they coupled a transformer with a VAE.
Using contrastive learning from NLP to generate new molecules with multiple SMILES representations, enhancing molecular novelty and validity.\citet{kyro_chemspaceal_2024} presented ChemSpaceAL, an active learning method for protein-specific molecular generation, efficiently identifying molecules with desired characteristics without prior knowledge of inhibitors. \citet{yan_predicting_2024} proposed the GMIA framework, which improves prediction accuracy and interpretability in drug-drug interactions through a graph mutual interaction attention decoder. These innovations represent significant strides in addressing key challenges in molecular generation, such as chirality recognition, molecular novelty, and drug-drug interaction prediction. By integrating new techniques like VAEs, contrastive learning, and active learning into transformer-based models, they have improved both the accuracy and interpretability of molecular design.

Building on these developments, \citet{shen_automoldesigner_2024} reported on AutoMolDesigner, an open-source tool for small-molecule antibiotic design, further emphasizing the role of automation in molecular generation. This work serves as a precursor to more complex models, such as Taiga\cite{Mazuz2023-gb} and cMolGPT,\cite{Wang2023-jd} which employ advanced methods like autoregressive mechanisms and reinforcement learning for molecular generation and property optimization.

For a deeper dive into decoder-only transformer architecture in chemistry, we highlight the May 2023 ``Taiga'' model by \citet{Mazuz2023-gb}, and cMolGPT by \citet{Wang2023-jd}. Taiga first learns to map SMILES strings to a vector space, and then refines that space using a smaller, labeled dataset to generate molecules with targeted attributes. It uses an autoregressive mechanism, predicting each SMILES character in sequence based on the preceding ones. For property optimization, Taiga employs the REINFORCE algorithm,\cite{Zhang2020-ji} which helps refine molecules to enhance specific features. While this reinforcement learning (RL) approach may slightly reduce molecular validity, it significantly improves the practical applicability of the generated compounds. Initially evaluated using the Quantitative Estimate of Drug-likeness (QED) metric,\cite{bickerton2012quantifying} Taiga has also demonstrated promising results in targeting IC50 values,\cite{Gaulton2012-og} the BACE protein,\cite{subramanian2016computational} and anti-cancer activities they collected from a variety of sources. This work underscores the importance of using new models to address applications that require a higher level of chemical sophistication, to illustrate how such models could ultimately be applied outside of the available benchmark datasets. It also builds on the necessary use of standardized datasets and train-validation-test splitting, to demonstrate progress, as explained by \citet{wu_moleculenet_2018}. Yet, even the MoleculeNet benchmarks\cite{wu_moleculenet_2018} are flawed, and we point the reader here to a more detailed discussion on benchmarking,\cite{pat_walters_blog3} given that a significant portion of molecules in the BACE dataset have undefined stereo centers, which, at a deeper level, complicates the modeling and prediction accuracy.

While models like Taiga demonstrate the power of autoregressive learning and reinforcement strategies to generate molecules with optimized properties, the next step in molecular design incorporates deeper chemical domain knowledge. This approach is exemplified by \citet{Wang2023-jd}. They introduced cMolGPT, a conditional generative model that brings a more targeted focus to drug discovery by integrating specific protein-ligand interactions, which underscores the importance of incorporating chemical domain knowledge to effectively navigate the vast landscape of drug-like molecules. Using self-supervised learning and an auto-regressive approach, cMolGPT generates SMILES guided by predefined conditions based on target proteins and binding molecules. Initially trained on the MOSES dataset\cite{Polykovskiy2020-fv} without target information, the model is fine-tuned with embeddings of protein-binder pairs, focusing on generating compound libraries and target-specific molecules for the EGFR, HTR1A, and S1PR1 protein datasets.\cite{xu20084, yu2022structural, xu2021structural, sun2017excape}

Their approach employs a QSAR model\cite{yang_concepts_2019} to predict the activity of generated compounds, achieving a Pearson correlation coefficient over $0.75$. However, despite the strong predictive capabilities, this reliance on a QSAR model, with its own inherent limitations, highlights the need for more extensive experimental datasets. cMolGPT\cite{Wang2023-jd} tends to generate molecules within the sub-chemical space represented in the original dataset, successfully identifying potential binders but struggling to broadly explore the chemical space for novel solutions. This underscores the challenge of generating diverse molecules with varying structural characteristics while maintaining high binding affinity to specific targets. While cMolGPT advances the integration of biological data and fine-tuned embeddings for more precise molecular generation, models like Taiga and cMolGPT differ in their approach. Taiga\cite{Mazuz2023-gb} employs reinforcement learning to optimize generative models for molecule generation, while cMolGPT uses target-specific embeddings to guide the design process. Both highlight the strengths of decoder-only models but emphasize distinct strategies; Taiga optimizes molecular properties through autoregressive learning, and cMolGPT focuses on conditional generation based on protein-ligand interactions.

In contrast, \citet{yu_llasmol_2024} follow a different approach with LlaSMol,\cite{yu_llasmol_2024} which utilizes pretrained models (for instance Galactica, LlaMa2, and Mistral) and performs parameter efficient fine-tuning (PEFT) techniques\cite{Han2024-ix, Ding2023-oa} such as LoRa\cite{Hu2021-hi}. PEFT enables fine-tuning large language models with fewer parameters, making the process more resource-efficient while maintaining high performance. LlaSMol demonstrated its potential by achieving state-of-the-art performance in property prediction tasks, particularly when fine-tuned on benchmark datasets like MoleculeNet.\cite{wu_moleculenet_2018}.

There continue to be significant advancements being made in using transformer-based models to tackle chemical prediction tasks with optimized computational resources, including more generalist models, such as Tx-LLM,\cite{Chaves2024-xi} designed to streamline the complex process of drug discovery. For additional insights on how these models are shaping the field, we refer the reader to several excellent reviews, \cite{guzman-pando_deep_2023, fromer_computer-aided_2023, vogt_exploring_2023-1, Das2024-ad} with \citet{goel_efficient_2023} highlighting the efficiency of modern machine learning methods in sampling drug-like chemical space for virtual screening and molecular design. \citet{goel_efficient_2023} discussed the effectiveness of generative models, including large language models (LLMs), in approximating the vast chemical space, particularly when conditioned on specific properties or receptor structures. 

We provide a segue from this section by introducing the work by \citet{Jablonka2023-bm}, which showcases a decoder-only GPT model that, despite its training on natural language rather than specialized chemical languages, competes effectively with decoder-only LLMs tailored to chemical languages. The authors finetuned GPT-3 to predict properties and conditionally generate molecules and, therefore, highlight its potential as a foundational tool in the field. This work sets the stage for integrating natural language decoder-only LLMs, like GPT, into chemical research, where they could serve as central hubs for knowledge discovery.

Looking ahead, this integration foreshadows future developments that pair LLMs with specialized tools to enhance their capabilities, paving the way for the creation of autonomous agents that leverage deep language understanding in scientific domains. Decoder-only models have already significantly advanced inverse molecular design, from improving property prediction to enabling target-specific molecular generation. Their adaptability to various chemical tasks
demonstrates their value in optimizing drug discovery processes and beyond. As models like LlaSMol and cMolGPT continue to evolve, integrating chemical domain knowledge and biological data, they offer exciting opportunities for more precise molecular generation. The growing potential for combining large language models like GPT-4 with specialized chemical tools signals a future where AI-driven autonomous agents could revolutionize chemical research, making these models indispensable to scientific discovery.

\subsection{Synthesis Prediction and Encoder-decoder Mol-LLMs}\label{sec:sciEncDec}

The encoder-decoder architecture is designed for tasks involving the translation of one sequence into another, making it ideal for predicting chemical reaction outcomes or generating synthesis pathways from given reactants. We begin with a background on optimal synthesis prediction and describe how earlier machine learning has approached this challenge. Following that, we explain how LLMs have enhanced chemical synthesis prediction and optimization. Although, our context below is aptly chosen to be synthesis prediction, other applications exist. For example, SMILES Transformer (ST)\cite{Honda2019-ek} is worth a mention, historically, because it explored the benefits of self-supervised pretraining to produce continuous, data-driven molecular fingerprints from large SMILES-based datasets.

\begin{table*}[ht]
    \small
    \caption{\ Encoder-decoder scientific LLMs. The release date column displays the date of the first publication for each paper. When available, the publication date of the last updated version is displayed between parentheses.
    $a$:``Model Size'' is reported as the number of parameters. $b$: The total number of parameters was not reported.}
    \label{tab:enc_dec_llms}
    \begin{tabular*}{\textwidth}{@{\extracolsep{\fill}}m{2cm}|m{1.5cm}m{4.5cm}m{1.5cm}m{3.5cm}m{2cm}}
        LLM   & Model Size$^a$ & Training Data & Architecture & Application & Release date\\\hline
        BioT5+\cite{Pei2024-ph} 
        & 252M &ZINC20, UniRef50, 33M PubMed articles, 339K mol-text pairs from PubChem, 569K FASTA-text pairs from Swiss-prot& T5 & Molecule Captioning, Molecule Generation, Property Prediction,  & 2024.02 (2024.08) \\
        nach0\cite{Livne2023-bu}
        & 250M & MoleculeNet, USPTO, ZINC & T5 & Property prediction, Molecule generation, Question answering, NER & 2023.11 (2024.05) \\
        ReactionT5\cite{Sagawa2023-fa}
        & 220M & ZINC and ORD & T5 & Property prediction and Reaction prediction & 2023.11 \\
        BioT5\cite{Pei2023-bp} 
        & 252M &ZINC20, UniRef50, full-articles from BioRxiv and PubMed, mol-text-IUPAC information from PubChem & T5 & Molecule Captioning, Property Prediction & 2023-10 (2024.12) \\
        MOLGEN\cite{Fang2023-ty}
        & $b$ & ZINC15 & BART & Molecule Generation & 2023.01 (2024.03) \\
        Text+Chem T5\cite{Christofidellis2023-ea}
        & 60M, 220M & 11.5M or 33.5M samples curated from \citet{Vaucher2020-gz}, \citet{Toniato2023-in}, and CheBI-20 & T5 & Molecule Captioning, Product Prediction, Retrosynthesis, Molecule Generation & 2023.01 (2023.06) \\
        MolT5\cite{Edwards2022-yu} 
        & 60M, 770M & C4 dataset & T5 & Molecule Captioning and Molecule Generation & 2022.04 (2022.12) \\
        T5Chem\cite{Lu2022-ms}
        & 220M & USPTO & T5 & Product Prediction, Retrosynthesis, Property Prediction & 2022.03 \\
        Text2Mol\cite{edwards-etal-2021-text2mol}
        & $b$ & CheBI-20 & SciBERT w/ decoder & Molecule captioning and conditional molecule generation &  2021.11 \\
        ChemFormer\cite{irwin_chemformer_2022}
        & 45M, 230M & 100M SMILES from ZINC-15 & BART & Product Prediction, Property Prediction, Molecular Generation & 2021.07 (2022.01) \\
        SMILES transformer\cite{Honda2019-ek}
        & $b$ & ChEMBL24 & Transformer & Property prediction & 2019.11 \\
        Molecular Transformer\cite{Schwaller2019-lc} 
        & 12M & USPTO & Transformer & Product prediction & 2018.11 (2019.08) \\
    \end{tabular*}
\end{table*}

\subsubsection{Synthesis Prediction}

Once a molecule has been identified through property-directed inverse design, the next challenge is to predict its optimal synthesis, including yield. \citet{shenvi-2024-natural} describe how the demanding and elegant syntheses of natural products has contributed greatly to organic chemistry. However, in the past 20 years, the focus has shifted away from complex natural product synthesis towards developing new reactions applicable for a broader range of compounds, especially in reaction catalysis.\cite{shenvi-2024-natural} Yet, complex synthesis is becoming relevant again as it can be digitally encoded, mined by LLMs,\cite{ai_extracting_2024} and applied to new challenges. Unlike property prediction, reaction prediction is particularly challenging due to the involvement of multiple molecules. Modifying one reactant requires adjusting all others, with different synthesis mechanisms or conditions likely involved. Higher-level challenges exist for catalytic reactions and complex natural product synthesis.
Synthesis can be approached in two ways. Forward synthesis involves building complex target molecules from simple, readily available substances, planning the steps progressively. Retrosynthesis, introduced by E.J. Corey in 1988,\cite{corey_robert_1988} is more common. It involves working backward from the target molecule, breaking it into smaller fragments whose re-connection is most effective. Chemists choose small, inexpensive, and readily available starting materials to achieve the greatest yield and cost-effectiveness. As a broad illustration, the first total synthesis of discodermolide\cite{nerenberg_total_1993} involved 36 such steps, a 24-step longest linear sequence, and a 3.2\% yield. There are many possible combinations for the total synthesis of the target molecule, and the synthetic chemist must choose the most sensible approach based on their expertise and knowledge. However, this approach to total synthesis takes many years.
LLMs can now transform synthesis such that structure-activity relationship predictions can be coupled in lock-step with molecule selection based on easier synthetic routes. This third challenge of predicting the optimal synthesis can also lead to the creation of innovative, non-natural compounds, chosen because of such an easier predicted synthesis but for which the properties are still predicted to meet the needs of the application. Thus, these three challenges introduced above are interconnected.

\subsubsection{Encoder-decoder mol-LLMs}
Before we focus on transformer use, some description is provided on the evolution from RNN and Gated Recurrent Unit (GRU) approaches in concert with the move from template-based to semi-template-based to template-free models. \citet{nam_linking_2016} pioneered forward synthesis prediction using a GRU-based translation model. In contrast, \citet{liu_retrosynthetic_2017} reported retro-synthesis prediction with a Long Short-Term Memory (LSTM) based seq2seq model incorporating an attention mechanism, achieving 37.4\% accuracy on the USPTO-50K dataset. The reported accuracies of these early models highlighted the challenges of synthesis prediction, particularly retrosynthesis. \citet{schneider-2016-fingerprint} further advanced retrosynthesis by assigning reaction roles to reagents and reactants based on the product.

Building on RNNs and GRUs, the field advanced with the introduction of template-based models. In parallel with the development of the Chematica tool \cite{august_2016_software_nodate, klucznik_efficient_2018} for synthesis mapping, \citet{Segler2017-dq} highlighted that traditional rule-based systems often failed by neglecting molecular context, leading to "reactivity conflicts." Their approach emphasized transformation rules that capture atomic and bond changes, applied in reverse for retrosynthesis. Trained on 3.5 million reactions, their model achieved 95\% top-10 accuracy in retrosynthesis and 97\% for reaction prediction on a validation set of nearly 1 million reactions from the Reaxys database (1771–2015). Although not transformer-based, this work laid the foundation for large language models (LLMs) in synthesis. However, template-based models depend on explicit reaction templates from known reactions, limiting their ability to predict novel reactions and requiring manual updates to incorporate new data.

Semi-template-based models offered a balance between rigid template-based methods and flexible template-free approaches. They used interpolation or extrapolation within template-defined spaces to predict a wider range of reactions and to adjust based on new data. In 2021, \citet{somnath_learning_2021} introduced a graph-based approach recognizing that precursor molecule topology is largely unchanged during reactions. Their model broke the product molecule into ``synthons'' and added relevant leaving groups, making results more interpretable.\cite{Wellawatte2023-rw} Training on the USPTO-50k dataset, \cite{schneider-2016-fingerprint} they achieved a top-1 accuracy of 53.7\%, outperforming previous methods.

However, the template-free approaches align well with transformer-based learning approaches because they learn retrosynthetic rules from raw training data. This provides significant flexibility and generalizability across various types of chemistry.  Template-free models are not constrained by template libraries and so can uncover novel synthetic routes that are undocumented or not obvious from existing reaction templates.
To pave the way for transformer use in synthesis, \citet{cadeddu_organic_2014} drew an analogy between fragments in a compound and words in a sentence due to their similar rank distributions. \citet{schwaller_found_2018} further advanced this with an LSTM network augmented by an attention-mechanism-based encoder-decoder architecture, using the USPTO dataset.\cite{schneider-2016-fingerprint} They introduced a new ``regular expression'' (or regex) for tokenizing molecules, framing synthesis (or retrosynthesis) predictions as translation problems with a data-driven, template-free sequence-to-sequence model. They tracked which starting materials were actual reactants, distinguishing them from other reagents like solvents or catalysts, and used the regular expression to uniquely tokenize recurring reagents, as their atoms were not mapped to products in the core reaction. This regex for tokenizing molecules is commonly used today in all mol-based LLMs.

In 2019, going beyond the ``neural machine'' work of \citet{nam_linking_2016}, \citet{Schwaller2019-lc} first applied a transformer for synthesis prediction, framing the task as translating reactants and reagents into the final product. Their model inferred correlations between chemical motifs in reactants, reagents, and products in the dataset (USPTO-MIT,\cite{jin_predicting_2017} USPTO-LEF, \cite{bradshaw_generative_2019} USPTO-STEREO\cite{schwaller_found_2018}). It required no handcrafted rules and accurately predicted subtle chemical transformations, outperforming all prior algorithms on a common benchmark dataset. The model handled inputs without a reactant-reagent split, following their previous work,\cite{schwaller_found_2018} and accounted for stereochemistry, making it valuable for universal application.
Then, in 2020, for automated retrosynthesis, \citet{schwaller_predicting_2020} developed an advanced Molecular Transformer model with a hyper-graph exploration strategy. The model set a standard for predicting reactants and other entities, evaluated using four new metrics. ``Coverage'' measured how comprehensively the model could predict across the chemical space, while ``class diversity'' assessed the variety of chemical types the model could generate, ensuring it was not limited to narrow subsets of reactions. ``Round-trip accuracy'' checked whether the retrosynthetically predicted reactants could regenerate the original products, ensuring consistency in both directions. ``Jensen–Shannon divergence'' compared the predicted outcomes to actual real-world distributions, indicating how closely the model’s predictions matched reality. Constructed dynamically, the hypergraph allowed for efficient expansion based on Bayesian-like probability scores, showing high performance despite training data limitations. Notably, accuracy improved when the re-synthesis of the target product from the generated precursors was factored in, a concept also employed by \citet{chen_deep_2021} and \citet{westerlund_chemformers_2024}. Also in 2020, \citet{zheng_predicting_2020} developed a ``template-free self-corrected retrosynthesis predictor'' (SCROP) using transformer networks and a neural network-based syntax corrector, achieving 59.0\% accuracy on a benchmark dataset.\cite{lowe_extraction_2012, schneider-2016-fingerprint} This approach outperformed other deep learning methods by over 2\% and template-based methods by over 6\%.

We now highlight advancements in synthesis prediction using the BART Encoder-Decoder architecture, starting with Chemformer by \citet{irwin_chemformer_2022}. This paper emphasized the computational expense of training transformers on SMILES and the importance of pretraining for efficiency. It showed that models pretrained on task-specific datasets or using only the encoder stack were limited for sequence-to-sequence tasks. After transfer learning, Chemformer achieved state-of-the-art results in both sequence-to-sequence synthesis tasks and discriminative tasks, such as optimizing molecular structures for specific properties. They studied the effects of small changes on molecular properties using pairs of molecules from the ChEMBL database\cite{Gaulton2012-og} with a single structural modification. Chemformer’s performance was tested on the ESOL, Lipophilicity, and Free Solvation datasets.\cite{wu_moleculenet_2018} \citet{irwin_chemformer_2022} also described their use of an in-house property prediction model, but when models train on calculated data for ease of access and uniformity, they abstract away from real-world chemical properties. We again emphasize the importance of incorporating experimentally derived data into Chemistry LLM research to create more robust and relevant models. Continuously curating new, relevant datasets that better represent real-world chemical complexities will enhance the applicability and transferability of these models.

In 2023, \citet{Toniato2023-in} also applied LLMs to single-step retrosynthesis as a translation problem, but increased retrosynthesis prediction diversity by adding classification tokens, or ``prompt tokens,'' to the target molecule's language representation, guiding the model towards different disconnection strategies. Increased prediction diversity has high value by providing out-of-the-box synthetic strategies to complement the human chemist’s work.
To measure retrosynthesis accuracy, \citet{li_retro-bleu_2024} introduced Retro-BLEU, a metric adapted from the BLEU (Bilingual Evaluation Understudy) score used in machine translation.\cite{Papineni2002-le} Despite progress in computer-assisted synthesis planning (CASP), not all generated routes are chemically feasible due to steps like protection and deprotection needed for product formation. Widely accepted NLP metrics like BLEU\cite{Papineni2002-le} and ROUGE\cite{Lin2004-md} focus on precision and recall by computing n-gram overlaps between generated and reference texts. Similarly, in retrosynthesis, reactant-product pairs can be treated as overlapping bigrams. Retro-BLEU uses a modified BLEU score, emphasizing precision over recall, as there is no absolute best route for retrosynthesis. Although not yet applied to LLM-based predictions, this approach has value by allowing future performance comparison with a single standard.

Finally, by expanding the use of encoder-decoder architectures outside synthesis prediction into molecular generation, \citet{Fang2023-ty} introduced MOLGEN, a BART-based pretrained molecular language model, in a 2023 preprint updated in 2024. MOLGEN addressed three key challenges: generating valid SMILES strings, avoiding an observed bias that existed against natural product-like molecules, and preventing hallucinations of molecules that didn't retain the intended properties. Pretrained on 100 million molecules using SELFIES\cite{krenn_self-referencing_2020} and a masked language model approach, MOLGEN predicts missing tokens to internalize chemical grammar. An additional highlight of this work is how MOLGEN uses ``domain-agnostic molecular prefix tuning.'' This technique integrates domain knowledge directly into the model's attention mechanisms by adding molecule-specific prefixes, trained simultaneously with the main model across various molecular domains. The model's parameters would thus be adjusted to better capture the complexities and diversities of molecular structures, and domain-specific insights would be seamlessly integrated. To prevent molecular hallucinations, MOLGEN employs a chemical feedback mechanism, to autonomously evaluate generated molecules for appropriate properties, to guide learning and optimization. Such feedback foreshadows a core aspect of autonomous agents, which is their capacity for reflection. We will explore this further below.

The advancements in synthesis prediction and molecular generation using encoder-decoder architectures have revolutionized the field, moving from rigid, template-based models to more flexible, template-free approaches. Early work with LSTMs and GRUs laid the foundation, while transformer-based models like Molecular Transformer and Chemformer set new benchmarks in accuracy and versatility. New metrics, such as Retro-BLEU, and domain-aware techniques, like MOLGEN’s prefix tuning, have further refined predictions and molecular design. These innovations, coupled with self-correcting mechanisms, point to a future of autonomous molecular design, where AI agents can predict, evaluate, and optimize synthetic pathways and molecular properties, accelerating chemical discovery.

\subsection{Multi-Modal LLMs}

We have demonstrated the impact of LLMs on chemistry through their ability to process textual representations of molecules and reactions. However, LLMs can also handle diverse input modalities, representing molecular and chemical data in various formats.\cite{David2020-zu, Hart2021-kl, Eraso2023-jc}
In chemistry, data can be represented in various forms, each providing unique insights and information (see Section~\ref{sec:benchmark}).
Chemical representations can be broadly classified into 1D, 2D, and 3D categories, depending on how much structural detail they convey.\cite{Karthikeyan2022-mr, Li2022-io}
1D representations include basic numerical descriptors, such as molecular features and fingerprints, as well as textual representations like SMILES\cite{weininger_smiles_1988}, SELFIES\cite{krenn_self-referencing_2020}, and IUPAC names. These descriptors vary in the amount of chemical information they carry.\cite{Lo2018-mi}
2D representations involve graph-based structures and visual formats, which can be extended with geometric information to produce 3D representations. 
Examples of 3D representations include molecular graphs enriched with spatial data, molecular point clouds, molecular grids, and 3D geometry files.\cite{Flam-Shepherd2023-sv} 

Some of these representations can be input into models in different ways. For instance, a point cloud can be expressed either as a vector of coordinates (numerical input) or as a text-based PDB file.
However, due to the distinct nature of the information conveyed, we treat textual descriptions of different molecular representations as separate modalities, even though both are technically strings.
Additionally, molecule images have been utilized to train transformer-based models.\cite{Rajan2021-tz} However, spectral data—such as Nuclear Magnetic Resonance (NMR), Infrared (IR) spectroscopy, and mass spectrometry, remain underexplored as inputs for LLM-based applications.

Multi-modal LLMs leverage and integrate these diverse data types to enhance their predictive and analytical capabilities. This integration improves the accuracy of molecular property predictions and facilitates the generation of novel compounds with desired properties.
A key example is Text2Mol proposed by \citet{edwards-etal-2021-text2mol} in 2021, which integrates natural language descriptions with molecular representations, addressing the cross-lingual challenges of retrieving molecules using text queries. The researchers created a paired dataset linking molecules with corresponding text descriptions and developed a unified semantic embedding space to facilitate efficient retrieval across both modalities. This was further enhanced with a cross-modal attention-based model for explainability and reranking. One stated aim was to improve retrieval metrics, which would further advance the ability for machines to learn from chemical literature.

In their 2022 follow-up, MolT5, \citet{Edwards2022-yu} expanded on earlier work by utilizing both SMILES string representations and textual descriptions to address two tasks: generating molecular captions from SMILES and predicting molecular structures from textual descriptions of desired properties. However, several key challenges remain. Molecules can be described from various perspectives, such as their therapeutic effects, applications (e.g., aspirin for pain relief or heart attack prevention), chemical structure (an ester and a carboxylic acid connected to a benzene ring in ortho geometry), or degradation pathways (e.g., breaking down into salicylic acid and ethanoic acid in moisture).\cite{carstensen_decomposition_1988} This complexity demands expertise across different chemistry domains, unlike typical image captioning tasks involving everyday objects (e.g., cats and dogs), which require minimal specialized knowledge. Consequently, building large, high-quality datasets pairing chemical representations with textual descriptions is a challenging task.

Moreover, standard metrics like BLEU, effective in traditional NLP, are insufficient for evaluating molecule-text tasks. To address these challenges, \citet{Edwards2022-yu} employed a denoising objective, training the model to reconstruct corrupted input data, thereby learning the structure of both text and molecules. Fine-tuning on gold-standard annotations further improved the model’s performance, enhancing previous Text2Mol metrics\cite{edwards-etal-2021-text2mol} and enabling MolT5 to generate accurate molecular structures and their corresponding captions.

Other multimodal approaches similarly target the fusion of chemical and linguistic data to advance applications in molecular design. \citet{seidl_enhancing_2023} developed CLAMP, which combines separate chemical and language modules to predict biochemical activity, while \citet{xu_multilingual_2023} 
presented BioTranslator, a tool that translates text descriptions into non-text biological data to explore novel cell types,  protein function, and drug targets. These examples highlight the growing trend of using language-based interfaces to enhance molecular exploration. The potential of multimodal LLMs extends beyond chemistry into more interactive and accessible tools. ChatDrug, by \citet{liu_chatgpt-powered_2023}, integrates multimodal capabilities through a prompt module, a retrieval and domain feedback module, and a conversation module for systematic drug editing. 
It identifies and manipulates molecular structures for better interpretability in pharmaceutical research.
Similarly, \citet{Christofidellis2023-ea} introduced a multi-domain, multi-task language model capable of handling tasks across both chemical and natural language domains without requiring task-specific pretraining.describe Joint Multi-domain Pre-training (JMP), which operates on the hypothesis that pre-training across diverse chemical domains enhances generalization toward a more robust foundational model. In this context, \citet{liu_molxpt_2023} developed MolXPT, introduced MolXPT, which further demonstrated the power of multimodal learning by achieving robust zero-shot molecular generation.

Finally, models that integrate even more diverse data types, such as GIT-Mol,\cite{Liu_2024_gitmol} which combines graphs, images, and text, and MolTC\cite{fang_moltc_2024}, which integrates graphical information for molecular interaction predictions illustrate how multimodal data improves accuracy and generalizability. Moreover, multimodal fusion models like PremuNet\cite{zhang_pre-trained_2024} and 3M-Diffusion, \citet{zhu_3m-diffusion_2024} which use molecular graphs and natural language for molecule generation, represent a significant leap forward in the creation of novel compounds. \citet{gao_dockingga_2024} advanced targeted molecule generation with DockingGA, combining transformer neural networks with genetic algorithms and docking simulations for optimal molecule generation, utilizing Self-referencing Chemical Structure Strings to represent and optimize molecules. \citet{zhou_instruction_2024} developed TSMMG, a teacher-student LLM designed for multi-constraint molecular generation, leveraging a large set of text-molecule pairs to generate molecules that satisfy complex property requirements. \citet{gong_text-guided_2024}introduced TGM-DLM, a diffusion model for text-guided molecule generation that overcomes limitations of autoregressive models in generating precise molecules from textual descriptions. These advances culminate in works like
MULTIMODAL-MOLFORMER by \citet{soares_beyond_2023},  which integrates chemical language and physicochemical features with molecular embeddings from MOLFORMER,\cite{Ross2022-yp}, significantly enhancing prediction accuracy for complex tasks like biodegradability and PFAS toxicity.

Overall, the shift to multimodal LLMs represents a robust approach to molecular design. By integrating diverse data sources, these models significantly enhance accuracy, interpretability, and scalability, opening new avenues for drug discovery, material design, and molecular property prediction. Combining linguistic, chemical, and graphical data into unified frameworks enables AI-driven models to make more informed predictions and generate innovative molecular structures.

\subsection{Textual Scientific LLMs}

LLMs are large neural networks known for their performance across various machine learning tasks, with the main advantage of not requiring well-structured data like molecular descriptors.\cite{Riedl2023-fc}
Their true power lies in their ability to handle more challenging tasks, such as extracting insights from less structured data sources like scientific texts or natural language descriptions. In chemistry, this opens doors to new methods of data extraction, classification, and generation, although it depends heavily on the availability of high-quality and diverse datasets (as discussed in Section~\ref{sec:benchmark}.
Unfortunately, many datasets are locked behind paywalls or are not machine-readable, limiting the full potential of LLMs in scientific applications. Encouraging open data initiatives and standardization of formats will play a vital role in expanding LLM applications in chemistry and related fields.

\subsubsection{Text Classification}
One of the key uses of LLMs in science is text classification, where models sift through vast amounts of scientific literature to extract structured data. For example, \citet{Huang2019-gw} applied LLMs to predict patient readmission using clinical data from MIMIC-III.\cite{mimic-iii} ClinicalBERT\cite{Huang2019-gw} used a combination of masked language modeling and next-sentence prediction, followed by fine-tuning on the readmission prediction task. Similarly, \citet{Zhao2024-hi} developed EpilepsyLLM by fine-tuning LLaMA using epilepsy data, demonstrating how instruction-based fine-tuning enables models to specialize in highly specific fields.
In another application, SciBERT\cite{Beltagy2019-ri} and ScholarBERT\cite{Hong2022-mv} adapted BERT to handle scientific literature. SciBERT, developed by \citet{Beltagy2019-ri} utilized a specialized tokenizer built for scientific texts from Semantic Scholar,\cite{Kinney2023-fj} and demonstrated superior performance over fine-tuned BERT models\cite{Devlin2018-nl} on scientific tasks. This improvement highlighted the importance of tailored vocabularies in model performance.\citet{Hong2022-mv} later developed ScholarBERT by pretraining on scientific articles from \texttt{Public.Resource.Org} and using RoBERTa optimizations\cite{Liu2019-ur} to improve pretraining performance. ScholarBERT was further fine-tuned on the tasks used for evaluation.
Despite using a larger dataset, ScholarBERT did not outperform LLMs trained on narrower domain datasets. However, ScholarBERT performed well on specific tasks, such as named entity recognition (NER) within the ScienceExamCER dataset,\cite{Smith2019-vu} which involved 3rd to 9th grade science exam questions.

\citet{Guo2022-kr} argue that manually curating structured datasets is a sub-optimal, time-consuming, and labor-intensive task. 
Therefore, they automated data extraction and annotation from scientific papers using \texttt{ChemDataExtractor}\cite{Swain2016-se} and their in-house annotation tool\cite{MTurk}. Text extraction tasks, like NER, can be formulated as multi-label classification tasks, which motivates using NER-like approaches and LLMs to extract structured data directly from unstructured text. LLMs developed for data mining include the work of \citet{Zhang2024-ab} and \citet{Chen2024-pi}

Text extraction tasks, like NER, can be formulated as multi-label classification tasks, which motivates using NER-like approaches and LLMs to extract structured data directly from unstructured text.
LLMs developed for data mining include the work of \citet{Zhang2024-ab} and \citet{Chen2024-pi}. Building upon this, \citet{Wang2024-ds} conducted a study comparing GPT-4 and ChemDataExtractor\cite{Swain2016-se} for extracting band gap information from materials science literature. They found that GPT-4 achieved a higher level of accuracy (Correctness 87.95\% vs. 51.08\%) without the need for training data, demonstrating the potential of generative LLMs in domain-specific information extraction tasks. Additionally, LLMs with support for image inputs have been shown to enable accurate data extraction directly from images of tables.\cite{Circi2024-du} A detailed discussion can be found in the study by \citet{Schilling-Wilhelmi2024-cx}.

In contrast to broad domain models, some LLMs focus on narrow, specialized fields to improve performance. ChemBERT\cite{Guo2022-kr} was pretrained using a BERT model to encode chemical reaction information, followed by fine-tuning a NER head.
ChemBERT outperformed other models such as BERT\cite{Devlin2018-nl} and BioBERT\cite{Lee2020-vl} in the product extraction task, presenting an improvement of $\sim$~6\% in precision. For product role labeling, that is by identifying the role an extracted compound plays in a reaction, ChemBERT showed a $\sim$~5\% improvement in precision. This suggests that training on narrower datasets enables models to learn specific patterns in the data more effectively.

This trend continued with MatSciBERT,\cite{Gupta2022-hy} and MaterialsBERT.\cite{Shetty2023-ew} With MatSciBERT, \citet{Gupta2022-hy} fine-tuned SciBERT\cite{Beltagy2019-ri} on the Material Science Corpus (MSC), a curated dataset of materials extracted from Elsevier’s scientific papers and improved article subject classification accuracy by 3\% compared to SciBERT. In a similar vein, with MaterialsBERT, \citet{Shetty2023-ew} fine-tuned PubMedBERT\cite{Gu2021-yn} on 2.4 million abstracts, showing incremental precision improvements in NER tasks. BatteryBERT\cite{huang_batterybert_2022} also followed this strategy, outperforming baseline BERT models in battery-related tasks.

Considerable effort has also been devoted to developing LLMs for biology tasks, following a similar trend of training models on large corpora such as Wikipedia, scientific databases, and textbooks, and then fine-tuning them for specific downstream tasks.
\citet{Shin2020-cp} pretrained various sizes of Megatron-LM\cite{Shoeybi2019-hr}, another BERT-like LLM, to create the BioMegatron family of models. These models, which had 345M, 800M, and 1.2B parameters and vocabularies of either 30k or 50k tokens, were pretrained using abstracts from the PubMed dataset and full-text scientific articles from PubMed Central (PMC), similar to BioBERT.\cite{Lee2020-vl}

Surprisingly, the largest 1.2B model did not perform better than the smaller ones, with the 345M parameter model using the 50k tokens vocabulary consistently outperforming others in tasks like Named Entity Recognition (NER) and Relation Extraction (RE). NER identifies specific entities, such as chemicals or diseases, while RE determines the relationships between them—both crucial for structuring knowledge from unstructured data. These processes streamline research by converting raw textual information into structured, usable formats for further analysis. This suggests that, for certain tasks, increasing model size does not necessarily lead to better performance. The relevance of model size was more apparent in the SQuAD \cite{Rajpurkar2016-og} dataset, suggesting that LLMs trained on smaller, domain-specific datasets may face limitations in broader generalization.

BioBERT\cite{Lee2020-vl} pretrained using data from Wikipedia, textbooks, PubMed abstracts, and the PMC full-text corpus, outperformed the original BERT in all tested benchmarks, and in some cases even achieved state-of-the-art (SOTA) performance in benchmarks such as NCBI disease, 2010 i2b2/VA, BC5CDR, BC4CHEMD, BC2GM, JNLPBA, LINNAEUS, and Species-800.
\citet{Peng2020-bp} developed BlueBERT, a multi-task BERT model, which was evaluated on the Biomedical Language Understanding Evaluation (BLUE) benchmark.\cite{Peng2019-yb} BlueBERT was pretrained on PubMed abstracts and MIMIC-III,\cite{mimic-iii} and fine-tuned on various BLUE tasks, showing performance similar to BioBERT across multiple benchmarks.

PubMedBERT\cite{Gu2021-yn}, following the approach adopted in SciBERT, created a domain-specific vocabulary using 14M abstracts from PubChem for pretraining. In addition to pretraining, the team curated and grouped biomedical datasets to develop BLURB, a comprehensive benchmark for biomedical natural language processing (NLP) tasks, including NER, sentence similarity, document classification, and question-answering. \citet{Gu2021-yn} demonstrated that PubMedBERT significantly outperformed other LLMs in the BLURB benchmark, particularly in the PubMedQA and BioQSA datasets. The second-best model in these datasets was BioBERT, emphasizing the importance of domain-specific training for high-performance LLMs in biomedical applications.

Text classification using LLMs, particularly in biomedicine and materials science, has demonstrated that domain-specific pretraining is most effective for enhancing model performance. Models like BioBERT, BlueBERT, and PubMedBERT highlight how focusing on specialized datasets, such as PubMed and MIMIC-III, improves accuracy in tasks like NER, RE, and document classification. These advances illustrate how narrowing the training scope to relevant data enables more effective extraction of structured information from unstructured scientific texts.

In the broader context of this work, text classification serves as a key element that allows AI models to interface with chemical, biological, and medical literature, thereby accelerating progress in drug design, materials discovery, and other research fields. This ability to classify and extract relevant information from scientific texts directly impacts the efficiency and precision of data interpretation, facilitating real-world applications across multiple domains.

\subsubsection{Text Generation}

Text generation in scientific LLMs offers unique capabilities beyond simply encoding and retrieving information. Unlike encoder-only models, which focus primarily on extracting insights from structured data, decoder models introduce generative abilities that allow them to create new text, answer questions, and classify documents with generated labels. This capability is particularly valuable in scientific fields, where LLMs must not only interpret data but also generate coherent and contextually accurate outputs based on domain-specific instructions. The following models demonstrate how decoder-based architectures enhance generative tasks in natural science, biology, and medical applications.

The Darwin model, as outlined by \citet{Xie2023-nh}, is one such example. It fine-tunes LLaMA-7B on FAIR, a general QA dataset, followed by specific scientific QA datasets. Instructions for scientific QA were sourced from SciQ\cite{Welbl2017-jv} and generated using the Scientific Instruction Generation (SIG) model, a tool fine-tuned from Vicuna-7B that converts full-text scientific papers into question-answer pairs. This multi-step training process significantly improved Darwin's performance on regression and classification benchmarks. Notably, LLaMA-7B fine-tuned only on FAIR achieved nearly the same results as the fully fine-tuned model on six out of nine benchmarks, indicating that the integration of domain-specific datasets may not always require extensive fine-tuning for performance gains.

Similarly, \citet{Song2023-et} created HoneyBee by fine-tuning LLaMA-7B and LLaMa-13B on MatSci-Instruct, a dataset with $\sim$52k instructions curated by the authors. HoneyBee outperformed other models, including MatBERT, MatSciBERT, GPT, LLaMa, and Claude, within its specialized dataset. However, \citet{Zhang2024-ks} showed that HoneyBee did not generalize well to other benchmarks, such as MaScQA\cite{Zaki2024-mt} and ScQA\cite{Auer2023-wq}, highlighting the limitations of models trained on narrow domains in terms of broader applicability.

In biology, BioGPT\cite{Luo2022-sv} pretrained a GPT-2 model architecture using 15M abstracts from PubChem corpus.
BioGPT was evaluated across four tasks and five benchmarks, including end-to-end relation extraction on BC5CDR, KD-DTI, and DDI, question-answering on PubMedQA, document classification on HoC, and text generation on all these benchmarks. After fine-tuning on these tasks (excluding text generation), BioGPT consistently outperformed encoder-only models like BioBERT and PubMedBERT, particularly in relation extraction and document classification. Focusing specifically on text generation, the authors compared BioGPT's outputs to those of GPT-2, concluding that BioGPT was superior, although no quantitative metric was provided for this comparison.

Building on these ideas, \citet{Wu2023-tb} pretrained LLaMA2 with the MedC-k dataset, which included 4.8M academic papers and 30K textbooks. This model was further refined through instruction tuning using the MedC-I dataset, a collection of medical QA problems. PMC-LLaMA\cite{Wu2023-tb} outperformed both LLaMa-2 and ChatGPT on multiple biomedical QA benchmarks, even though it was $\sim$~10 times smaller in size. Notably, the model's performance on MedQA\cite{Jin2020-he}, MedMCQA\cite{Pal2022-te}, and PubMedQA\cite{Jin2019-uw} benchmarks improved progressively as additional knowledge was incorporated, the model size increased, and more specific instructions were introduced during tuning.

Text generation through decoder models has significantly expanded the applications of LLMs in scientific fields by enabling the generation of contextual answers and labels from scientific data. Unlike encoder-only models that rely on predefined classifications, decoder models such as Darwin, HoneyBee, and BioGPT can produce outputs tailored to domain-specific needs. This capability is important in fields like biomedicine, where accurate question-answering and document generation are highly valued. By leveraging multi-step pretraining and fine-tuning on specialized datasets, decoder models offer greater flexibility in handling both general and domain-specific tasks.

In the broader context of this work, text generation marks a key methodological advance that complements other LLM tasks, such as classification and extraction. The ability to generate structured responses and create new text from scientific literature accelerates research and discovery across chemistry, biology, and medicine. This generative capacity bridges the gap between raw data and meaningful scientific insights, equipping AI-driven models with a more comprehensive toolkit for addressing complex research challenges.

\subsection{The use of ChatGPT in Chemistry}

With the rise of ChatGPT, we review here how many researchers have wanted to test the capability of such an accessible decoder-only LLM. \citet{castro_nascimento_large_2023} wrote the first notable paper on ChatGPT's impact on Chemistry. The authors emphasize that LLMs, trained on extensive, uncurated datasets potentially containing errors or secondary sources, may include inaccuracies limiting their ability to predict chemical properties or trends. The paper highlighted that while LLMs could generate seemingly valid responses, they lacked true reasoning or comprehension abilities and would perpetuate existing errors from their training data. However, the authors suggested that these limitations could be addressed in the future.
The work serves as a benchmark to qualitatively assess improvements in generative pretrained transformers. For example, five tasks were given to ChatGPT (GPT-3). The accuracy for converting compound names to SMILES representations and vice versa was about 27\%, with issues in differentiating alkanes and alkenes, benzene and cyclohexene, or \textit{cis} and \textit{trans} isomers. ChatGPT found reasonable octanol-water partition coefficients with a 31\% mean relative error, and a 58\% hit rate for coordination compounds' structural information. It had a 100\% hit rate for polymer water solubility and a 60\% hit rate for molecular point groups. Understandably, the best accuracies were achieved with widely recognized topics. The authors concluded that neither experimental nor computational chemists should fear the development of LLMs or task automation; instead, they advocated for enhancing AI tools tailored to specific problems and integrating them into research as valuable facilitators.

The use of ChatGPT in chemistry remains somewhat limited. Studies by \citet{humphry_potential_2023}, \citet{emenike_was_2023}, and \citet{fergus_evaluating_2023} focus on its role in chemical education. Some research also explores ChatGPT's application in specific areas, such as the synthesis and functional optimization of Metal-Organic Frameworks (MOFs), where computational modeling is integrated with empirical chemistry research.\cite{zheng_shaping_2023, zheng_chatgpt_2023, xie_fine-tuning_2024, zheng_gpt-4_2023} 
\citet{deb2024-uc} offer a detailed yet subjective evaluation of ChatGPT’s capabilities in computational materials science. They demonstrate how ChatGPT assisted with tasks like identifying crystal space groups, generating simulation inputs, refining analyses, and finding relevant resources. Notably, the authors emphasize ChatGPT’s potential to write code that optimizes processes and its usefulness for non-experts, particularly in catalyst development for CO$_2$ capture.

Three key points emerge regarding the use of ChatGPT alone. First, reliable outputs depend on precise and detailed input, as \citet{deb2024-uc} found when ChatGPT struggled to predict or mine crystal structures. Second, standardized methods for reproducing and evaluating GPT-based work remain underdeveloped. Third, achieving complex reasoning likely requires additional chemical tools or agents, aligning with Bloom’s Taxonomy.\cite{bloom1968taxonomy,bloom2010taxonomy}
Bloom’s Taxonomy organizes educational objectives into hierarchical levels: Remembering, Understanding, Applying, Analyzing, Evaluating, and Creating. These range from recalling facts to constructing new concepts from diverse elements. While LLMs and autonomous agents can support lower-level tasks, they currently fall short of replicating higher-order cognitive skills comparable to human expertise.

Currently, LLMs and autonomous agents are limited in replicating higher-level thinking compared to human understanding.
To better assess LLMs' capabilities in this domain, we propose using Bloom’s Taxonomy as a quality metric.\cite{bloom1968taxonomy,bloom2010taxonomy} This framework offers a structured approach for evaluating the sophistication of LLMs and autonomous agents, especially when addressing complex chemical challenges. It can help quantify their ability to engage in higher-level reasoning and problem-solving.

\subsubsection{Automation}\label{sec:automation}

The evolution of artificial intelligence in chemistry has fueled the potential for automating scientific processes. For example, in 2019, \citet{coley_robotic_2019} developed a flow-based synthesis robot proposing synthetic routes and assembling flow reaction systems, tested on medically relevant molecules, and in 2020, \citet{gromski_universal_2020} provided a useful exploration of how chemical robots could outperform humans when executing chemical reactions and analyses. They developed the Chemputer, a programmable batch synthesis robot handling reactions like peptide synthesis and Suzuki coupling. In 2021, \citet{grisoni_combining_2021} combined deep learning-based molecular generation with on-chip synthesis and testing. The Automated Chemical Design (ACD) framework by \citet{goldman_defining_2022} provides a useful taxonomy for automation and experimental integration levels.
Thus, automation promises to enhance productivity through increased efficiency, error reduction, and the ability to handle complex problems, as described in several excellent reviews regarding automation in chemistry,\cite{schneider_automating_2018, janet_artificial_2023, coley_autonomous_2020, coley_autonomous_2020-1,thakkar_artificial_2021, shen_automation_2021, liu_microfluidics_2021} 

This increased productivity may be the only possible approach to exploring the vastness of all chemical space. 
To fully leverage AI in property prediction, inverse design, and synthesis prediction, it must be integrated with automated synthesis, purification, and testing. This automation should be high-throughput and driven by AI-based autonomous decision-making (sometimes called ``lights-out'' automation). \citet{janet_artificial_2023} highlighted challenges in multi-step reactions with intermediate purifications, quantifying uncertainty, and the need for standardized recipe formats. They also stated the limitations of automated decision-making.
Organa\cite{Darvish2024-cp} addresses some of these challenges. It can significantly reduce physical workload and improve users' lab experience by automating diverse common lab routine tasks such as solubility assessment, pH measurement, and recrystallization. Organa interacts with the user through text and audio. The commands are converted into a detailed LLM prompt and used to map the goal to the robot's instructions. Interestingly, Organa is also capable of reasoning over the instructions, giving feedback about the experiments, and producing a written report with the results.

Other limitations exist, like a machine being restricted to pre-defined instructions, its inability to originate new materials, and the lower likelihood of lucky discoveries. Yet, when dedicated tools can be connected to address each step of an automated chemical design, these limitations can be systematically addressed through advancements in LLMs and autonomous agents, discussed in the next section.

\section{LLM-based Autonomous Agents}\label{sec:agents}

The term ``agent'' originates in philosophy, referring to entities capable of making decisions\cite{Salamon2011-lz}. Hence, in artificial intelligence, an ``agent'' is a system that can perceive its environment, make decisions, and act upon them in response to external stimuli\cite{Xi2023-ee}. Language has enabled humans to decide and act to make progress in response to the environment and its stimuli, and so LLMs are naturally ideal for serving as the core of autonomous agents. Thus, in agreement with \citet{Gao2024-zp}, we define a ``language agent'' as a model or program (typically based on LLMs) that receives an observation from its environment and executes an action in this environment. Here, environment means a set of tools and a task.
Hence, ``LLM-based autonomous agents'' refer to language agents whose core is based on an LLM model.
Comprehensive analyses of these agents are available in the literature\cite{Wang2023-kc, Xi2023-ee, Gao2024-zp}, but this section highlights key aspects to prepare the reader for future discussions.

There is no agreed definition of the nomenclature to be used to discuss agents.
For instance, \citet{Gao2024-zp} created a classification scheme that aims to group agents by their autonomy in biological research. This means a level 0 agent has no autonomy and can only be used as a tool, while a level 3 agent can independently create hypotheses, design experiments, and reason.

\begin{figure}[h!]
    \centering
    \includegraphics[width=0.99\textwidth]{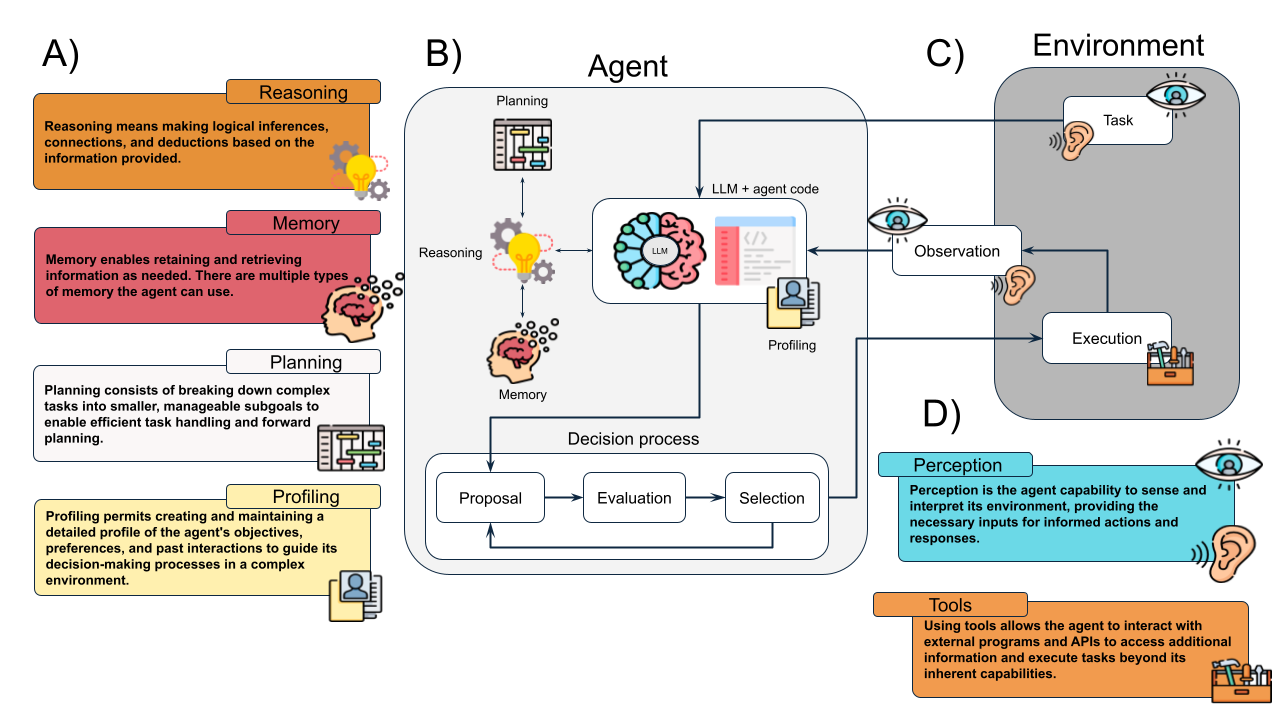}
    \caption{
    Agent's architecture as defined in this review. According to our definition, an agent is composed of a central program (typically an LLM and the code to implement the agent's dynamic behavior) and the agent modules. The agent continuously receives observations from the environment and decides which action should be executed to complete the task given to it. Here, we define the agent as the set of elements whose decision is trainable, that is, the LLM, the agent code, the decision process, and the agent modules. Given a task, the agent uses the agent modules (memory, reasoning, planning, profiling) and the LLM to decide which action should be executed. This action is executed by calling a tool from the environment. After the action is executed, an observation is produced and fed back to the agent. The agent can use perception to receive inputs in different modalities from the environment. A) Description of agent modules, B) illustration of the agent architecture, C) illustration of the environment components, D) description of tools elements present in the environment.
    }
    \label{fig:agents}
\end{figure}

Following this perspective, \citet{Wang2023-kc} categorizes agent components into four modules: profiling, memory, planning, and action. In contrast, \citet{Weng2023-ic} also identifies four elements — memory, planning, action, and tools — but with a different emphasis. Meanwhile, \citet{Xi2023-ee} proposes a division into three components: brain, perception, and action, integrating profiling, memory, and planning within the brain component, where the brain is typically an LLM. 
Recently, \citet{Sumers2023-ct} proposed Cognitive Architectures for Language Agents (CoALA), a conceptual framework to generalize and ease the design of general-purpose cognitive language agents. In their framework, a larger cognitive architecture composed of modules and processes is defined. CoALA defines a memory, decision-making, and core processing module, in addition to an action space composed of both internal and external tools. While internal tools mainly interact with the memory to support decision-making, external tools make up the environment, as illustrated in Figure~\ref{fig:agents}.
Given a task that initiates the environment, the ``decision process'' runs continuously in a loop, receiving observations and executing actions until the task is completed. For more details, read \citet{Sumers2023-ct}.

In this review, we define an autonomous agent system as a model (typically an LLM) that continuously receives observations from the environment and executes actions to complete a provided task, as described by \citet{Gao2024-zp}. Nevertheless, in contrast to CoALA\cite{Gao2024-zp}, we will rename ``internal tools'' as ``agent modules'' and ``external tools'' simply as ``tools'', for clarity. The agent consists of trainable decision-making components such as the LLM itself, policy, memory, and reasoning scheme. In contrast, the environment comprises non-trainable elements like the task to be completed, Application Programming Interface (API) access, interfaces with self-driving labs, dataset access, and execution of external code. 
By referring to decision-making components as agent modules, we emphasize their inclusion as parts of the agent. By referring to non-trainable elements as tools, we highlight their role as part of the environment. We discuss six main types of actions. As shown in Figure~\ref{fig:agents}, four of the six, memory, planning, reasoning, and profiling are agent modules. The remaining two actions (or tools) and perception are part of the environment. Since the perception is how the agent interacts with the environment and is not a trainable decision, we therefore included it as part of the environment.

\subsection{Memory Module}

The role of the memory module is to store and recall information from past interactions and experiences to inform future decisions and actions. There are multiple types of memory in agents, namely sensory memory, short-term memory, and long-term memory. 
A major challenge in using agents is the limited context window, which restricts the amount of in-context information and can lead to information loss, thereby impacting the effectiveness of short-term and long-term memory. Solutions involve summarizing memory content\cite{Liang2023-in}, compressing memories into vectors\cite{Zhang2023-yw, Zhu2023-tx, Zhong2023-pa}, and utilizing vector databases\cite{Han2023-yf} or combinations thereof\cite{Zhao2023-ra}, with various databases available such as ChromaDB, FAISS, Pinecone, Weaviate, Annoy, and ScaNN.\cite{ANN-benchmark-eo} Addressing these challenges to enhance agent memory continues to be a significant area of research\cite{Hatalis2023-fh}.
Sensory, or procedural memory is knowledge embedded into the model's parameters during pretraining and/or in heuristics implemented into the agent's code. 
Short-term, or working, memory includes the agent's finite knowledge during a task, incorporating interaction history and techniques like in-context learning\cite{Brown2020-ej} (ICL), which leverages the limited input's context length for information retention. 
Long-term memory involves storing information externally, typically through an embedded vector representation in an external database. In the original CoALA\cite{Gao2024-zp} paper, long-term memory is further categorized as episodic, which registers previous experiences, and semantic, which stores general information about the world.

\subsection{Planning and Reasoning Modules}
The planning and reasoning module is made of two components. Planning involves identifying a sequence of actions required to achieve a specified goal. In the context of language agents, this means generating steps or strategies that the model can follow to solve a problem or answer a question, which can be enhanced with retrieval from previous experiences,\cite{Park2023-fr} and from feedback from post-execution reasoning\cite{Raman2022-zo, Dhuliawala2023-wp}. We note that Retrieval-Augmented Generation (RAG) enhances the planning phase by enabling models to access external knowledge bases, integrating retrieved information into the generation process. This approach improves accuracy and relevance, especially when handling complex or knowledge-intensive tasks.
Reasoning refers to the process of drawing conclusions or making decisions based on available information and logical steps. For example, there are studies that demonstrate the benefits of LLM reasoning for question answering, where new context tokens can be integrated in a step-by-step way to guide the model towards more accurate answers\cite{Huang2022-rs, Kojima2022-bd, Wei2022-rm, Wang2022-lq, Yao2022-db, Hao2023-ng}. One popular reasoning strategy is Chain-of-Thought (CoT),\cite{Wei2022-rm,Liu2023-zb, Yao2023-ph, Shinn2023-kt, Kang2023-ol, Qian2024-pi} a reasoning strategy which substantially boosts QA performance by generating intermediate reasoning steps in a sequential manner. CoT involves breaking down complex problems into smaller, manageable steps, allowing the model to work through reasoning one step at a time rather than attempting to solve the entire problem at once. CoT thereby reduces hallucinations and enhances interpretability, as demonstrated by improved results in models like PaLM\cite{Anil2023-ja} and GPT-3 with benchmarks like GSM8K\cite{Cobbe2021-ef}, SVAMPs\cite{Jie2022-fy}, and MAWPS\cite{Lan2021-as}.

In advanced reasoning, final tasks are often decomposed into intermediary ones using a cascading approach, similar to Zero-shot-CoT\cite{Kojima2022-bd} and RePrompt\cite{Raman2022-zo}. 
However, while CoT is considered as single-path reasoning, CoT extensions like Tree-of-Thoughts\cite{Yao2022-db}, Graph-of-Thoughts\cite{Besta2023-sd}, Self-consistent CoT\cite{Wang2022-lq}, and Algorithm-of-Thoughts\cite{Sel2023-vr} offer multi-path reasoning. Furthermore, other models have pitted multiple agents against each other to debate or discuss various reasoning paths,\cite{Liang2023-mz, Du2023-wj, Chan2023-qf} while others use external planners to create plans\cite{Song2022-uw, Liu2023-cp}. 
A feedback step during the execution of the plan was a further extension of the CoT ideas; this enables agents to refine their actions based on environmental responses adaptively, which is crucial for complex tasks.\cite{Madaan2023-kd, Xi2023-sj}

Another interesting reasoning scheme is the Chain-of-Verification(CoVe), \cite{Dhuliawala2023-wp} where once an answer is generated, another LLM is prompted to generate a set of verification questions to check for agreement between the original answer and the answers to the verification questions such that the final answer can be refined.
The ReAct\cite{Yao2022-db} -- Reason+Act -- model proposes adding an observation step after acting. This means the LLM first reasons about the task and determines the necessary step for its execution, it performs the action and then observes the action's result. Reasoning on that result, it can subsequently perform the following step. 
Similarly, Reflexion\cite{Shinn2023-kt} also implements a reasoning step after executing an action. However, Reflexion implements an evaluator and self-reflection LLMs to not only reason about each step but also to evaluate the current trajectory the agent is following using a long-term memory module.
As the context increases, it may become challenging for agents to deal with the long prompt. Aiming to solve this issue, the Chain-of-Agents (CoA)\cite{Zhang2024-xw} extends reasoning schemes that leverage multi-agent collaboration to reason over long contexts. This framework employs workers and manager agents to process and synthesize information to generate the final response. CoA demonstrated improvements of up to 10\% when compared against an RAG baseline.

ReAct and Reflexion are closed-ended approaches where the agent starts with all the tools and must determine which to use. To address more open-world challenges, \citet{Wang2023-ni} introduced the Describe, Explain, Plan, and Select (DEPS) method, which extends this approach. Lastly, human inputs can also be used to provide feedback to the agent. Providing feedback using a human-in-the-loop approach is particularly interesting in fields where safety is a main concern.

\subsection{Profiling Module}
LLMs can be configured to perform in specific roles, such as coders, professors, students, and domain experts, through a process known as profiling. Language agents can thus incorporate the profile through the LLM or through the agent code. The profiling approach involves inputting psychological characteristics to the agent, significantly impacting its decision-making process \citep{Deshpande2023-iz, Hong2023-ma, Li2023-uj, Jinxin2023-wp}. Profiling enables the creation of multi-agent systems that simulate societal interactions, with each agent embodying a unique persona within the group \citep{Park2023-fr, Qian2023-sd}. The most prevalent technique for profiling, called ``handcrafting'', requires manually defining the agent's profile, often through prompts or system messages \citep{Shao2023-ic, Ke2023-mp}. While profiling can also be automated with LLMs \citep{Wang2023-jq}, that automation method may only be suited for generating large numbers of agents since it offers less control over their overall behavior. An interesting application of profiling is the development of agent sets that reflect demographic distributions \citep{Argyle2023-gf}.

\subsection{Perception}
Perception is an analog to the human sensory system, which interprets multimodal information such as text, images, or auditory data, transforming it into a format comprehensible by LLMs, as demonstrated by SAM\cite{Kirillov2023-jc}, GPT4-V\cite{GPT4-v}, LLaVa\cite{Liu2023-op}, Fuyu8B\cite{FUYU8B-2023}, and BuboGPT\cite{Zhao2023-fn}. In our proposed architecture, the perception is responsible for converting the task and the observations to a data representation that can be understood by the agent. Moreover, advancements in LLMs have led to the development of even more versatile models, such as the any-to-any Next-GPT\cite{Wu2023-pw} and the any-to-text Macaw-LLM\cite{Lyu2023-sa}. Employing such multimodal LLMs in decision-making processes can simplify perception tasks for agents, with several studies exploring their use in autonomous systems\cite{Wang2024-ct, Gao2023-gk}.

\subsection{Tools}

In our proposed definition (see Figure \ref{fig:agents}b), tools or actions are part of the environment. The agent can interact with this environment by deciding which action to execute through the decision-making process. The set of all possible actions that can be selected is also known as the ``action space''. 

The decision process is composed of three main steps: proposal, evaluation, and selection. During the proposal, one or more action candidates are selected using reasoning\cite{Yao2022-db}, code structures\cite{Wang2023-os, Park2023-fr}, or simply by selecting every tool available\cite{Ahn2022-oa}.\cite{Huang2022-rs, Chen2021-vs, Wang2022-lq} The evaluation process consists of evaluating each selected action according to some metric to predict which action would bring more value to the agent. Lastly, the action is selected and executed.

Given that pretrained parameters (sensory memory) are limited, the model must use tools for complex tasks in order to provide reliable answers.
However, LLMs need to learn how to interact with the action space and how and when to use those tools most accurately.\cite{Qin2023-ag} 
LLMs can be pretrained or fine-tuned with examples of tool use, enabling them to operate tools and directly retrieve tool calls from sensory memory during a zero-shot generation.\cite{Nakano2021-od} Recent studies investigate this approach, particularly focusing on open-source LLMs.\cite{Schick2023-oe, Parisi2022-pj, Qian2023-oa} 

As foundational AI models become more advanced, their abilities can be expanded. It was shown that general-purpose foundation models can reason and select tools even with no fine-tuning. 
For example, MRKL\cite{Karpas2022-uc} implements an extendable set of specialized tools known as neuro-symbolic modules and a smart ``router'' system to retrieve the best module based on the textual input. 
Specifically, this router smartly parses the agent's output and selects which neuro-symbolic module is more suitable to perform the task following some heuristic.
These neuro-symbolic modules are designed to handle specific tasks or types of information and are equipped with built-in capabilities and task-relevant knowledge. 
This pre-specialization allows the model to perform domain-specific tasks without needing a separate, domain-specific dataset.
This design addresses the problem of LLMs lacking domain-specific knowledge and eliminates the need for the costly and time-consuming LLM fine-tuning step, using specialized data annotation.\cite{Chen2023-df} The router can receive support from a reasoning strategy to help select the tools\cite{Chen2023-df} or follow a previously created plan\cite{Wang2023-ni}. Recent advances have shown that LLMs can develop new tools of their own\cite{Cai2023-ui, Qian2023-no, Yuan2023-xq}, enabling agents to operate, as needed, in dynamic and unpredictable ``open-worlds'', on unseen problems as illustrated by Voyager\cite{Wang2023-os}. This capability allows agents to evolve and improve continually.

\section{LLM-Based Autonomous Agents in Scientific Research}

The previous section introduced key concepts relevant to any description of the development of autonomous agents. Here, we now focus on which agents were developed for scientific purposes, and ultimately for chemistry. Previous sections of this review have discussed how LLMs could be powerful in addressing challenges in molecular property prediction, inverse design, and synthesis prediction. When we consider the value of agents in chemistry and the ability to combine tools that, for example, search the internet for established synthetic procedures, look up experimental properties, and control robotic synthesis and characterization systems, we can see how autonomous agents powerfully align with the broader theme of automation, which will lead to an acceleration of chemical research and application.

\begin{longtable}{@{\extracolsep{\fill}}m{3.2cm}|cccm{6cm}m{2cm}}

    \caption{Scientific LLM systems and agents. We identify the studies we classified as an agent with the icon \agent\cite{Hilmy-bot} and multi-agent systems with the icon \agent\agent. \W, \S, and \L\cite{Laisa-memory} mean the agent bases his behavior on sensory, short, and long memory components, respectively. Besides the textual capabilities of LLM-based agents, \audio\cite{Freepik-audio} and \view\cite{Kiranshastry-view} mean the agent has additional audio and visual perception, respectively. The release date column displays the date of the first publication for each paper. When available, the publication date of the last updated version is displayed between parentheses.}\label{tab:agents}\\

    Agent   & Memory & Planning & Reasoning & Action & Release date \\\hline
    \endfirsthead

    \multicolumn{6}{c}{{\bfseries \tablename\ \thetable{} -- continued from previous page}} \\
    Agent   & Memory & Planning & Reasoning & Action & Release date \\\hline
    \endhead

    \hline \multicolumn{6}{r}{{Continued on next page}} \\ \hline
    \endfoot
    
    \hline
    \endlastfoot

    PaperQA2\cite{Skarlinski2024-an}\agent
    &\S\L   &\c &\c & Tools to search the scientific literature, gather evidence, and answer questions   & 2024.09\\
    LLaMP\cite{Chiang2024-yq}\agent
    &\S\L &   &\c & Tools for database access, literature search, and atomistic simulations  & 2024.01 (2024.06) \\
    SGA\cite{Ma2024-lj}\agent
    &\S &  & & Employ the LLMs in a optimization loop  & 2024.05 \\
    CRISPR-GPT\cite{Huang2024-pg}\agent\agent
    &\S   &\c & \c  & Tool for gene editing experiments design & 2024.04\\
    TAIS\cite{Liu2024-xm}\agent\agent
    &\S   &\c & \c  & Tools for gene expression data analysis  & 2024.02\\
    ChemReasoner\cite{Sprueill2024-sj}\agent
    &\W\S   &\c & \c  & Tools for heuristic search, 3D structure generation, and prediction using GNNs & 2024.02 (2024.06)\\
    SciAgent\cite{Ma2024-no}\agent
    &\W\L   &\c & \c  & Trained Mistral for tool usage. Evaluated it using MathToolBench's tools & 2024.02\\
    STORM\cite{Shao2024-qk}
    &\S\L   &\c &\c  & Article writing using retrieval from multi-LLM conversations and pre-generated outline & 2024.02 (2024.04)\\
    \citet{Volker2024-nv}
    &\S     &   &   & Regression with ICL and text retrieval  & 2024.02\\
    ProtAgent\cite{Ghafarollahi2024-kj} \agent\agent
    &\S\L   & \c & \c & Tools for proteins information retrieval, analyzing, \textit{de novo} design, and 3D folded structure generation & 2024.01 (2024.05)\\
    Organa\cite{Darvish2024-cp} \agent\audio\view
    &\S\L   & \c & \c & Tools for common lab procedures, reasoning about experimental results, and report writing & 2024.01\\
    PaperQA\cite{Lala2023-ru}\agent
    &\S\L   &\c &\c & Tools to search the scientific literature, gather evidence, and answer questions   & 2023.12\\
    WikiCrow\cite{wikicrow2024}\agent          
    &\S\L   &\c &\c & Uses PaperQA as a tool & 2023.12\\
    Coscientist\cite{Boiko2023-ot}\agent     
    &\S     &\c &\c & Tools for running Python code, web-searching, and interacting with lab equipment & 2023.04 (2023.12)\\
    Eunomia\cite{Ansari2023-qi}\agent
    &\S     &   &\c & Tools for literature and dataset searching and a chain-of--verification loop & 2023.12\\
    CALMS\cite{Prince2023-ar}\agent
    &\L     &   &   & Tools for using the Materials Project API, designing experiments, and using a hardware API to perform the experiment & 2023.12\\
    CoQuest\cite{Liu2023-ye}\agent
    &\S & & \c & Research question generations and tools for literature visualization using a graph organization & 2023.10 (2024.03) \\
    eXpertAI\cite{Wellawatte2023-rw}\agent
    &\S     &   &\c & Tools for applying XAI methods & 2023.11\\
    BioPlanner\cite{ODonoghue2023-tx}\agent
    &\S     &\c &\c & Tool for protocol searching in the BioProt dataset  & 2023.10\\
    IBM ChatChem\cite{Janakarajan2023-fk}\agent
    &\S     &   &   & Tools for cheminformatics and accessing GT4SD and HuggingFace models & 2023.09\\
    ChatMOF\cite{Kang2023-zs}\agent
    &\W\S   &   & \c  & Tools for database search, property prediction, and MOF's structure generation & 2023.08 (2024.06)\\
    AmadeusGPT\cite{Bran2023-jk}\agent\view
    &\S\L   &   &\c & Tools for writing and executing code for computer vision, machine learning, and spatial-temporal reasoning & 2023.07 \\
    i-Digest\cite{Mourino2023-fj} \audio
    &\S     &   &   & Uses the whisper model to process audio transcription from classes and write summaries and following up questions  & 2023.06\\
    BOLLaMa \cite{Rankovic2023-bd}          
    &\W     &   &   & Implements an LLM interface to ease the usage of their BO code & 2023.06\\
    text2concrete\cite{Kruschwitz2023-da}   
    &\S     &   &   & Uses ICL to predict compressive strength from concrete formulation  & 2023.06\\
    MAPI\_LLM\cite{Ramos2023-ps}\agent
    &\S\L   &   &\c &Database access and LLM prediction using ICL& 2023.06\\
    BO-LIFT\cite{Ramos2023-cs}              
    &\S     &   &   &Regression using ICL and text retrieval & 2023.04\\
    ChemCrow\cite{Bran2023-jk}\agent
    &\S     &   &\c & Molecular, cheminformatics, search and critique tools & 2023.04 (2024.05) \\
\end{longtable}

It was \citet{Hocky2021-mr} who discussed the early stages of models that could automate programming and, hence, the expected impacts in chemistry.
Then, early work by \citet{White2023-bi} applied LLMs that could generate code to a benchmark set of chemical problems. In that case, not only were LLMs demonstrated to possess a notable understanding of chemistry, based on accurate question answering, but \citet{White2023-bi} imagined a potential to use them as base models to control knowledge augmentation and a variety of other tools. Thus, these LLMs could be used to execute routine tasks, optimize procedures, and enhance the retrieval of information from scientific literature across a range of scientific domains. To the best of our knowledge, this is the first review of autonomous agents in chemistry that have evolved since these two visionary conceptual perspectives. A deeper exploration follows below.
One driving motivation for the need to augment LLMs with a more pertinent and dedicated knowledge base is the need to circumvent problems of a limited context prompt window, and the restriction that once an LLM is trained, any new information is beyond it's reach since it necessarily has fallen outside its corpus of training data. Furthermore, LLMs are also known to hallucinate. Their predictions are probabilistic and, in science, if experimental evidence is available, then there is great value in building from known domain-specific information. Some improved prompt engineering can aid in the generation of results that are more likely to be accurate, but the use of autonomous agents may solve such problems completely in this next phase of AI in chemistry.
In fact, even adding one or two components when building an agent, as opposed to a whole suite, has shown some significant gains.

Building on this foundation, \citet{Ramos2023-cs} illustrated that LLMs could directly predict experimental outcomes from natural language descriptions, incorporating this ability into a Bayesian optimization (BO) algorithm to streamline chemical processes.
Using in-context learning (ICL), where a model learns from examples provided during inference without requiring retraining, their approach avoided additional model training or fine-tuning, simplifying the optimization process. In a similar vein, \citet{Kristiadi2024-mf} demonstrated similar results with a smaller, domain-specific model, using parameter-efficient fine-tuning (PEFT) rather than ICL.
\citet{Rankovic2023-ge} also explored BO using natural language. They used an LLM to encode chemical reaction procedures, described using natural language, and then trained a Gaussian process (GP) head to predict the reaction yield from the latent encoded representation of the procedure. By keeping the LLM frozen and only updating the multilayer perceptron (MLP) head, this approach minimized training time. \citet{Volker2024-nv} extended these ideas by sampling multiple model completions and adding a verifier model to select the next best step in the BO algorithm. They also used ICL and a short-term memory component to optimize alkali-activated concrete mix design. These examples demonstrate how agent-based systems can execute complex optimization algorithms step by step, directly contributing to automation and more efficient experimental design.

To better promote new ideas regarding AI in scientific research, \citet{Jablonka2023-ng} organized a one-day hackathon in March 2023 where participants developed 14 innovative projects addressing chemical problems centered on predictive modeling, automation, knowledge extraction, and education. Several agent-based approaches emerged from this hackathon.  First, MAPI\_LLM\cite{Ramos2023-ps} is an agent with access to the Materials Project API (MAPI) database that receives a query asking for a property of a material and then retrieves the relevant information from the dataset. If the material is not available on MAPI, the agent can search for similar materials and use in-context learning (ICL) to provide a prediction of the requested property. Additionally, MAPI\_LLM also has a reaction module for synthesis proposal. Second, \citet{Rankovic2023-bd} used LLMs to make BO algorithms more accessible to a broader group of scientists; BOLLaMa implements a natural language interface to easily interact with BO software developed by their group\cite{Rankovic2023-qo}. Third, and similar to \citet{Ramos2023-cs} and \citet{Rankovic2023-ge} who employed LLMs in BO, \citet{Weiser2023-ac} focused on genetic algorithms (GA), a different optimization algorithm. In GA, pieces of information are stochastically combined and evaluated to guide the algorithm during the optimization. For chemistry, these pieces are often molecular fragments that are combined to compose a final whole molecular structure. Thus, \citet{Weiser2023-ac} used LLMs to implement common GA operators under the hypothesis that LLMs can generate new combined molecules better than random cross-over due to their sensory memory. Fourth, InsightGraph\cite{Circi2023-ja} can draw general relationships between materials and their properties from JSON files. \citet{Circi2023-ja} showed that LLMs can understand the structured data from a JSON format and reorganize the information in a knowledge graph. Further refinement of this tool could automate the process of describing relationships between materials across various scientific reports, a task that remains labor-intensive today. Fifth, \citet{Kruschwitz2023-da} used ICL and LLMs to accurately predict the compressive strength of concrete formulations; Text2Concrete achieved predictive accuracy comparable with a Gaussian process regression (GPR) model, with the advantage that design principles can be easily added as context. This model was successfully applied in a BO algorithm following the \citet{Ramos2023-cs} approach.\cite{Volker2024-nv}
For education purposes, multiple authors have raised the discussion about how LLMs can be used to support educators' and instructors' daily work.\cite{Zaabi2023-mx, Kasneci2023-oh, Hellas2023-ks, Dan2023-gs, Tian2024-en} Finally, in this direction, \citet{Mourino2023-fj} developed i-Digest, an agent whose perception module can understand audio tracks and video recordings. These audio recordings are transcribed to text using the Whisper\cite{Radford2022-wq} model, and therefore, i-Digest is a digital tutor that generates questions to help students test their knowledge about the course material. These are just a few examples to showcase the capabilities of AI systems to innovate and generate solutions rapidly.

More recently, \citet{Ma2024-no} showed that agents can be trained to use tools. SciAgent\cite{Ma2024-no} was developed under the premise that finetuning LLMs for domain-specific applications is often impractical. Nevertheless, the agent can be fine-tuned with a set of tools that will enable them to perform well in a domain-specific task. These tools, typically Python functions, enable SciAgent to plan, retrieve, and use these tools to facilitate reasoning and answer domain-related questions effectively. The benchmark developed for SciAgent, known as \texttt{SciToolBench}, includes five distinct domains, each equipped with a set of questions and corresponding tools. The development of its retrieval and planning modules involved finetuning different LLMs on the MathFunc benchmark, resulting in a notable performance improvement of approximately $\sim$~20\% across all domains in \texttt{SciToolBench} compared to other LLMs.

These examples demonstrate the rapidly growing potential of autonomous agents to drive innovation and automation across scientific tasks, from optimizing experiments and materials discovery to enhancing education. As these tools advance, they streamline processes, generate new insights, and empower researchers to tackle complex challenges. By combining reasoning, optimization, and tool usage in real time, agents mark a significant leap in AI-driven research. In the next section, we focus on how agents are transforming literature review processes, a critical aspect of scientific discovery.

\subsection{Agents for Literature Review}

Another fantastic opportunity for automation in the sciences is associated with high-quality literature review, a pivotal aspect of scientific research that requires reading and selecting relevant information from large numbers of papers, and thereby distilling the current state of knowledge relevant to a particular research direction. This extremely time-consuming task is being revolutionized by advanced AI tools designed to automate and enhance such analysis and summarization.

PaperQA introduces a robust model that significantly reduces misinformation while improving the efficiency of information retrieval. This agent retrieves papers from online scientific databases, reasons about their content, and performs question-answering (QA) tasks. Its mechanism involves three primary components—``\texttt{search}'', ``\texttt{gather\_evidence}'', and ``\texttt{answer\_question}'' and the authors adapted the Retrieval-Augmented Generation (RAG)\cite{Lewis2020-bh} algorithm to include inner loops on each step. For instance, PaperQA can perform multiple rounds of \texttt{search} and \texttt{gather\_evidence} if, upon reflection, not have enough evidence has been acquired to successfully \texttt{answer\_question}.

To further validate its capabilities, the authors developed a new benchmark called LitQA, specifically designed to evaluate the performance of models like PaperQA in solving complex, real-world scientific questions. LitQA focuses on tasks that mimic the intricacy of scientific inquiry, comprising 50 multiple-choice questions derived from biomedical papers published post-September 2021, ensuring that these papers were not included in the training data of LLMs. In this challenging setting, PaperQA not only meets but exceeds human performance, achieving a precision rate of 87.9\% and an accuracy score of 69.5\%, compared to the human baseline of 66.8\%.\cite{Lala2023-ru} By applying the RAG technique to full-text scientific papers, PaperQA sets a new standard in QA capabilities, achieving human-like performance in curated datasets without hallucination or selecting irrelevant citations.\cite{Lala2023-ru}

Building on top of PaperQA, WikiCrow exemplifies the practical application of AI in generating concise and relevant Wikipedia-style summaries. The authors show that while 16\% of a human-created Wikipedia article comprises irrelevant statements, WikiCrow displays irrelevant information only 3\% of the time. Their system also added 5\% more correct citations when compared with original articles. Moreover, thanks to its foundation in the PaperQA framework\cite{Lala2023-ru}, WikiCrow achieves remarkable cost-efficiency. The authors estimate that WikiCrow can accomplish in a few days what would take humans approximately 60,000 hours, or about 6.8 years, thereby underscoring its ability to rapidly produce extensive scientific content. This efficiency exemplifies the reliability and transformative potential of AI in content creation \cite{wikicrow2024}.

Following a different approach, the STORM model also addressed the problem of writing Wikipedia-like summaries, where the STORM acronym represents the Synthesis of Topic Outlines through Retrieval and Multi-perspective questions.\cite{Shao2024-qk} This approach implements a two-step procedure. First, STORM retrieves multiple articles on a topic and uses an LLM to integrate various perspectives into a cohesive outline. Second, this outline is used to write each section of the Wikipedia-like summary individually.
To create the outline, multiple articles discussing the topic of interest are retrieved by an ``expert'' LLM, which processes each one to create $N$ perspectives. Each perspective is then fed to a ``writer'' LLM, and a conversation is initiated between writer and expert. Finally, the $N$ conversations are used to design the final outline.
The outline and the set of references, accessed by RAG, are given to the writer LLM. The writer LLM is prompted to use these inputs to generate each section of the article sequentially. Following this, all sections are merged and refined to eliminate redundancies and enhance coherence. Upon human evaluation, STORM is reported to be $\sim$~25\% more organized and present $\sim$~10\% better coverage when compared to a pure RAG approach. However, it was also less informative than human-written Wikipedia pages, and STORM presented a transfer of internet-borne biases, producing emotional articles, which is a major concern.

\subsection{Agents for Chemical Innovation}

Transitioning from literature synthesis to practical chemistry applications, we next explore how LLM-based agents have proven their capabilities to revolutionize routine chemical tasks toward an acceleration of molecular discovery and scientific research. Agents are flexible entities capable of developing prompt-specific workflows and executing a plan toward accomplishing a specific task. ChemCrow\cite{Bran2023-jk} introduced a significant shift in how LLMs would be applied in chemistry, given that LLMs alone do not access information outside of their training data nor can they directly perform chemistry-related tasks. 

By augmenting LLMs with common chemical tools, computational or robotic, ChemCrow automates a broad spectrum of routine chemical tasks, demonstrating a significant leap in LLM applicability. Under human evaluation, ChemCrow consistently outperformed GPT-4, achieving an accuracy score of 9.24/10 compared to 4.79/10.\cite{Bran2023-jk}
The developers of ChemCrow have also considered the ethical implications and potential risks associated with its capabilities. ChemCrow's high potential could be misused and exploited for malicious objectives, and therefore the authors have implemented safety checks and guidelines to prevent such misuse, or ``dual usage''. Additionally, they acknowledge that ChemCrow, relying on an LLM, may not always provide completely accurate answers due to gaps in its chemical knowledge. As such, they recommend careful and responsible use of the tool, along with thorough scrutiny of its outputs. In summary, while ChemCrow presents a powerful new chemical assistant\cite{Bran2023-jk}, oversight of its use is required, and this agent's access to tools has been deliberately limited to enhance security and avoid misuse.

Similarly to ChemCrow\cite{Bran2023-jk}, Chemist-X\cite{Chen2023-cg} uses RAG to get up-to-date literature information and use it to reliably solve a user's questions. Nevertheless, Chemist-X focuses on designing chemical reactions to achieve a given molecule. It works in three phases: (1) First, the agent searches molecule databases for similar molecules to the given molecule, then (2) it searches online literature searching for chemical reactions capable of converting the list of similar molecules in the target. Lastly, (3) machine learning models are used to propose the reaction conditions. To validate their agent, the authors used Chemist-X to design a High-Throughput Screening (HTS) experiment aiming to produce 6-(1-methyl-1H-indazol-4-yl), resulting in a maximum yield of 98.6\%.

Another system called Coscientist\cite{Boiko2023-ot} system exemplifies the integration of semi-autonomous robots in planning, conceiving, and performing chemical reactions with minimal human intervention. At its core, the system features a main module named `PLANNER', which is supported by four submodules. These submodules, or tools, are responsible for performing actions such as searching the web for organic synthesis, executing Python code, searching the hardware documentation, and performing a reaction in an automated lab.\cite{Boiko2023-ot} Utilizing this framework, the Coscientist successfully conducted two types of chemical coupling reactions, Suzuki-Miyaura and Sonogashira, in a semi-automated fashion, with manual handling of initial reagents and solvents. Additionally, Coscientist was also used to optimize reaction conditions.
In contrast to \citet{Ramos2023-cs}, who used LLMs within a Bayesian Optimization (BO) algorithm as a surrogate model, \citet{Boiko2023-ot} approached the optimization task as a strategic ``game'' aimed at maximizing reaction yield by selecting optimal reaction conditions. This demonstrates the ability of GPT-4 to effectively reason about popular chemical reactions -- possibly via comprehensive coverage in pretraining.
The authors have indicated that the code for their agent will be released following changes in U.S. regulations on AI and its scientific applications. At the time of writing, the code remains unreleased, but a simple example that calculates the square roots of random numbers has been provided to illustrate their approach.\cite{Boiko2023-ot}.
These examples underscore the transformative role of LLMs in enhancing and automating chemical processes, which will likely accelerate chemical discovery.

Automated workflows in protein research have also been explored. ProtAgent\cite{Ghafarollahi2024-kj} is a multi-agent system designed to automate and optimize protein design with minimal human intervention. This system comprises three primary agents: \texttt{Planner}, \texttt{Assistant}, and \texttt{Critic}. The \texttt{Planner} is tasked with devising a strategy to address the given problem, the \texttt{Assistant} executes the plan using specialized tools and API calls, and the \texttt{Critic} supervises the entire process, providing feedback and analyzing outcomes. These agents collaborate through a dynamic group chat managed by a fourth agent, the \texttt{Chat Manager}. Tasks executed by this team include protein retrieval and analysis, \textit{de novo} protein design, and conditioned protein design using Chroma\cite{Ingraham2023-hf} and OmegaFold\cite{Wu2024-mc}.

Similarly to ProtAgent, \citet{Liu2024-xm} created a team of AI-made scientists (TAIS) to conduct scientific discovery without human intervention. However, their agents have roles analogous to human roles, such as project manager, data engineer, code reviewer, statistician, and domain expert. While in ProtAgent\cite{Ghafarollahi2024-kj} agents interact through the \texttt{Chat Manager} only, TAIS\cite{Liu2024-xm} enables AI scientists to interact between themselves directly using pre-defined collaboration pipelines. To evaluate TAIS, the authors curated the Genetic Question Exploration (GenQEX) benchmark, which consists of 457 selected genetic data questions. As a case study, the authors show TAIS's answer to the prompt \texttt{``What genes are associated with Pancreatic Cancer when considering conditions related to Vitamin D Levels?''}. The system identified 20+ genes with a prediction accuracy of 80\%. 

Innovation can also be achieved by looking into data from a different point-of-view to get new insights. Automating querying databases was investigated by \citet{Ramos2023-ps} with a ReAct agent with access to the MAPI dataset. This concept was extended by \citet{Chiang2024-yq} using LLaMP,\cite{Chiang2024-yq} which is a RAG-based ReAct agent that can interact with MAPI, arXiv, Wikipedia, and has access to atomistic simulation tools. The authors showed that grounding the responses on high-fidelity information (a well-known dataset) enabled the agent to perform inferences without fine-tuning. 

The agents in chemistry, as exemplified by ChemCrow\cite{Bran2023-jk} and Coscientist\cite{Boiko2023-ot}, highlight a significant shift towards automation and enhanced efficiency in molecular discovery and scientific research. These systems demonstrate the potential of integrating LLMs with chemical tools and automation frameworks, achieving impressive accuracy and effectiveness in tasks ranging from routine chemical operations to complex reaction optimizations. Similarly, ProtAgent\cite{Ghafarollahi2024-kj} and TAIS\cite{Liu2024-xm} systems showcase the versatility of multi-agent frameworks in automating protein design and genetic research, pushing the boundaries of what AI-driven scientific discovery can achieve. These studies collectively showcase the incredible potential of agents in chemical and biological research, promising automation of routine tasks, easing the application of advanced techniques and analyses, and accelerating discoveries. However, they also underscore the necessity for meticulous oversight and responsible development to harness their full potential while mitigating risks.\cite{Tang2024-nx}

\subsection{Agents for Experiments Planning}

Building on the capabilities of ChemCrow and Coscientist in automating chemistry-related tasks, recent advances have focused on bridging the gap between virtual agents and physical laboratory environments.
For example, Context-Aware Language Models for Science (CALMS)\cite{Prince2023-ar}, BioPlanner\cite{ODonoghue2023-tx}, and CRISPR-GPT\cite{Huang2024-pg} focus on giving support to researchers with wet-lab experimental design and data analysis.

CALMS\cite{Prince2023-ar} focuses on improving laboratory efficiency through the operation of instruments and management of complex experiments, employing conversational LLMs to interact with scientists during experiments. In addition, this agent can perform actions using lab equipment after lab equipment APIs have been provided to the agent as tools. 
CALMS was designed to enhance instrument usability and speed up scientific discovery, providing on-the-spot assistance for complex experimental setups, such as tomography scans, and enabling fully automated experiments. For instance, its capability was showcased through the operation of a real-world diffractometer. Although CALMS excelled in several tasks, a comparison between GPT-3.5 and Vicuna 1.5 revealed Vicuna's limitations in handling tools.

In contrast, BioPlanner\cite{ODonoghue2023-tx} significantly improves the efficiency of scientific experimentation by creating pseudocode representations of experimental procedures, showcasing AI's capacity to streamline scientific workflows. Therefore, Rather than interacting directly with lab equipment through APIs, BioPlanner creates innovative experimental protocols that can be expanded upon within a laboratory setting.
The initial step in BioPlanner's process involves assessing the capability of LLMs to produce structured pseudocode based on detailed natural language descriptions of experimental procedures.

In testing, BioPlanner successfully generated correct pseudocode for 59 out of 100 procedures using GPT-4, although the most common errors involved omitted units.
Afterward, the authors used BioPlanned to generate a procedure for culturing an E.coli bacteria colony and storing it with cryopreservation, which ran successfully.

Focusing on gene editing experiments, CRISPR-GPT\cite{Huang2024-pg} is an agent developed to design experiments iteratively with constant human feedback.
CRISPR-GPT\cite{Huang2024-pg} aims to bridge the gap for non-experts by simplifying this process into manageable steps solvable by an LLM with access to useful tools. This agent operates in three modes based on user prompts: ``Meta mode'' provides predefined pipelines for common gene-editing scenarios; ``Auto mode'' uses the LLM to plan a sequence of tasks; and ``Q\&A mode'' answers general questions about the experimental design. The authors demonstrate that based on human evaluations, CRISPR-GPT outperforms GPT-3.5 and GPT-4 in accuracy, reasoning, completeness, and conciseness. Additionally, they applied CRISPR-GPT to design real-world experiments for knocking out TGFBR1, SNAI1, BAX, and BCL2L1 in the human A375 cell line, achieving an editing efficiency of approximately 70\% for each gene.

Following the ideas of developing agents for automating experimental protocol generation, \citet{Ruan2024-lw} created a multi-agents system composed of 6 agents: \texttt{Literature Scouter}, \texttt{Experiment Designer}, \texttt{Hardware Executor}, \texttt{Spectrum Analyzer}, \texttt{Separation Instructor}, and \texttt{Result Interpreter}. The Large Language Models-based Reaction Development Framework (LLM-RDF)\cite{Ruan2024-lw} automates every step of the synthesis workflow. While other studies focus on the literature review\cite{Zheng2023-ka, Zhang2024-ab, Zheng2024-oz}, HTS\cite{Chen2023-cg}, and reaction optimization\cite{Zheng2023-gb, Boiko2023-ot}, LLM-RDF can support researchers from literature search until the product purification. Using this system, the authors showed they could design a copper/TEMPO catalyzed alcohol oxidation reaction, optimize reaction conditions, engineer a scale-up, and purify the products, obtaining a yield of 86\% and a purity >98\% while producing 1 gram of product.

Interestingly, despite covering different fields and having diverse goals, all of these studies, from the fully automated systems like CALMS,\cite{Prince2023-ar} and LLM-RDF, \cite{Ruan2024-lw} to human-driven protocols in BioPlanner\cite{ODonoghue2023-tx} and CRISPR-GPT\cite{Huang2024-pg}, share a ``human-in-the-loop'' approach. This ensures the researcher remains integral to the development process, enhancing reliability and mitigating potential agent limitations, such as errors or hallucinations. Moreover, this approach addresses risks and dual-use concerns, as humans can assess whether the agents' suggestions are safe.\cite{Wang2023-kc, Nascimento2023-vo}
On a slightly different track, Organa\cite{Darvish2024-cp} fully automates the laboratory workload while providing feedback to the researcher and producing reports with the results, as discussed on Section \ref{sec:automation}.

Autonomous agents significantly enhance productivity and efficiency in scientific research, but human creativity and decision-making remain vital to ensure quality and safety. In the next section, we explore agents designed to automate cheminformatics tasks, continuing our focus on how AI systems are reshaping the chemical sciences.

\subsection{Agents for Automating Cheminformatics Tasks}

Cheminformatics consists of applying information technology techniques to convert physicochemical information into knowledge. The process of solving cheminformatics problems commonly involves retrieving, processing, and analyzing chemical data.\cite{Niazi2023-is}
Getting inspiration from ChemCrow\cite{Bran2023-jk} ideas, Chemistry Agent Connecting Tool Usage to Science (CACTUS)\cite{McNaughton2024-qj} focused on assisting scientists by automating cheminformatics tasks. CACTUS automates the applications of multiple cheminformatics tools, such as property prediction and calculation, while maintaining the human-in-the-loop for molecular discovery. The authors investigated the performance of a diverse set of open-source LLMs, where Gemma-7B and Mistral-7B demonstrated superior performance against LLaMA-7B and Falcon-7B. In addition, the authors reported that adding domain-specific information in the prompt to align the agent to chemistry problems considerably increases a model's performance. For instance, predicting drug-likeness with a Gemma-7B agent improves the accuracy of  $\sim$ 60\% when aligning the agent in this way, and prompt alignment improved the prediction of all properties they studied.

Further illustrating the versatility of AI in scientific research and domain-specific tools usage is ChatMOF,\cite{Kang2023-zs} which focuses on the prediction and generation of Metal-Organic Frameworks (MOFs). ChatMOF integrates MOF databases with its MOFTransformer\cite{Kang2023-xd} predictor module, thereby showcasing the innovative use of genetic algorithms in guiding generative tasks from associated predictions. The authors showed that ChatMOF achieved an accuracy of $\sim$~90\% in search and prediction tasks while generative tasks have an accuracy of $\sim$~70\%. The genetic algorithm used by ChatMOF allows for the generation of a diverse array of MOF structures, which can be further refined based on specific properties requested by users. For instance, when prompted to, ``generate structures with the largest surface area'', the system initially generated a broad distribution of structures with surface area centered in 3784 m$^2$/g, and the GA evolves it to a narrower distribution with a peak at 5554 m$^2$/g after only three generations. It is important to note that even though ChatMOF has access to a dataset of experimental values for MOFs, language model predictions guide their GA, and no further validation has been made.
Lastly, \citet{Ansari2023-qi} developed Eunomia, another domain-specific autonomous AI agent that leverages existing knowledge to answer questions about materials. Eunomia\cite{Ansari2023-qi} can use chemistry tools to 
access a variety of datasets, scientific papers and unstructured texts to extract and reason about material science information. The authors implemented a CoVe\cite{Dhuliawala2023-wp} (Consistency Verification) scheme to evaluate the model's answer and minimize hallucination. The authors showed that including CoVe increased the model's precision by $\sim$~20\% when compared to previous methods such as an agent using ReAct only.\cite{Yao2022-db}

Promoting molecular discovery is a topic with great attention in the literature devoted to it and, as described extensively above, LLMs have leveraged a large amount of unstructured data to accelerate that discovery. \citet{Janakarajan2023-fk} discuss the advantages of using LLMs in fields such as \textit{de novo} drug design, reaction chemistry, and property prediction, but they augment the LLM in \texttt{IBM ChemChat}, a chatbot with the capability of using common APIs and python packages commonly used daily by a cheminformatics researcher to access molecular information. \texttt{ChemChat} has access to tools such as Generative Toolkit for Scientific Discovery (GT4SD)\cite{Manica2023-hv}, a package with dozens of trained models generative models for science, rxn4chemistry\cite{rxn4chemistry}, a package for computing chemistry reactions tasks, HuggingMolecules\cite{Gainski2022-an}, a package developed to aggregate molecular property prediction LMs, and RDKit\cite{Landrum2013-nt}, a package to manipulating molecules. Since \texttt{ChemChat} implements an agent in a chat-like environment, users can interactively refine design ideas. Despite being developed to target \textit{de novo} drug design, \texttt{ChemChat} nonetheless is a multi-purpose platform that can be more broadly used for molecular discovery.

In addition to the capabilities described above, LLM-based agents can empower users to tackle tasks that typically require extensive technical knowledge. In previous work, \citet{Wellawatte2023-rw} and \citet{Gandhi2022-mx} showed that including natural language explanations (NLE) in explainable AI (XAI) analysis can improve user understanding. More recently, \citet{Wellawatte2023-rw} developed XpertAI\cite{Wellawatte2023-rw} to seamlessly integrate XAI techniques with LLMs to interpret and explain raw chemical data autonomously. Applying XAI techniques is usually restricted to technical experts but by integrating such techniques with an LLM-based agent to automate the workflow, the authors made XAI accessible to a wider audience.

Their system receives raw data with labels for physicochemical properties. The raw data is used to compute human-interpretable descriptors and then calculate SHAP (or SHapley Addictive exPlanations) values or Z-scores for Local Interpretable Model-agnostic Explanations (LIME). By calculating SHAP values, a value can be assigned to each feature, indicating its contribution to a model’s output. LIME interprets a model by making a local approximation, around a particular prediction, to indicate what factors have contributed to that prediction in the model. It may use, for example, a surrogate local linear regression fit to recognized features.\cite{Gandhi2022-mx}
In addition to XAI tools, XpertAI can search and leverage scientific literature to provide accessible natural language explanations (NLEs).
While ChatGPT provides scientific justifications with similar accuracy, its explanation is often too broad. On the other hand, XpertAI provides data-specific explanations and visual XAI plots to support its explanations.\cite{Wellawatte2023-rw}
With a similar goal, \citet{Zheng2023-em} prompted the LLM to generate explanatory rules from data.

These developments signify a growing trend in the integration of tools and LLMs in autonomous AI within scientific research. By automating routine tasks, enhancing information retrieval and analysis, and facilitating experimentation, AI is expanding the capabilities of researchers and accelerating the pace of scientific discovery. This review underscores the transformative impact of AI across various scientific domains, heralding a new era of innovation and efficiency in chemical research.

\subsection{Agents for Hypothesis Creation}
Following the agent's classification proposed by \citet{Gao2024-zp}, the studies we have discussed previously lie mainly in level 1, i.e. AI agents as a research assistant. Therefore, such agents can support researchers in executing predefined tasks, but they lack the autonomy to propose, test, and refine new scientific hypotheses. New research has been focusing on making agents able to refine scientists' initial hypotheses collaboratively, which is a required skill to achieve level 2 in the \citet{Gao2024-zp} classification.

The idea of an ``AI scientist'' who can generate new, relevant research questions (RQ) has been pursued by groups such as \citet{Wang2023-vc}, who developed a framework called Scientific Inspiration Machines Optimized for Novelty (SciMON). SciMON uses LLMs to produce new scientific ideas grounded in existing literature. It retrieves inspirations from past papers and iteratively refines generated ideas to optimize novelty by comparing them with prior work. Extending these ideas, \citet{Gu2024-cb} used LLMs to search over a knowledge graph for inspiration to propose new personalized research ideas. 
Aligned with this vision, \citet{Liu2023-ye} developed CoQuest, partially automating the brainstorming for new the RQ process. This system uses a human-computing interface (HCI) to allow the agent to create new RQs that can be further enhanced by human feedback. They developed two strategies for RQ generation: breadth-first, where the agent generates multiple RQs simultaneously following the original user's prompt, and depth-first, where multiple RQs are created sequentially, building on the top of the previously generated RQ. For each RQ generation, the agent implements a ReAct\cite{Yao2022-db} framework with tools for literature discovery, hypothesis proposition, refinement, and evaluation. Upon evaluation of 20 HCI doctoral researchers by a post-interaction survey, the breadth-first approach was preferred by 60\% of the evaluators. Interestingly, despite the evaluators' report that the breadth-first approach gave them more control and resulted in more trustworthy RQs, the depth-first had better scores for novelty and surprise. This difference might be caused by the fact that the depth-first uses its own RQ to iterate. This process can introduce new keywords that users have not considered.

Focusing on generating and testing hypotheses, ChemReasoner\cite{Sprueill2024-sj} uses a domain-specific reward function and computational chemistry feedback to validate agent responses. The authors combined a Monte Carlo thought search\cite{Sprueill2023-yd} for catalysis with a reward function from atomistic GNNs trained to predict adsorption energy or reaction energy barriers. While the search is responsible for exploiting literature information and allowing the model to propose new materials, the hypothetic material is further tested by the GNN. This framework was applied to suggest materials for adsorbates, biofuel catalysts, and catalysts for CO$_2$ to methanol conversion. The LLM generated the top five catalysts for each task, with ChemReasoner significantly outperforming GPT-4 based on the reward score.

Similarly, \citet{Ma2024-lj} developed Scientific Generative Agent (SGA) to generate hypotheses and iteratively refine them through computational simulations. Initially, the LLM generates a hypothesis. In the use cases considered, it can be a code snippet or a molecule. In the sequence, a search algorithm is used to find a better initial hypothesis for solving the initial query. Finally, this hypothesis --- code or molecule --- is optimized using a gradient-based algorithm. Lastly, the optimization output serves as feedback to the LLM to iterate. In their molecule design task, the goal was to generate a molecule with a specified HOMO-LUMO gap. The hypothesis is a molecule, that is, a SMILES string and a set of atomic coordinates. The gap is predicted by employing UniMol\cite{Zhou2023-zl}. They showed that SGA could generate molecules based on quantum mechanical properties, but the results were not validated.

\section{Challenges and Opportunities}
LLMs hold great potential in chemistry due to their ability to both predict properties, new molecules, and their syntheses and to orchestrate existing computational and experimental tools. These capabilities enhance the accuracy and efficiency of chemical research and open up new avenues for discovery and innovation. By encapsulating AI models, data analysis software, and laboratory equipment within agent-based frameworks, researchers can harness these sophisticated tools through a unified interface. This approach not only simplifies the interaction with complex systems but also democratizes the immense capabilities of modern computational tools, thereby maximizing their utility in advancing chemical research and development.
Other publications and reviews also shared their opinions on the challenges for the future of LLMs and LLM-based agents in chemistry.\cite{Sumers2023-ct, Miret2024-nt, Du2024-uq, Minaee2024-ii, Zhang2024-te, Hira2024-hj}
Nonetheless, some important challenges and opportunities for progress remain, which we summarize here.

\paragraph{Data Quality and Availability} 
Quality and availability of data are critical factors that influence the efficacy of LLMs. Indeed, scaling both the model size and the amount of training data used has proven to improve capabilities.\cite{OpenAI2023-md} However, current AI models are not trained on large amounts of chemical data, which limits their capabilities to reason about advanced chemical concepts.\cite{Mirza2024-qw}

There are two types of datasets commonly used to train LLMs: unlabeled and labeled datasets.
Unlabeled datasets, or pretraining data, are used during the semi-supervised training, which focuses on creating a ``prior belief'' about a molecule. Currently, we have huge datasets composed of hypothetical and/or theoretical data. When a model is trained on data that is not grounded in real chemical information, this might cause the model to learn a wrong prior belief.\cite{Gao2020-ey}

Labeled datasets, often used in benchmarks, also suffer from their inclusion of hypothetical and calculated data. Benchmarks are necessary for quantifying improvements in AI modeling and prediction within a competitive field. However, dominant benchmarks like MoleculeNet,\cite{wu_moleculenet_2018} have significant limitations that may restrict the generalizability and applicability of evolving models. In his blog, \citet{pat_walters_blog3} brings to light numerous errors and inconsistencies within the MoleculeNet data, which substantially impact model performance and reliability.\cite{Gloriam2019-yf, irwin_chemformer_2022, liu_multi-modal_2023, Livne2023-bu}. Walters also argues that the properties present in these benchmarks do not directly correlate with real chemistry improvement. As such, new benchmarks need to translate to practical chemistry problems directly. For instance, increasing accuracy in predicting LogP is not necessarily mapped to drugs with greater bioavailability. Some promising work has come from the Therapeutic Common Data (TDC),\cite{Huang2021-ug, TDC-ss} which includes data from actual therapeutic essays, providing a more practical foundation for model training. 

The community continues to work to organize and curate datasets to prepare data for LLM training and evaluation. Scientific benchmarks\cite{Peng2019-yb, Lala2023-ru, Liu2024-hv}, repositories with curated datasets\cite{_jablonka_awesome-chemistry-datasets}, and packages for model evaluation\cite{Mirza2024-qw} have been developed.
However, the challenges concerning grounded truth and consistent datasets remain.
With advancements in scientific document processing,\cite{Swain2016-se} there is now the opportunity to obtain new datasets from peer-reviewed scientific papers\cite{Laurent2024-dj, Alampara2024-zx, Song2023-oh, Zaki2024-mt}.
Due to the multi-modal capabilities of such AI models, these new benchmarks can comprise multiple data types, potentially enhancing the applicability and transferability of these models. The continual curation of new, relevant datasets that represent the complexities of real-world chemical problems will further enhance the robustness and relevance of LLMs in chemistry.

\paragraph{Model Interpretability} 
Model interpretability is a significant challenge for LLMs due to their ``black-box'' nature, which obscures the understanding of how predictions are made.
However, innovative approaches are being developed to enhance LLMs' interpretability. For instance, \citet{Schwaller2021-sr} and \citet{Schilter2023-ux} used information from the different multi-attention heads. While \citet{Schwaller2021-sr} connected atoms from reactants to atoms in the products, \citet{Schilter2023-ux} assigned H-NMR peaks to specific hydrogens in a molecule to indicate how spectra were comprehended, or structures deduced. Additionally, since the LLMs use language, which is intrinsically interpretable, LLMs may be incrementally modified to explain their reasoning processes directly, exemplified with tools like eXpertAI\cite{Wellawatte2023-rw} and or simply adjusting prompting\cite{Wei2022-rm, Chen2023-df}. These methods address the critical need for transparency in the mechanism of understanding for a good prediction beyond the good prediction itself.

\paragraph{Integration with Domain Knowledge and Cross-disciplinary Applications} While LLMs excel at pattern recognition, integrating explicit chemical rules and domain knowledge into these systems remains challenging. This integration is essential to make predictions that are not only statistically valid but also chemically reasonable. It was shown by \citet{Beltagy2019-ri} and \citet{Gu2021-yn} that better performance on common NLP tasks can be achieved by developing a vocabulary and pretraining on a domain-specific training corpus. While pretraining with domain-specific datasets that include chemical properties, reaction mechanisms, and experimental results may better capture the nuances of chemistry, but using AI to foster multi-disciplinary research remains a significant challenge. The Galactica LLM \cite{Taylor2022-pu} also used special tokens for delineating chemical information, to relatively good success on chemistry tasks. \citet{Aryal2024-na} also progress by creating an ensemble of specialist agents with different domains of knowledge, allowing them to interact to better answer the user query. Specifically, \citet{Aryal2024-na} used agents with chemistry, physics, electrochemistry, and materials knowledge. 

\paragraph{Tool Development} 
The effectiveness of a combined LLM/autonomous agent approach hinges significantly on the availability and quality of the tools, as well as on the complexity and diversity of the chemical tasks at hand.
Some emphasis should be placed on refining standalone tools, with the confidence that overarching frameworks, like a GPT-4-type wrapper, or ``assistant'', will eventually integrate these tools seamlessly. Developers should stay informed about existing tools and design their tools to interface effectively with such a wrapper. This ensures that each tool is ready to contribute its unique capabilities to a cohesive agent system.

\paragraph{Reinforcement Learning} 
RL has been successfully used in LLMs\cite{Bohm2019-ue, Pan2023-lb, OpenAI2022-RLHF}, with a few applications also proposed for use in agents\cite{Hu2023-sw, Xu2023-et}. The next frontier is applying RL to agents directly, to improve their ability on specific tasks. \citet{bou2024acegen} provided a recent framework and example for generative molecular design when viewed as an RL problem (similar to RLHF) and some early success has been seen in applying the RLHF algorithm directly to protein language models where the reward model comes from scientific tasks.\cite{hayes2024simulating} Neither of these are direct RL on language model agents, but are a step towards this goal.

\paragraph{Agent Evaluation} 
Comparing different agent systems is challenging due to the lack of robust benchmarks and evaluation schemes. Consequently, it is difficult to define what constitutes a ``superhuman'' digital chemist and reach a consensus on the criteria for success.\cite{Morris2023-jt, Morris2023-jt, IBM2024-su}
This issue is similar to the ongoing discussions about defining artificial general intelligence (AGI) and the expected capabilities of cognitive architectures.\cite{Langley2009-eh, Goertzel2014-nm}
Once a reliable metric for evaluating such AI systems is established, it is crucial for the AI scientific community to set clear guidelines for conducting research. Currently, assessing success is challenging because the goals are not well defined. Building on this, we propose using Bloom’s taxonomy\cite{bloom1968taxonomy,bloom2010taxonomy} as a reference point for developing a metric to evaluate more complex reasoning and tool use in autonomous agents. This educational framework categorizes cognitive skills in a hierarchical manner, from basic recall to creative construction, providing a structured approach to assess higher-order thinking and reasoning capabilities in these systems. This adaptation could significantly enhance the evaluation of LLMs and autonomous agents, especially when tackling complex chemical challenges.

\paragraph{Ethical and Safety Concerns} 
As with all AI technologies, deploying LLMs involves ethical considerations, such as biases in predictions and the potential misuse of AI-generated chemical knowledge.
\citet{Ruan2023-oq} and \citet{Tang2024-nx} highlight the need for multi-level regulation, noting that current alignment methods may be insufficient for ensuring safety and that human evaluation alone is not scalable.

The absence of specialized models for risk control and reliable safety evaluations poses a significant challenge to ensuring the safety of tool-using LLMs. This highlights the urgent need to automate red-teaming strategies to reinforce AI safety protocols.
Additionally, the development of safe AI systems should prioritize minimizing harmful hallucinations. While managing dual-use risks is a human responsibility and should be controlled through safety assessments at publication or indirect regulation by the scientific community.

\paragraph{Human-AI Collaboration in Chemical Research}
LLMs are poised to transform fields such as drug discovery, materials science, and environmental chemistry due to their ability to predict chemical properties and reactions with remarkable accuracy. Models based on architectures like BERT have demonstrated their capability to achieve state-of-the-art performance in various property prediction tasks.\cite{li_mol-bert_2021,wang_smiles-bert_2019} Furthermore, studies by \citet{Jablonka2023-bm} and \citet{Born2023-nc} have showcased the predictive power of LLMs by reformulating traditional regression and classification tasks as generative tasks, opening up new avenues for chemical modeling. However, as emphasized by \citet{Weng2023-ic}, maintaining the reliability of LLM outputs is essential, as inaccuracies in formatting, logical reasoning, or content can significantly impede their practical utility.
Hallucination is also an intrinsic issue with LLMs.\cite{Ji2023-cg} Though agents can deal with hallucinations to some extent by implementing sanity-checking tools, it does not make the response hallucination-proof. A possible approach to address this issue is to use a human-in-the-loop approach, where steps of human-agent interaction are added to the workflow to check if the agent is in the correct pathway to solve the request.\cite{Cai2023-pw, Xiao2023-qg, Drori2024-pk}

The potential of LLMs to design novel molecules and materials was highlighted by the AI-powered robotic lab assistant, A-Lab, which synthesized 41 new materials within just 17 days.\cite{szymanski_autonomous_2023} Nonetheless, this achievement has sparked debates about the experimental methods and the actual integration of atoms into new crystalline materials, raising questions about the authenticity of the synthesized structures.\cite{peplow_robot_2023} These controversies underline the necessity for rigorous standards and the critical role of human expertise in validating AI-generated results. Again, the integration of advanced AI tools with the oversight of seasoned chemists is crucial, suggesting that a hybrid approach could significantly enhance both the innovation and integrity of materials science research.

In parallel, we have seen how LLM-based agents are increasingly capable of automating routine tasks in chemical research, which traditionally consume significant time and resources. These models excel in real-time data processing, managing vast datasets, and even conducting comprehensive literature reviews with minimal human intervention. Advances in AI technology now allow agents not only to perform predefined tasks but also to adapt and develop new tools for automating additional processes. For instance, tasks such as data analysis, literature review, and elements of experimental design are now being automated.\cite{Wellawatte2023-rw, Hong2024-lg, Qi2024-qe, Lala2023-ru, ODonoghue2023-tx, Prince2023-ar, Bran2023-jk, Boiko2023-ot} This automation liberates chemists to focus on more innovative and intellectually engaging aspects of their work, and the opportunity is to expand productivity and creativity in their science.

\paragraph{Promotion of Impactful Discoveries}
AI technologies offer experimental chemists significant opportunities to streamline repetitive tasks like data collection and analysis, freeing up time for innovation.\cite{Darvish2024-cp}
AI-powered tools can suggest novel experiments and chemical pathways,\cite{Ruan2024-lw} but the black-box nature of many models raises concerns about trust and transparency.
Human expertise remains essential to validate AI-generated results, especially in critical experiments.

A key challenge is translating AI predictions into real-world experiments, where factors like reagent quality and equipment limitations must be considered.
To integrate AI effectively in the lab, stronger collaboration between computational and experimental chemists is essential, ensuring AI tools are practical and aligned with lab conditions.\cite{Tai2020-gu}
Clear communication will help identify the most impactful AI advancements, ensuring tools address the real needs of experimentalists. 
AI's ability to explore new chemical spaces also offers exciting opportunities for discovery, allowing chemists to harness these insights while maintaining oversight for accuracy and reliability.

\paragraph{AI in Everyday Chemistry}

In the near future, AI tools will become integral to the daily workflow of chemists, transforming how routine challenges are approached and resolved. Today's chemist may soon find themselves interacting directly with AI-driven systems, leveraging advanced simulations, literature analyses, and predictive models to accelerate discovery. While this may sound like a glimpse into the future, the reality is that such tools are already emerging, and their widespread adoption has likely already begun. To illustrate this transformation, we propose the following scenario, based largely on prior experience.\cite{zheng_impact_2017}

A chemist working on synthesizing a challenging target molecule encounters suboptimal yields and an unexpected side product. Despite verifying solvent purity, reaction conditions, and ruling out possible causes such as steric hindrance, or leaving group viability, the issue remains unresolved. The chemist plans a comprehensive systematic study, varying the leaving group and adjusting the length of a a bulky alkyl chain in one of the secondary amines.\cite{zheng_impact_2017} This would require weeks of repeated testing and data analysis, creating two lengthy projects for PhD students, diverting 2-3 months of effort. Nonetheless, the starting materials are ordered, and are expected to arrive within a fortnight.

In contrast, another chemist, equipped with methods described here, approaches the problem differently. Through \textit{in-silico} studies, they evaluate the chemical properties of reactants and intermediates using a selection from the chemistry-specific LLMs described above. This strategy allows for rapid hypothesis testing and simulation of reaction conditions. With a human-in-the-loop workflow, the chemist refines the predictions, dismissing implausible pathways and focusing on a promising hypothesis. They use tools like PaperQA2\cite{Skarlinski2024-an} to verify the reaction mechanism against existing literature, ensuring a solid foundation in prior knowledge.
This AI-driven workflow enables the chemist to design three targeted experiments, each validating a critical model prediction, thus bypassing the need for a larger methodological studies. Using an automated ChemCrow system,\cite{Bran2023-jk} the required starting materials are synthesized overnight. The following day, a PhD chemist performs the reactions, swiftly confirming the AI-derived hypothesis and achieving the desired product within 24 hours. The entire process, from problem identification to successful synthesis, concludes in just one week. Meanwhile, the first group of PhD students continues their extensive exploration of reaction conditions, gaining methodological insights but without directly achieving their original goal.

This comparison underscores how creativity and efficiency in research may benefit from a hybrid approach where there is some computational heavy lifting, along with a team of virtual chemistry experts to help hone and test ideas.

\section{Conclusions}

Since this review is targeted in part to an audience of chemists, who may not have yet embraced AI technology, we consider it valuable to point out our perspective that AI in chemistry is definitely here to stay. 
We predict that its use will only grow as a necessary tool that will inevitably lead to more jobs and greater progress. We hope to facilitate the change by connecting the technology to the chemical problems that our readership is already addressing through more traditional methods.

Large Language Models (LLMs) have demonstrated remarkable potential in reshaping chemical research and development workflows.
These models have facilitated significant advancements in molecular simulation, reaction prediction, and materials discovery.
In this review, we discussed the evolution of LLMs in chemistry and biochemistry.
Successful cases where LLMs have proven their potential in promoting scientific discovery were shown with caveats of such models.

Adopting LLM-based autonomous agents in chemistry has enhanced the accuracy and efficiency of traditional research methodologies and introduced innovative approaches to solving complex chemical problems. 
Looking forward, the continued integration of LLMs promises to accelerate the field's evolution further, driving forward the frontiers of scientific discovery and technological innovation in chemistry. 
We have shown how agents have been used in chemistry and proposed a framework for thinking about agents as a central LLM followed by interchangeable components. 

However, despite the community's astonishing advances in this field, many challenges still require solutions. 
We identified the main challenges and opportunities that need to be addressed to promote the further development of agents in chemistry.
Addressing the challenges related to model transparency, data biases, and computational demands will be crucial for maximizing their utility and ensuring their responsible use in future scientific endeavors.

While there are significant challenges to be addressed, the opportunities presented by LLMs in chemistry are vast and have the potential to fundamentally alter how chemical research and development are conducted. Effectively addressing these challenges will be crucial for realizing the full potential of LLMs in this exciting field.
To keep pace with the ever-growing number of relevant publications, we will maintain a repository with an organized structure listing new studies regarding LLMs and LLM-based agents focused on scientific purposes.
The repository can be found in \url{https://github.com/ur-whitelab/LLMs-in-science}

\subsection*{Author contribution}
All authors contributed to writing this review article.

\subsection*{Competing interests}
The authors have no conflicts to declare.

\subsubsection*{Acknowledgments}
M.C.R. and A.D.W. acknowledge the U.S. Department of Energy, Grant No. DE-SC0023354, and C.J.C. gratefully acknowledges the Jane King Harris Endowed Professorship at Rochester Institute of Technology for the support provided for this publication. We are grateful for feedback from early drafts of this review from the following colleagues: Kevin Jablonka, Philippe Schwaller, Michael Pieler, Ryan-Rhys Griffiths, Geemi Wellawatte, and Mario Krenn.

\bibliographystyle{unsrtnat}
\bibliography{refs, Chris3}



\end{document}